\documentclass[sigconf]{acmart}

\usepackage[utf8]{inputenc} 
\usepackage[T1]{fontenc}    
\usepackage{CJKutf8}        

\usepackage{amsmath, amsthm, amsfonts}
\usepackage{graphicx}
\usepackage{subcaption}
\usepackage{booktabs}
\usepackage{array, makecell, multirow, colortbl}
\usepackage{xcolor} 
\usepackage{tikz}
\usetikzlibrary{trees}
\usepackage{url}
\usepackage{hyperref}
\usepackage{cleveref}
\usepackage{nicefrac, microtype, enumitem, appendix, pifont}
\usepackage{listings, tcolorbox}
\usepackage{changepage}
\usepackage[ruled,vlined]{algorithm2e}

\newtheorem{definition}{Definition}

\definecolor{blue1}{HTML}{EEF4FA}
\definecolor{blue2}{HTML}{E3EDF7}
\definecolor{blue3}{HTML}{D7E5F4}
\definecolor{blue4}{HTML}{D0E1EE}
\definecolor{blue5}{HTML}{C9DDF0}

\definecolor{green1}{HTML}{E4F7F9}
\definecolor{green2}{HTML}{D8F3F5}
\definecolor{green3}{HTML}{D1EFF2}
\definecolor{green4}{HTML}{E2F3E7}
\definecolor{green5}{HTML}{D6EADB}

\definecolor{yellow1}{HTML}{FFFDF8}
\definecolor{yellow2}{HTML}{FFF9E3}
\definecolor{yellow3}{HTML}{FFF6D8}
\definecolor{yellow4}{HTML}{FFF2C8}
\definecolor{yellow5}{HTML}{FFEEC0}

\tikzset{
  every node/.style = {draw, rounded corners, font=\small, align=left, text width=4cm},
  level 1/.style = {sibling distance=55mm, level distance=15mm},
}

\setcopyright{none}
\renewcommand\footnotetextcopyrightpermission[1]{} 

\acmConference[KDD'26]{KDD 2026}{August 9 to 13, 2026}{Jeju, Korea}

\definecolor{mygreen}{RGB}{51,102,0}
\definecolor{myred}{RGB}{204, 0, 0}
\definecolor{myblue}{RGB}{222,242,255}

\newcommand{\projectname}{\texttt{FinWorld}\xspace}

\begin{document}

\title{FinWorld: An All-in-One Open-Source Platform for End-to-End Financial AI Research and Deployment}


\author{Wentao Zhang}
\authornote{These authors contributed equally to this work.}
\affiliation{%
\institution{Nanyang Technological University \\ Skywork AI}
\country{Singapore}}
\email{zhangwent963@gmail.com}

\author{Yilei Zhao}
\authornotemark[1]
\affiliation{%
\institution{Nanyang Technological University}
\country{Singapore}}
\email{YILEI002@e.ntu.edu.sg}

\author{Chuqiao Zong}
\affiliation{%
\institution{Nanyang Technological University}
\country{Singapore}}
\email{ZONG0005@e.ntu.edu.sg}

\author{Xinrun Wang}
\authornote{Corresponding author. Email: xrwang@smu.edu.sg}
\affiliation{%
\institution{Singapore Management University}
\country{Singapore}}
\email{xrwang@smu.edu.sg}

\author{Bo An}
\affiliation{%
\institution{Nanyang Technological University \\ Skywork AI}
\country{Singapore}}
\email{boan@ntu.edu.sg}


\begin{abstract}

  Financial AI holds great promise for transforming modern finance, with the potential to support a wide range of tasks such as market forecasting, portfolio management, quantitative trading, and automated analysis. However, existing platforms remain limited in task coverage, lack robust multimodal data integration, and offer insufficient support for the training and deployment of large language models (LLMs). In response to these limitations, we present \projectname, an all-in-one open-source platform that provides end-to-end support for the entire financial AI workflow, from data acquisition to experimentation and deployment. \projectname distinguishes itself through native integration of heterogeneous financial data, unified support for diverse AI paradigms, and advanced agent automation, enabling seamless development and deployment. Leveraging data from 2 representative markets, 4 stock pools, and over 800 million financial data points, we conduct comprehensive experiments on 4 key financial AI tasks. These experiments systematically evaluate deep learning and reinforcement learning algorithms, with particular emphasis on RL-based finetuning for LLMs and LLM Agents. The empirical results demonstrate that \projectname significantly enhances reproducibility, supports transparent benchmarking, and streamlines deployment, thereby providing a strong foundation for future research and real-world applications. Code is available at Github~\footnote{https://github.com/DVampire/FinWorld}.
  
\end{abstract}





\maketitle

\section{Introduction}

\begin{table}[htbp]
    \centering
    \footnotesize
    \renewcommand{\arraystretch}{1.0}
    \setlength{\tabcolsep}{5pt}
    \caption{Comparison of \projectname and related platforms. \textcolor{mygreen}{\ding{51}}: Supported, \textcolor{myred}{\ding{55}}: Not supported, \textcolor{gray}{Partial}: Partially supported.}
    \vspace{-1em}
    \label{tab:platform}
    \begin{tabular}{lcccc}
    \toprule
    \textbf{Key Features} & \textbf{FinWorld} & \textbf{TradeMaster} & \textbf{Qlib} & \textbf{FinRL-Meta} \\
    \midrule
    Multi-task support & \textcolor{mygreen}{\ding{51}}      & \textcolor{gray}{Partial} & \textcolor{gray}{Partial}      & \textcolor{gray}{Partial} \\
    
    Multi-modal data & \textcolor{mygreen}{\ding{51}}      & \textcolor{myred}{\ding{55}}        & \textcolor{myred}{\ding{55}}      & \textcolor{myred}{\ding{55}}        \\
    ML, DL, RL  & \textcolor{mygreen}{\ding{51}}      & \textcolor{gray}{Partial}       & \textcolor{gray}{Partial}     & \textcolor{gray}{Partial}        \\
    
    LLMs  & \textcolor{mygreen}{\ding{51}}      & \textcolor{myred}{\ding{55}} & \textcolor{myred}{\ding{55}}      & \textcolor{myred}{\ding{55}} \\
    
    LLMs Agent & \textcolor{mygreen}{\ding{51}}      & \textcolor{myred}{\ding{55}}        & \textcolor{myred}{\ding{55}}      & \textcolor{myred}{\ding{55}}        \\
    
    Extensibility & \textcolor{mygreen}{\ding{51}}      & \textcolor{mygreen}{\ding{51}}        & \textcolor{gray}{Partial} & \textcolor{gray}{Partial} \\
    
    Auto presentation    & \textcolor{mygreen}{\ding{51}}      & \textcolor{gray}{Partial} & \textcolor{myred}{\ding{55}}      & \textcolor{myred}{\ding{55}}        \\
    
    Distributed Training   & \textcolor{mygreen}{\ding{51}}      & \textcolor{myred}{\ding{55}} & \textcolor{myred}{\ding{55}}      & \textcolor{myred}{\ding{55}}        \\
    
    Benchmarking  & \textcolor{mygreen}{\ding{51}}      & \textcolor{mygreen}{\ding{51}}        & \textcolor{mygreen}{\ding{51}}     & \textcolor{gray}{Partial} \\
    
    \bottomrule
    \end{tabular}
    \vspace{-2.5em}
    \end{table}

Financial AI has revolutionized how we approach market analysis, trading strategies, and investment decisions. From traditional quantitative machine learning (ML) models~\cite{fama2015five,ke2017lightgbm} to modern deep learning (DL) and reinforcement learning (RL) architectures~\cite{jiang2017deep, yu2019model, liu2020adaptive, liu2020finrl}, the field has witnessed remarkable progress across diverse domains including time series forecasting~\cite{feng2023multi, ding2020hierarchical}, algorithmic trading \cite{liu2020finrl, qin2024earnhft}, portfolio management~\cite{zhang2024reinforcement, kim2025semi}, and natural language analysis for financial documents~\cite{liu2023fingpt,liu2025fin}. However, this rapid advancement has created a fragmented ecosystem where researchers and practitioners must navigate multiple specialized tools, each optimized for specific tasks but lacking seamless integration.

Current financial AI platforms, while valuable in their respective domains, face several critical limitations. For instance, TradeMaster~\cite{sun2023trademaster} provides RL methods for four financial trading tasks but lacks compatibility with traditional quantitative models and financial LLMs. Qlib~\cite{yang2020qlib} provides robust quantitative investment tools but offers limited support for modern AI paradigms such as LLMs and RL. Similarly, FinRL-Meta~\cite{liu2022finrl} focuses specifically on RL for trading but suffers from a rigid framework architecture that makes it difficult to integrate new algorithms and lacks support for DL and LLM-based approaches. Most existing platforms are designed for narrow task scopes: some excel at time series forecasting but lack trading simulation, while others offer robust algorithmic trading or portfolio management yet have limited support for modern LLMs and agent paradigms. Furthermore, the integration of heterogeneous financial data ranging from structured market data to unstructured news and reports remains challenging, often requiring complex preprocessing pipelines and custom development efforts.

In summary, current financial AI platforms still face four major challenges: i) \textbf{Limited Task Coverage}: Insufficient support for emerging paradigms such as large language models and autonomous agents. ii) \textbf{Heterogeneous Data Integration}: Inadequate integration of heterogeneous data sources, including structured market data, unstructured news, and multimodal financial information. iii) \textbf{Rigid Framework Architecture}: Framework architectures that impede the seamless integration of novel algorithms and methodologies. iv) \textbf{Standardized Evaluation and Presentation}: Lack of standardized evaluation protocols and presentation frameworks for comprehensive performance assessment.

To address these challenges, we propose \projectname, an all-in-one open-source platform for end-to-end financial AI research and deployment. \projectname provides a unified framework that seamlessly integrates diverse AI paradigms, heterogeneous data sources, and modern technologies to enable comprehensive financial AI development and evaluation. As shown in Table~\ref{tab:platform}, \projectname offers comprehensive support across all key features compared to existing platforms. The key features of \projectname include:
\begin{itemize}[leftmargin=*, itemsep=0.6em, topsep=0.3em]
    \item \textbf{Multi-task Support}: Unified platform supporting time series forecasting, algorithmic trading, portfolio management, and LLM applications, while existing platforms focus on narrow task scopes.

    \item \textbf{Multimodal Data Integration}: Native support for heterogeneous financial data, including structured market data, unstructured news, and multimodal information.

    \item \textbf{Comprehensive AI Paradigms}: Full support for ML, DL, RL, LLMs, and LLM agents, enabling seamless integration of traditional and modern AI approaches.

    \item \textbf{High Extensibility}: Modular and extensible framework design enabling rapid integration of novel algorithms, with flexibility facilitating research prototyping and real-world applications.

    \item \textbf{Advanced Automation}: With automated presentation and reporting features, and support for distributed multi-GPU training and testing, the system enables efficient exploration in multi-environments and supports high-performance computing.
\end{itemize}

To demonstrate the effectiveness of our proposed platform, we conduct extensive experiments and provide comprehensive benchmarks. Our main contributions are threefold:
\begin{itemize}[leftmargin=*, itemsep=0.6em, topsep=0.3em]
  \item \textbf{Unified Framework}: We propose a unified, end-to-end framework for training and evaluation of ML, DL, RL, LLMs, and LLM agents, covering four critical financial AI task types including time series forecasting, algorithmic trading, portfolio management, and LLM applications.

  \item \textbf{Modular Design}: The framework features a modular architecture that enables flexible construction of custom models and tasks, including the development of personalized LLM agents. The system supports efficient distributed training and testing across multiple environments.

  \item \textbf{Comprehensive Benchmark}: We provide support for multimodal heterogeneous data with over 800 million samples, establishing a comprehensive benchmark for the financial AI community. Extensive experiments across four task types demonstrate the framework's flexibility and effectiveness.
\end{itemize}

\section{Related Work}

\subsection{ML, DL, and RL in Financial AI}
Financial time series forecasting has evolved from traditional statistical methods like ARIMA~\cite{box1976time} and GARCH~\cite{engle1982autoregressive} to modern DL approaches. ML methods such as LightGBM~\cite{ke2017lightgbm} and XGBoost~\cite{chen2016xgboost} demonstrated superior performance in capturing complex financial patterns, while DL models including LSTM~\cite{hochreiter1997long} and Transformer~\cite{vaswani2017attention} architectures have revolutionized temporal modeling. Recent advances in financial forecasting include TimesNet~\cite{wu2022timesnet}, DLinear~\cite{zeng2023are}, and Timexer~\cite{liu2024timexer}, which leverage attention mechanisms for improved prediction accuracy. In algorithmic trading, RL methods such as PPO~\cite{schulman2017proximal} and DQN~\cite{mnih2015human} have enabled adaptive trading strategies~\cite{jiang2017deep, liu2020finrl}, with recent work exploring transformer-based approaches~\cite{qin2024earnhft}. Portfolio management has similarly progressed from classical mean-variance optimization~\cite{markowitz1952portfolio} to data-driven approaches using ML~\cite{heaton2017deep} and RL~\cite{zhang2024reinforcement, kim2025semi}. Despite these advances, existing approaches often focus on individual tasks, lacking unified frameworks for comprehensive financial AI development.

\subsection{LLMs and LLM Agents in Financial AI}
The integration of LLMs into financial decision-making has followed two main paths. The traditional approach uses pre-training and supervised fine-tuning (SFT), with models like FinBERT~\cite{araci2019finbert} and FLANG~\cite{shah2022flang} excelling in financial text understanding. Recent progress includes FinGPT~\cite{liu2023fingpt}, BloombergGPT~\cite{wu2023bloomberggpt} with domain-adapted tokenization, and FinQA~\cite{chen2021finqa} featuring numerical reasoning. A newer direction combines pre-training with reinforcement learning (RL) to enhance reasoning, as shown by Fin-R1~\cite{liu2025fin} and Fino1~\cite{qian2025fino1}. Despite their strengths, LLM-based methods often lack sequential decision-making, are computationally expensive, and struggle with non-stationary markets.

To overcome the limitations of traditional LLMs in sequential decision-making, researchers have introduced agentic mechanisms such as memory, tool use, and self-reflection. Early systems like FinMem~\cite{yu2024finmem} and FinRobot~\cite{zhou2024finrobot} rely on text-only LLMs enhanced with layered memory, profiling, and chain-of-thought reasoning to improve single-asset trading performance. Building on this, FinAgent~\cite{zhang2024multimodal} represents the first multimodal trading agent, which jointly processes news, structured price data, and K-line chart visuals via a tool-augmented dual-reflection architecture, achieving significant improvements in profit. Recent efforts shift toward collaborative multi-agent systems: FinCon~\cite{yu2024fincon} adopts a manager–analyst hierarchy with verbal reinforcement, while TradingAgents~\cite{xiao2024tradingagents} models a full trading firm with specialized agents coordinating to enhance returns and risk control. Beyond these architectural innovations, recent work has also explored using RL to finetune LLMs Agents~\cite{feng2025group,agent-r1}, enabling dynamic exploration and reasoning capabilities in real-world environments.


\subsection{Financial AI Platforms}
The development of financial AI platforms has been driven by the need to provide standardized environments for research and deployment. Qlib~\cite{yang2020qlib} provides a quantitative investment platform that integrates traditional ML pipelines with financial data processing, offering tools for feature engineering, model training, and portfolio optimization.TradeMaster~\cite{sun2023trademaster} offers a comprehensive RL-based trading framework with modular components for strategy development, backtesting, and evaluation across multiple financial markets. FinRL-Meta~\cite{liu2022finrl} focuses on reinforcement learning environments and benchmarks, providing standardized market simulators and evaluation metrics for RL-based trading strategies. These platforms have significantly contributed to the advancement of financial AI research by providing researchers and practitioners with accessible tools and standardized evaluation frameworks. However, most existing platforms primarily focus on traditional ML, DL, and RL methods, with limited support for LLMs and LLMs Agents, and lack comprehensive multimodal data integration capabilities. This motivates our proposal of \projectname, a comprehensive framework that covers most critical financial models and tasks. 

\section{Preliminaries}
\label{sec:preliminaries}
In this section, we formally define the core tasks considered in this work. For completeness, we provide mathematical formulations for time series forecasting, algorithmic trading, portfolio management, and LLMs applications in the financial domain.

We first introduce the notations used throughout this paper. For single asset scenarios, $\mathbf{x}_{1:T} = \{x_1, ..., x_T\} \in \mathbb{R}^{T \times D}$ denote the historical endogenous time series (e.g., open, high, low, close price), and for multi-asset scenarios, $\mathbf{x}_{1:T} \in \mathbb{R}^{N \times T \times D}$ denote the historical price sequences for $N$ assets. $\mathbf{z}_{1:T_{\mathrm{ex}}} = \{\mathbf{z}_1, ..., \mathbf{z}_{T_{\mathrm{ex}}}\} \in \mathbb{R}^{T_{\mathrm{ex}} \times C}$ denote exogenous covariate series (e.g., technical indicators, financial factors, news-based sentiment scores, or macroeconomic variables), where $D$ is the dimension of the endogenous series and $C$ is the number of exogenous features. Other symbols will be defined as needed in subsequent sections.

\subsection{Time Series Forecasting Task}
\label{sec:forecasting_task}
Unlike traditional time series forecasting, which often aims to predict the future value of a single series, financial forecasting has two key distinctions. First, directly forecasting asset prices is problematic due to their non-stationary nature and sensitivity to external shocks, so financial models typically predict returns, which are more stable and meaningful for investment decisions. Second, financial forecasting usually involves predicting the future returns of multiple assets simultaneously, reflecting the real-world requirements of portfolio management and quantitative strategies.

\begin{definition}[Time Series Forecasting]
Given the historical prices of $N$ stocks $\mathbf{x}_{1:T} \in \mathbb{R}^{N \times T \times D}$ and multiple exogenous variables $\mathbf{z}_{1:T_{\mathrm{ex}}}$, the goal of time series forecasting in finance is to predict the future $S$-day relative returns for all assets:
\begin{equation}
\hat{\mathbf{R}}_{T+1:T+S} = \mathcal{F}_\theta(\mathbf{x}_{1:T}, \mathbf{z}_{1:T_{\mathrm{ex}}}),
\end{equation}
where $\hat{\mathbf{R}}_{T+1:T+S} \in \mathbb{R}^{N \times S}$ denotes the predicted relative returns (e.g., $\mathbf{R}_{t} = \frac{\mathbf{x}_t}{\mathbf{x}_T} - 1$), and $\mathcal{F}_\theta$ is a forecasting model parameterized by $\theta$.
\end{definition}
    
\subsection{Algorithmic Trading Task}
\label{sec:trading_task}
This task involves the design, simulation, and evaluation of systematic trading strategies across single asset, emphasizing real-time decision making and risk management.

\begin{definition}[Algorithmic Trading]
Given historical observations $\mathbf{x}_{1:T}$ and exogenous variables $\mathbf{z}_{1:T_{\mathrm{ex}}}$, the algorithmic trading problem can be addressed via two mainstream approaches:

\textbf{(a) ML\&DL-based Approach.} The goal is to predict future price returns or movement directions using a predictive model $\mathcal{F}_\theta$:
\begin{equation}
    \hat{y}_{T+1:T+S} = \mathcal{F}_\theta(\mathbf{x}_{1:T}, \mathbf{z}_{1:T_{\mathrm{ex}}}),
\end{equation}
where $\hat{y}_t$ can represent the predicted return, probability of upward/downward movement, or other trading signals. The final trading actions $a_{T+1:T+S}$ are then determined by applying a pre-defined decision rule to $\hat{y}_{T+1:T+S}$ (e.g., buy if $\hat{y}_t > \tau$, sell if $\hat{y}_t < -\tau$, hold otherwise).

\textbf{(b) RL-based Approach.} The trading task is modeled as a Markov decision process (MDP). At each time step $t$, the agent observes state $s_t$ (constructed from $\mathbf{x}{1:t}$ and $\mathbf{z}{1:t}$), chooses action $a_t \in \mathcal{A}$, earns reward $r_t$, and moves to the next state $s_{t+1}$. The goal is to learn a policy $\pi_\theta(a_t \mid s_t)$ that maximizes expected cumulative reward:
\begin{equation}
    \max_\theta~ \mathbb{E}_{\pi_\theta} \left[\sum_{t=T+1}^{T+S} r_t\right].
\end{equation}
\end{definition}
    
\subsection{Portfolio Management Task}
\label{sec:portfolio_task}
This task focuses on the construction, optimization, and dynamic rebalancing of investment portfolios, subject to real-world operational constraints (e.g., transaction costs, position limits), and supports various objective functions and risk measures such as return maximization, volatility minimization, and sharpe ratio optimization.

\begin{definition}[Portfolio Management]
Given historical price sequences $\mathbf{x}_{1:T} \in \mathbb{R}^{N \times T \times D}$ for $N$ assets and exogenous variables $\mathbf{z}_{1:T_{\mathrm{ex}}} \in \mathbb{R}^{N \times T_{\mathrm{ex}} \times C}$, the portfolio management problem can be approached in two mainstream ways:

\textbf{(a) ML\&DL-based Approach.}
The goal is to predict future asset returns or risk estimates using a predictive model $\mathcal{F}_\theta$:
\begin{equation}
    \hat{\mathbf{y}}_{T+1:T+S} = \mathcal{F}_\theta(\mathbf{x}_{1:T}, \mathbf{z}_{1:T_{\mathrm{ex}}}),
\end{equation}
where $\hat{\mathbf{y}}_t$ may represent the predicted return, risk score, or other signals for the $N$ assets. The allocation weights $\mathbf{w}_{T+1:T+S}$ (where $\sum_{i=1}^N w_{t,i} = 1$ and $w_{t,i} \geq 0$ for all $t$) are then determined by applying an optimization procedure or rule to $\hat{\mathbf{y}}_{T+1:T+S}$ (e.g., mean-variance optimization, risk parity, or rule-based allocation).

\textbf{(b) RL-based Approach.}
This task is formulated as a sequential decision process, where at each time $t$, the agent observes a state $s_t$ (e.g., constructed from $\mathbf{x}_{1:t}$ and $\mathbf{z}_{1:t}$), selects allocation weights $\mathbf{w}_t \in \Delta^{N+1}$ (where $\sum_{i=1}^{N+1} w_{t,i} = 1$ and $w_{t,i} \geq 0$, with $w_{t,0}$ representing cash position), receives a reward $r_t$ (e.g., portfolio return or risk-adjusted reward), and transitions to the next state $s_{t+1}$. The objective is to learn a policy $\pi_\theta(\mathbf{w}_t \mid s_t)$ that maximizes expected cumulative utility:
\begin{equation}
    \max_\theta~ \mathbb{E}_{\pi_\theta} \left[\sum_{t=T+1}^{T+S} U(\mathbf{w}_t, \mathbf{x}_t)\right],
\end{equation}
where $U(\cdot)$ represents cumulative portfolio return.
\end{definition}

\subsection{LLMs Applications}
\label{sec:llms_task}
Encompassing two main categories of LLM applications in finance: i) General language understanding tasks, including SFT and RL training for LLMs and LLM agents on financial text analysis, financial QA, and similar language comprehension tasks, and ii) Sequential decision-making tasks, involving RL training for LLMs or LLM agents through real-world environment interactions, such as trading in live market environments.

\begin{definition}[LLMs Applications]
Given financial-domain inputs from multiple modalities, such as unstructured text (e.g., news, reports), structured time series (e.g., open, high, low, close price), images (e.g., Kline charts), and audio or video data (e.g., financial broadcasts), the goal is to train and deploy large language models $\mathcal{M}_\phi$ for two primary application types: general language understanding and sequential decision-making. Formally,
\begin{equation}
    \mathbf{y} = \mathcal{M}_\phi(\mathbf{D}_1, \mathbf{D}_2, ..., \mathbf{D}_K),
\end{equation}
where each $\mathbf{D}_k$ represents an input modality, and $\mathbf{y}$ is the task-specific output in this financial context.
\end{definition}
Depending on the specific downstream task, LLMs can flexibly serve as predictive models within ML\&DL-based pipelines, or act as autonomous agents capable of RL-based decision-making and advanced tool utilization in finance.

\subsection{Reinforcement Learning for LLMs}

Group Relative Policy Optimization (GRPO)~\cite{shao2024deepseekmath} is a policy optimization algorithm designed to efficiently train LLMs in both language and environment-interactive settings. In the context of financial AI, GRPO can be leveraged to endow LLMs with either general financial knowledge and reasoning ability, and specialized trading skills. For example, when applied to financial document analysis or financial question answering, GRPO enables the LLM to align with environment feedback and develop robust understanding of domain-specific content. In these tasks, the LLM is trained to produce high-quality financial responses, with group-level reward normalization providing stable learning signals for fine-tuning.

Moreover, when applied in simulated trading environments, GRPO can be used to train LLMs through direct environment interaction. The LLM receives financial observations (such as price time series, technical indicators, and news), generates actions (e.g., BUY, HOLD, SELL), and is rewarded according to realized trading returns or risk-adjusted performance. This allows the LLM to acquire real trading capabilities, strategy adaptation, and risk management skills beyond static data learning.

GRPO is applied to our LLM training through a two-stage RL paradigm, where the first stage focuses on equip the model’s financial reasoning abilities using financial reasoning datasets, and the second stage immerses the model in real or simulated market environments to develop practical decision-making skills. For more details, please refer to Appendix~\ref{appx:llm_reasoning}.

\section{FinWorld}
\label{sec:finworld}


This section introduces \projectname, a unified, modular platform designed to overcome the limits of current financial-AI frameworks. A clean, layered architecture lets researchers combine traditional ML, DL, RL, LLMs, and LLMs Agents-based methods for mainstream financial task while keeping concerns clearly separated. Integrated dataset management, standardized model APIs, and a scalable training back-end support both academic studies and real-world deployment. Architectural details appear in Appendix \ref{appx:architecture}.

\begin{figure*}
    \centering
    \includegraphics[width=0.90\linewidth]{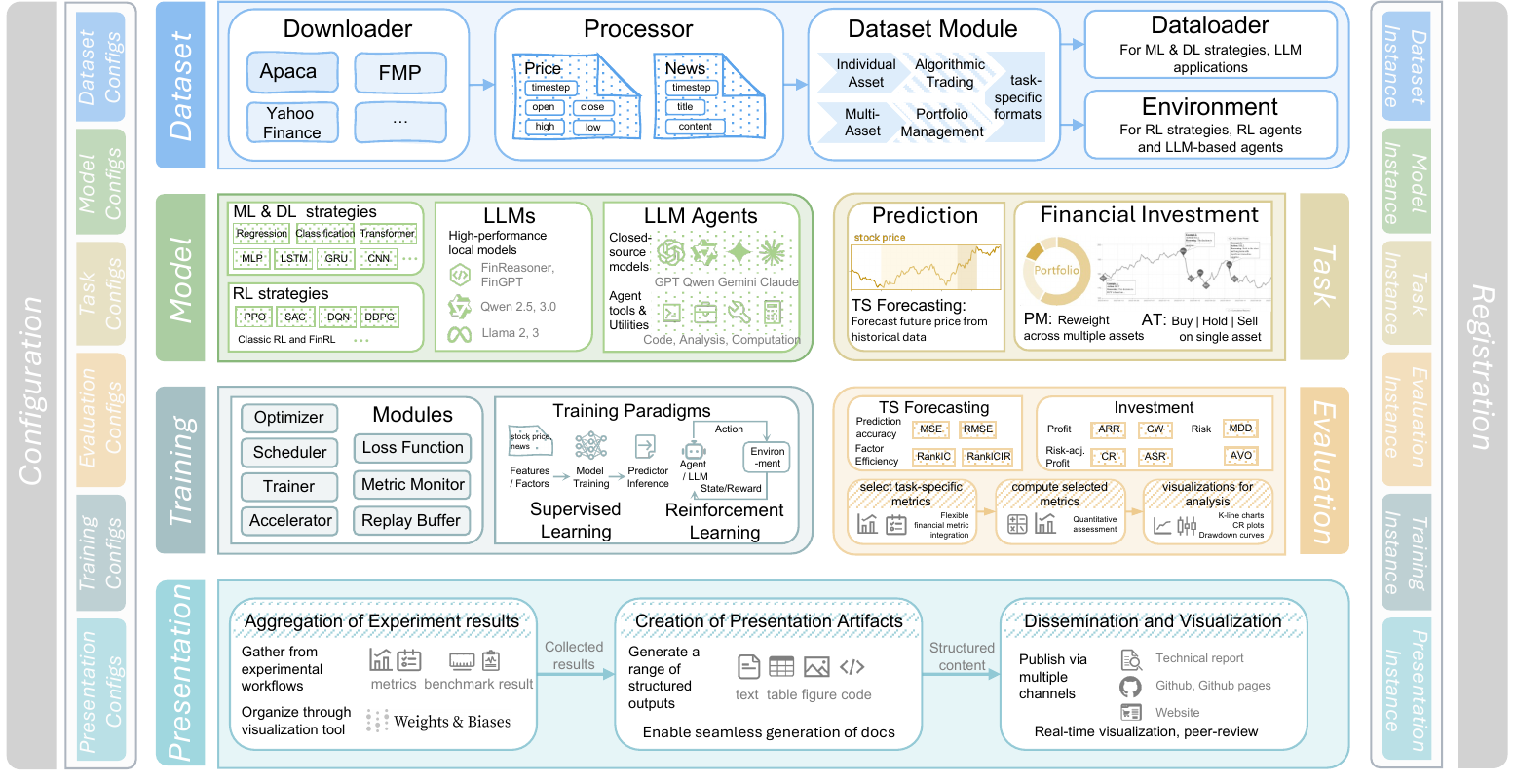}
    \vspace{-1em}
    \caption{Overview of \projectname.}
    \vspace{-1em}
    \label{fig:finworld}
\end{figure*}

\subsection{Design Principles}
The design of \projectname is underpinned by foundational principles, delivering a robust, extensible, research-oriented platform for the development and evaluation of financial AI models and systems:

\begin{itemize}[leftmargin=*,itemsep=0.5em]
    \item \textbf{Layered and Object-Oriented Architecture}. We employ a hierarchical, layered architecture with object-oriented design principles to ensure clear separation of concerns and facilitate both flexibility and scalability across the platform. 
    
          
      
      

    \item \textbf{Modular and Decoupled Design}. \projectname adopts a fully modular architecture. Each component is developed as a self-contained unit with well-defined interfaces, enabling separate optimization and streamlined integration of custom components.

    \item \textbf{Extensibility and Paradigm Fusion}. \projectname emphasizes extensibility at all levels, providing standardized extension points for seamless integration of novel algorithms and datasets. This architecture natively supports the fusion of diverse paradigms across financial AI research.
\end{itemize}

\subsection{Configuration Layer}
The configuration layer of \projectname is built on \texttt{mmengine}~\cite{mmengine2022} and provides a unified and extensible system using dictionaries. It centralizes all experimental settings, including datasets, models, training, and evaluation, in a readable and modular format. Support for configuration inheritance and overrides ensures reproducibility, flexibility, and collaborative development. In the meantime, \projectname uses a registry mechanism to manage core components such as the dataset and the environment. Each class is registered by its type and instantiated from the configuration, allowing for flexible component management and a decoupled system architecture.

\subsection{Dataset Layer}

The dataset layer of \projectname comprises multiple functional modules designed to enable standardized, extensible, and task-oriented data management for financial AI research. This layer abstracts the complexities of diverse data sources and modalities, offering a unified interface for data acquisition, preprocessing, and task-specific organization. Specifically, it consists of five main modules: data downloader module, data processor module, dataset module, dataloader module, and environment module, each supporting flexible customization and extensibility.

\textbf{Downloader Module.} This module consists of two main components: i) Market Data Downloader, which standardizes access to heterogeneous financial market data sources, such as FMP~\cite{fmp2025} and Alpaca~\cite{alpaca2025}, supporting multiple time resolutions (e.g., daily, minute-level) and data types (e.g., OHLCV, news); and ii) LLM Reasoning Dataset Downloader, which provides access to financial reasoning datasets including FinQA, professional certification materials (e.g., ACCA, CFA), and other financial knowledge bases. The unified downloader abstraction streamlines data acquisition from various APIs and databases, ensuring consistency and scalability across research tasks.

\textbf{Processor Module.} Building on the standardized data access, this module enables users to configure essential preprocessing steps, such as factor computation (e.g., alpha158~\cite{yang2020qlib}), feature selection, and normalization, LLM reasoning data processing, and other preprocessing steps according to specific research needs. The processor abstraction facilitates reproducible and customizable data transformations for downstream modeling.

\textbf{Dataset Module.} This module organizes processed data into task-specific formats suitable for various financial applications. For algorithmic trading tasks, it encapsulates data as individual asset datasets; for portfolio management and multi-asset problems, it constructs unified multi-asset datasets. The module seamlessly handles multiple data modalities, including numerical time series, structured financial data, and unstructured textual information, enabling comprehensive modeling capabilities for both traditional financial AI tasks and LLM-based applications.

\textbf{Dataloader Module.} This module provides standardized dataloaders for ML, DL, or LLMs applications, such as time series forecasting. It enables efficient batching and sampling, facilitating seamless integration with other frameworks.

\textbf{Environment Module.} Designed for reinforcement learning paradigms, this module encapsulates data as interactive environments. It supports both conventional RL agents and LLM-based agent interactions, enabling unified experimentation and promoting reproducibility across a variety of agent-based learning tasks.

\subsection{Model Layer}

The model layer of \projectname\ consists of multiple specialized modules that enable unified definition, management, and invocation of diverse modeling paradigms within the platform. This layer abstracts the construction and orchestration of traditional ML models, deep neural network, and LLMs, providing standardized interfaces for consistent training, inference, and integration across multi-tasks. Specifically, it is organized into four main modules:

\textbf{ML Models.} This module defines classical model structures in a unified declarative form. It includes linear and logistic regression, decision trees, random forests, and gradient boosting ensembles (for example, XGBoost and LightGBM). Each specification exposes a consistent input and output schema and a task head (regression, binary classification, or multiclass classification), with structural attributes such as regularization.

\textbf{DL Models.} This module organizes neural architectures into components, layers, and model networks. Components include data embedding for financial time series~\cite{wu2022timesnet}, patch embedding~\cite{dosovitskiy2020image}, position encodings, attention, transformer blocks, and activation functions. Layers compose these into an embedding layer, an encoder, and a decoder with consistent interfaces. Model networks such as Autoformer, VAE, and GPT-style decoders are instantiated by wiring the layers under a common specification, independent of training and inference.

\textbf{RL Models.} This module defines RL network structures with an actor-critic abstraction. Policy and value networks are composed from DL models components, including embedding layers and backbones such as MLP, LSTM/GRU, and Transformer. Specifications cover discrete or continuous action heads and value heads, with shared or separate encoders and optional memory. Financial constraints are represented as structural hooks for transaction costs, slippage and market impact, risk limits, and trading calendars, supporting trading and portfolio tasks.

\textbf{LLMs Models.} This module centralizes access to both proprietary and open‑source LLMs via a unified interface. It supports seamless switching among commercial providers (e.g., GPT‑4.1, Claude‑4‑Sonnet, Gemini‑2.5‑Pro) and efficient local models (e.g., Qwen2.5, Qwen3), with standardized controls for context length, sampling, and cost/latency tracking. Built‑in tools include function calling, tool use, and retrieval‑augmented generation for financial documents. The module integrates naturally with agents and downstream models, enabling document understanding, information extraction, instruction following.

\subsection{Training Layer}

The training layer in \projectname offers a modular scaffold that abstracts every element required to optimise all method pipelines for financial applications. A uniform interface guarantees experiment reproducibility, smooth scaling from single GPU notebooks to distributed clusters, and quick transfer of best practices across tasks such as time series forecasting, trading, portfolio management, and LLM applications.

\textbf{Optimizer}. A rich catalogue of first order methods, ranging from classic stochastic gradient descent (SGD) to adaptive variants such as Adam and AdamW, lets practitioners match optimisation dynamics to task characteristics. A common wrapper normalises hyperparameter signatures and supports gradient centralisation, decoupled weight decay, and mixed precision updates, ensuring robust convergence on noisy and non stationary market data.

\textbf{Loss}. A flexible factory produces objective functions that cover regression losses (mean squared error and mean absolute error), classification losses for event prediction, and reinforcement learning surrogates for policy and value objectives. Composite losses can be declared with a single line, enabling multi-task learning or the joint optimisation of risk adjusted return and prediction accuracy.

\textbf{Scheduler}. Static strategies (e.g., step, cosine, linear) and adaptive schemes (e.g., warm-up then decay) are available for both learning rates and regularisation coefficients. Schedulers are time-aware, resuming the exact trajectory when checkpoints are reloaded.

\textbf{Metrics}. In addition to generic accuracy and error scores, the library ships with finance specific diagnostics such as ARR, SR, and MDD. Each metric is logged at configurable intervals and can trigger early stopping or hyperparameter sweeps, enabling data-driven model selection.

\textbf{Trainer}. Acting as the orchestrator, the trainer pipes data loading, forward and backward passes, gradient clipping, metric evaluation, checkpointing, and experiment logging (TensorBoard or WandB). Specialised variants provide task specific logic, for example the forecasting trainer, trading trainer, portfolio trainer, and large language model trainer. Clear callback hooks let researchers inject custom steps, such as on-the-fly data augmentation or bespoke risk constraints, without changing the core loop.

Together, these components form a coherent architecture that accelerates experimentation, strengthens reproducibility, and lowers the barrier to launching state-of-the-art AI solutions in the demanding environment of financial markets.

\subsection{Evaluation Layer}

The evaluation layer in \projectname provides a comprehensive and extensible framework for assessing financial AI models and strategies. It dynamically selects and applies appropriate evaluation protocols and metrics based on the specific task and model type, supporting both established benchmarks and user-defined criteria. The framework incorporates a diverse library of financial and predictive metrics, such as ARR, MDD, SR, and MSE, which can be flexibly combined or extended as needed. In addition to quantitative assessment, the evaluation layer provides advanced visualization tools, including candlestick (K-line) charts, cumulative return plots, drawdown curves, and trade annotation overlays, to facilitate intuitive interpretation and diagnosis of model performance. This flexible and adaptive evaluation process not only streamlines model assessment for diverse financial tasks, but also facilitates systematic comparison, rapid diagnosis of strengths and weaknesses, and iterative improvement of financial AI methods within the platform. Integrated into the trainer's validation and test stages, it ensures consistency with training-phase evaluation and supports systematic comparison, diagnosis, and iterative improvement.


\subsection{Task Layer}

The task layer is responsible for the systematic definition, abstraction, and encapsulation of financial AI task types. It systematically supports a wide spectrum of financial AI tasks, providing unified abstractions and modular interfaces that facilitate integration with upstream data modules and downstream modeling components. This layer centers on several core financial tasks, including time series forecasting, algorithmic trading, portfolio management, and LLMs applications, which are formally defined in Section~\ref{sec:preliminaries}. 

Each task is specified by configurable input/output schemas and standardized evaluation protocols, supporting both established benchmarks and user-customized scenarios. The unified task architecture enables rapid prototyping, cross-task generalization, and reproducible research across financial AI applications. Additionally, the layer provides a flexible framework for deploying LLM-based agents in asynchronous, single-agent, or multi-agent configurations, allowing users to compose and customize financial agents for diverse research and application needs.

\subsection{Presentation Layer}

The presentation layer in \projectname is designed for automated dissemination and documentation of experimental results. Central to this layer is a dedicated presentation agent that orchestrates the aggregation of evaluation outputs, automatic generation of technical reports, and the creation of interactive web pages for result interpretation and dissemination. Experimental findings, along with key visualizations and benchmark summaries, are systematically compiled into structured documents and published to collaborative platforms such as GitHub, ensuring transparent sharing and long-term accessibility. Furthermore, seamless integration with experiment tracking tools like Wandb enables real-time visualization and comparison of metrics throughout the research lifecycle. This automated, multi-channel presentation workflow enhances reproducibility, supports peer review, and amplifies the visibility and impact of financial AI research conducted within the platform.

\section{Empirical Evaluation}

\subsection{Dataset}

We utilize two representative markets, the US and China, covering four major stock pools: DJ30, SP500, SSE50, and HS300. The dataset spans from 1995-05-01 to 2025-05-01, comprising daily and minute-level \textbf{price} as well as \textbf{news}, totaling over 800 million data points. This extensive dataset supports tasks such as time series forecasting, algorithmic trading, and portfolio management. Additionally, we collect a comprehensive \textbf{LLM Reasoning dataset} in both Chinese and English, covering diverse financial scenarios for training LLMs and agents, with over 80,000 samples across multiple reasoning benchmarks. All experiments are conducted on 2 NVIDIA H100 GPUs. Unless otherwise specified, all reported results are averaged over three runs with different random seeds, and the best results in each column are underlined. Detailed information can be found in Appendix~\ref{appx:dataset}.


\subsection{Time Series Forecasting}

\textbf{Dataset Setup}. We use daily OHLCV and Alpha158 features for DJ30, SP500, SSE50, and HS300 from 2015-05-01 to 2025-05-01, with per-stock normalization and data split at 2023-05-01 for training and validation. Detailed information can be found in Appendix~\ref{appx:forecasting}.

\noindent\textbf{Metrics}. We evaluate forecasting performance using four metrics: \textbf{MAE}, \textbf{MSE}, \textbf{RankIC}, and \textbf{RankICIR}. MAE and MSE focus on absolute prediction accuracy, while RankIC and RankICIR are particularly relevant for evaluating the model's effectiveness in capturing return-based or ranking-based financial relationships.

\noindent\textbf{Methods}. We evaluate several ML-based and DL-based time series forecasting models, including: i) ML-based methods: \textbf{LightGBM}~\cite{yang2020qlib}, \textbf{XGBoost}~\cite{yang2020qlib}; ii) DL-based methods: \textbf{Autoformer}~\cite{wu2021autoformer}, \textbf{Crossformer}~\cite{zhang2023crossformer}, \textbf{ETSformer}~\cite{woo2022etsformer}, \textbf{DLinear}~\cite{zeng2023transformers}, \textbf{TimesNet}~\cite{wu2022timesnet}, \textbf{PatchTST}~\cite{nie2022time}, \textbf{TimeMixer}~\cite{wang2024timemixer}, and \textbf{TimeXer}~\cite{wang2024timexer}. These models are selected based on their effectiveness in capturing complex patterns and their ability to handle high-dimensional data.

\noindent\textbf{Experiment Results}
As shown in the results Table~\ref{tab:forecasting}, on the DJ30 dataset, the TimeXer model achieves a MAE of 0.0529 and an MSE of 0.0062, significantly lower than LightGBM (MAE 0.1392, MSE 0.0235). TimeXer also attains the highest RankICIR of 0.4889, compared to 0.2017 for LightGBM. Similarly, on the HS300 dataset, models such as TimeMixer and TimeXer outperform ML methods, with MAEs of 0.3804 and 0.3727, and MSEs of 0.0442 and 0.0529, respectively. Overall, deep learning models achieve lower prediction errors and higher rank-based metrics than machine learning methods, highlighting the advantages of DL approaches for financial time series forecasting.

\begin{table}[htbp]
  \caption{Comparison of models for time series forecasting.}
  \label{tab:forecasting}
  \centering
  \footnotesize
  \setlength{\tabcolsep}{2pt}
  \renewcommand{\arraystretch}{0.8}
  \resizebox{0.98\linewidth}{!}{
  \begin{tabular}{lcccccccccccccccc}
  \toprule
  \multirow{3}{*}{\parbox{1.2cm}{\centering\textbf{Model}}} &
  \multicolumn{4}{c}{DJ30} &
  \multicolumn{4}{c}{SP500} \\
  \cmidrule(lr){2-5} \cmidrule(lr){6-9}
  & MAE~$\downarrow$ & MSE~$\downarrow$ & RankIC~$\uparrow$ & RankICIR~$\uparrow$ 
  & MAE~$\downarrow$ & MSE~$\downarrow$ & RankIC~$\uparrow$ & RankICIR~$\uparrow$ \\
  \midrule
  
  \rowcolor{gray!15}
  \multicolumn{9}{c}{\textit{\textbf{ML-based}}} \\
  LightGBM  & 0.1392 & 0.0235 & 0.0157 & 0.2017 & 0.6741 & 0.0788 & 0.0216 & 0.2242 \\
  XGBoost   & 0.1476 & 0.0251 & 0.0138 & 0.1684 & 0.7025 & 0.0814 & 0.0199 & 0.1925 \\
  \rowcolor{gray!15}
  \multicolumn{9}{c}{\textit{\textbf{DL-based}}} \\
  Autoformer   & 0.1044 & 0.0182 & 0.0193 & 0.3265 & 0.4568 & 0.0671 & 0.0312 & \underline{0.3379} \\
  Crossformer  & 0.1624 & 0.0173 & 0.0150 & 0.3561 & 0.6448 & 0.0774 & 0.0219 & 0.2984 \\
  ETSformer    & 0.0640 & 0.0115 & 0.0209 & 0.2411 & 0.2560 & 0.0448 & 0.0295 & 0.3125 \\
  DLinear      & 0.0867 & 0.0122 & 0.0144 & 0.2247 & 0.3537 & 0.0416 & 0.0177 & 0.2783 \\
  TimesNet     & 0.1242 & 0.0142 & 0.0211 & 0.2642 & 0.4013 & 0.0660 & \underline{0.0483} & 0.3512 \\
  PatchTST     & 0.0830 & 0.0112 & 0.0140 & 0.2134 & 0.3682 & 0.0498 & 0.0157 & 0.1922 \\
  TimeMixer    & 0.0620 & 0.0081 & 0.0294 & 0.4741 & 0.2278 & 0.0279 & 0.0349 & 0.1193 \\
  TimeXer      & \underline{0.0529} & \underline{0.0062} & \underline{0.0302} & \underline{0.4889} & \underline{0.1871} & \underline{0.0240} & 0.0425 & 0.3255 \\
  \bottomrule
  \toprule
  \multirow{3}{*}{\parbox{1.2cm}{\centering\textbf{Model}}} &
  \multicolumn{4}{c}{SSE50} &
  \multicolumn{4}{c}{HS300} \\
  \cmidrule(lr){2-5} \cmidrule(lr){6-9}
  & MAE~$\downarrow$ & MSE~$\downarrow$ & RankIC~$\uparrow$ & RankICIR~$\uparrow$ 
  & MAE~$\downarrow$ & MSE~$\downarrow$ & RankIC~$\uparrow$ & RankICIR~$\uparrow$ \\
  \midrule
  
  \rowcolor{gray!15}
  \multicolumn{9}{c}{\textit{\textbf{ML-based}}} \\
  LightGBM  & 0.1540 & 0.0267 & 0.0211 & 0.1688 & 0.7836 & 0.1097 & 0.0219 & 0.1961 \\
  XGBoost   & 0.1632 & 0.0293 & 0.0184 & 0.1497 & 0.8140 & 0.1123 & 0.0193 & 0.1724 \\
  \rowcolor{gray!15}
  \multicolumn{9}{c}{\textit{\textbf{DL-based}}} \\
  Autoformer   & 0.1163 & 0.0212 & 0.0336 & 0.1848 & 0.5696 & 0.1213 & \underline{0.0364} & \underline{0.2493} \\
  Crossformer  & 0.1756 & 0.0197 & 0.0225 & 0.1318 & 0.9829 & 0.1082 & 0.0187 & 0.1525 \\
  ETSformer    & 0.0737 & 0.0125 & \underline{0.0452} & 0.2107 & 0.4413 & 0.0546 & 0.0332 & 0.2406 \\
  DLinear      & 0.0932 & 0.0154 & 0.0343 & 0.1722 & 0.4315 & 0.0717 & 0.0285 & 0.2078 \\
  TimesNet     & 0.1329 & 0.0168 & 0.0251 & 0.1465 & 0.7118 & 0.0997 & 0.0241 & 0.1894 \\
  PatchTST     & 0.0907 & 0.0135 & 0.0377 & 0.1934 & \underline{0.3253} & 0.0595 & 0.0276 & 0.1971 \\
  TimeMixer    & 0.0724 & 0.0119 & 0.0402 & 0.2086 & 0.3804 & 0.0442 & 0.0348 & 0.2417 \\
  TimeXer      & \underline{0.0625} & \underline{0.0093} & 0.0425 & \underline{0.2173} & 0.3727 & \underline{0.0529} & 0.0352 & 0.2478 \\

  \bottomrule
  \end{tabular}
  }
  \end{table}
\begin{table}[htb]
\caption{Comparison of models for algorithmic trading.}
\vspace{-1em}
\label{tab:trading}
\centering
\footnotesize
\setlength{\tabcolsep}{1pt}
\renewcommand{\arraystretch}{0.85}
\resizebox{0.97\linewidth}{!}{
\begin{tabular}{lcccccccccccc}
\toprule
\multirow{2}{*}{\parbox{1.2cm}{\centering\textbf{Model}}}
    & \multicolumn{6}{c}{AAPL}
    & \multicolumn{6}{c}{AMZN} \\

\cmidrule(lr){2-7} \cmidrule(lr){8-13}
& ARR~$\uparrow$ & SR~$\uparrow$ & MDD~$\downarrow$ & CR~$\uparrow$ & SoR~$\uparrow$ & VOL~$\downarrow$
& ARR~$\uparrow$ & SR~$\uparrow$ & MDD~$\downarrow$ & CR~$\uparrow$ & SoR~$\uparrow$ & VOL~$\downarrow$ \\
\midrule

\rowcolor{gray!15}
\multicolumn{13}{c}{\textit{\textbf{Rule-based}}} \\
BUY\&HOLD & 13.0620 & 0.6654 & 17.6828 & 0.7387 & 0.9606 & 0.2210 & 43.9823 & 1.4190 & 19.6246 & 2.2412 & 2.1558 & 0.2854 \\
MACD & 9.7042 & 0.6964 & \underline{11.5621} & 0.8393 & 1.0251 & 0.1487 & -1.4856 & 0.0028 & \underline{19.4850} & -0.0762 & 0.0041 & 0.1761 \\

\rowcolor{gray!15}
\multicolumn{13}{c}{\textit{\textbf{ML-based}}} \\
LightGBM  & 7.4450 & 0.5112 & 16.7097 & 0.4456 & 0.7363 & 0.1679 & -2.6266 & -0.1886 & 38.1339 & -0.0689 & -0.2575 & \underline{0.1093} \\
XGBoost  & 15.7943 & 1.0359 & 17.6320 & 2.0695 & \underline{2.0720} & \underline{0.1185} & 20.8648 & 0.7175 & 26.5675 & 0.3134 & 0.8854 & 0.3823 \\

\rowcolor{gray!15}
\multicolumn{13}{c}{\textit{\textbf{DL-based}}} \\
Transformer   & 23.3469 & 0.8788 & 18.7073 & 1.2480 & 1.6204 & 0.2809 & 5.8822 & 0.3716 & 25.5574 & 0.0884 & 0.4436 & 0.3520 \\
LSTM   & 11.2387 & 0.5262 & 35.9190 & 0.2448 & 0.9589 & 0.2647 & 15.5243 & 0.8552 & 40.3636 & 0.3846 & 1.2941 & 0.1897 \\
DLinear   & 9.5539 & 0.4686 & 34.7311 & 0.2136 & 0.8594 & 0.2644 & 9.9999 & 0.7221 & 30.5673 & 0.3272 & 1.1297 & 0.1468 \\

\rowcolor{gray!15}
\multicolumn{13}{c}{\textit{\textbf{RL-based}}} \\
PPO & \underline{31.5838} & \underline{1.0483} & 18.7102 & \underline{1.6222} & 1.9040 & 0.2956 & 37.3429 & 1.2514 & 22.2379 & \underline{1.6793} & 1.7909 & 0.2864 \\
SAC & 28.2963 & 0.8758 & 37.7384 & 0.6467 & 1.5797 & 0.2970 & \underline{44.4502} & \underline{1.6389} & 23.4218 & 1.4709 & \underline{2.6872} & 0.1918 \\

\bottomrule
\toprule

\multirow{2}{*}{\parbox{1.2cm}{\centering\textbf{Model}}}
    & \multicolumn{6}{c}{GOOGL}
    & \multicolumn{6}{c}{META} \\
    
\cmidrule(lr){2-7} \cmidrule(lr){8-13}
& ARR~$\uparrow$ & SR~$\uparrow$ & MDD~$\downarrow$ & CR~$\uparrow$ & SoR~$\uparrow$ & VOL~$\downarrow$
& ARR~$\uparrow$ & SR~$\uparrow$ & MDD~$\downarrow$ & CR~$\uparrow$ & SoR~$\uparrow$ & VOL~$\downarrow$ \\
\midrule

\rowcolor{gray!15}
\multicolumn{13}{c}{\textit{\textbf{Rule-based}}} \\
BUY\&HOLD & 28.1837 & 1.0236 & 22.2309 & 1.1678 & 1.4589 & 0.2811 & 67.2719 & 1.7264 & 18.8226 & 1.5740 & 2.8285 & 0.3286 \\
MACD & 10.0432 & 0.6068 & \underline{12.6638} & 0.7931 & 0.8620 & 0.1862 & 40.3374 & 1.5468 & \underline{12.7635} & 1.1604 & 2.0533 & 0.2362 \\

\rowcolor{gray!15}
\multicolumn{13}{c}{\textit{\textbf{ML-based}}} \\
LightGBM  & 16.4245 & 1.0908 & 21.0885 & 0.7788 & 1.6732 & 0.1468 & 39.6612 & 1.1212 & 23.4416 & 0.7655 & 1.5979 & 0.3676 \\
XGBoost  & 7.5362 & 0.6783 & 20.6611 & 0.3648 & 1.2239 & \underline{0.1171} & 32.0035 & 1.6461 & 24.2472 & 1.3199 & 2.8858 & \underline{0.1778} \\

\rowcolor{gray!15}
\multicolumn{13}{c}{\textit{\textbf{DL-based}}} \\
Transformer   & 10.3923 & 0.6683 & 36.7980 & 0.2824 & 0.9652 & 0.1468 & 48.2834 & 1.6819 & 29.0003 & \underline{1.6649} & \underline{2.9686} & 0.2528 \\
LSTM   & 18.7416 & 0.8849 & 36.8169 & 0.5091 & 1.2871 & 0.2221 & 44.6391 & 1.1864 & 30.0527 & 0.8918 & 1.7019 & 0.3702 \\
DLinear   & 21.4708 & 1.0461 & 36.8203 & 0.5831 & 1.5575 & 0.2063 & 35.3426 & 1.445 & 34.0928 & 1.0366 & 2.6560 & 0.2267 \\

\rowcolor{gray!15}
\multicolumn{13}{c}{\textit{\textbf{RL-based}}} \\
PPO & 20.5532 & 1.0145 & 16.7733 & 1.2254 & 1.5855 & 0.2048 & \underline{72.0104} & \underline{2.1031} & 18.6421 & 1.5360 & 1.4293 & 0.2591 \\
SAC & \underline{30.6081} & \underline{1.2203} & 35.4036 & \underline{1.7233} & \underline{1.8833} & 0.2038 & 56.8891 & 1.5085 & 34.6565 & 1.6415 & 2.4020 & 0.3355 \\

\bottomrule
\toprule

\multirow{2}{*}{\parbox{1.2cm}{\centering\textbf{Model}}}
    & \multicolumn{6}{c}{MSFT}
    & \multicolumn{6}{c}{TSLA} \\

\cmidrule(lr){2-7} \cmidrule(lr){8-13}
& ARR~$\uparrow$ & SR~$\uparrow$ & MDD~$\downarrow$ & CR~$\uparrow$ & SoR~$\uparrow$ & VOL~$\downarrow$
& ARR~$\uparrow$ & SR~$\uparrow$ & MDD~$\downarrow$ & CR~$\uparrow$ & SoR~$\uparrow$ & VOL~$\downarrow$ \\
\midrule

\rowcolor{gray!15}
\multicolumn{13}{c}{\textit{\textbf{Rule-based}}} \\
BUY\&HOLD & 13.3812 & 0.6991 & 18.6840 & 0.7162 & 0.9635 & 0.2116 & 30.8051 & 0.7469 & 53.6797 & 0.5739 & 1.1537 & 0.5892 \\
MACD & 6.4616 & 0.4703 & \underline{13.7325} & 0.4705 & 0.6337 & \underline{0.1606} & 68.8393 & 1.5085 & \underline{21.7414} & \underline{2.1663} & 2.5809 & 0.3989 \\

\rowcolor{gray!15}
\multicolumn{13}{c}{\textit{\textbf{ML-based}}} \\
LightGBM  & 20.7821 & 0.9031 & 37.8864 & 0.5485 & 1.4983 & 0.2400 & 49.2144 & 1.3086 & 56.0589 & 0.8779 & 2.2514 & \underline{0.3529} \\
XGBoost  & 14.4887 & 0.6633 & 49.8944 & 0.2904 & 1.0676 & 0.2497 &  39.7010 & 0.9788 & 58.5673 & 0.5053 & 1.5409 & 0.4401 \\

\rowcolor{gray!15}
\multicolumn{13}{c}{\textit{\textbf{DL-based}}} \\
Transformer   & 23.2662 & 1.0207 & 32.8318 & 0.6037 & 1.4478 & 0.2180 & 73.1291 & 1.2581 & 48.5669 & 0.9308 & 1.8457 & 0.5683 \\
LSTM   & 16.5063 & 0.6185 & 31.8587 & 0.3910 & 1.0206 & 0.2162 & 44.7985 & 1.0768 & 53.6992 & 1.0904 & 1.5650 & 0.6473 \\
DLinear   & 9.4778 & 0.5599 & 31.7271 & 0.2987 & 0.7792 & 0.1962 & 41.1897 & 1.0612 & 62.3380 & 0.6608 & 1.7030 & 0.3998 \\

\rowcolor{gray!15}
\multicolumn{13}{c}{\textit{\textbf{RL-based}}} \\
PPO       & \underline{29.5326} & \underline{1.1044} & 18.6900 & \underline{1.5801} & \underline{1.7624} & 0.2654 & 58.5511 & 1.0402 & 48.5536 & 0.5703 & 1.6841 & 0.4298 \\
SAC       & 22.2784 & 0.9515 & 27.7286 & 0.8035 & 1.5859 & 0.2405 & \underline{70.548} & \underline{1.6474} & 57.4088 & 1.5065 & \underline{2.7099} & 0.5016 \\

\bottomrule
\end{tabular}
}
\vspace{-2em}
\end{table}

\subsection{Algorithmic Trading}
\textbf{Dataset Setup}. We use daily OHLCV and Alpha158 features for six US stocks, AAPL, AMZN, GOOGL, META, MSFT, and TSLA, from 1995-05-01 to 2025-05-01, with per-stock normalization and data split at 2023-05-01 for training and validation. Detailed information can be found in Appendix~\ref{appx:trading}.

\noindent\textbf{Metrics}. We evaluate trading performance using 6 metrics: \textbf{ARR}, \textbf{MDD}, \textbf{VOL}, \textbf{CR}, \textbf{SR}, and \textbf{SoR}. ARR measures profitability. MDD captures the worst drawdown, and VOL reflects return variability, both indicating risk exposure. CR evaluates returns relative to drawdown risk, and SR and SoR assess risk-adjusted returns based on total and downside volatility, respectively. Together, these metrics provide a comprehensive view of profitability and risk.

\noindent\textbf{Methods}. We systematically evaluate several Rule-based, ML-based, DL-based, and RL-based methods, including i) Rule-based methods: \textbf{BUY\&HOLD}, \textbf{MACD}; ii) ML-based Methods: \textbf{LightGBM}, \textbf{XGBoost}; iii) DL-based Methods: \textbf{Transformer}~\cite{vaswani2017attention}, \textbf{LSTM}, \textbf{DLinear}~\cite{zeng2023transformers};  iv) RL-based Methods: \textbf{PPO}~\cite{schulman2017proximal}, \textbf{SAC}~\cite{haarnoja2018soft}.

\noindent\textbf{Experiment Results}. Table~\ref{tab:trading} shows that RL methods achieve the best overall trading performance, with SAC and PPO delivering the highest annualized returns and SR across most stocks (e.g., SAC achieves 101.55\% ARR on TSLA and PPO attains 2.10 SR on META). DL models like Transformer also outperform ML-based and rule-based methods in both return and risk metrics (e.g., 73.13\% ARR on MSFT). Compared to ML, DL and rule-based strategies, RL methods show clear advantages in both returns and risk-adjusted performance.

\subsection{Portfolio Management}

\textbf{Dataset Setup}. Dataset setup is the same as in the time series forecasting task. Detailed information can be found in 
Appendix~\ref{appx:portfolio}.

\noindent\textbf{Metrics}. The evaluation metrics for portfolio management are consistent with those used in the algorithmic trading task. Specifically, we report \textbf{ARR}, \textbf{SR}, \textbf{MDD}, \textbf{CR}, \textbf{SoR}, and \textbf{VOL}, following the definitions and formulas described in the previous section.

\noindent\textbf{Methods}. We evaluate ML-based, DL-based, and RL-based methods. Since we can apply the Top-$k$ Dropout Strategy ~\cite{yang2020qlib} after any ML\&DL-based forecasting model to construct a portfolio, our ML\&DL-based methods include all ML\&DL-based forecasting models. Specifically, we consider: i) Rule-based methods: \textbf{BUY\&HOLD}; ii) ML-based Methods: \textbf{LightGBM}, \textbf{XGBoost}; iii) DL-based Methods: \textbf{Autoformer}~\cite{wu2021autoformer}, \textbf{Crossformer}~\cite{zhang2023crossformer}, \textbf{ETSformer}~\cite{woo2022etsformer}, \textbf{DLinear}~\cite{zeng2023transformers}, \textbf{TimesNet}~\cite{wu2022timesnet}, \textbf{PatchTST}~\cite{nie2022time}, \textbf{TimeMixer}~\cite{wang2024timemixer}, and \textbf{TimeXer}~\cite{wang2024timexer}; iv) RL-based Methods: \textbf{PPO}~\cite{schulman2017proximal}, \textbf{SAC}~\cite{haarnoja2018soft}.

\noindent\textbf{Experiment Results}
Overall, the RL-based methods, especially SAC, achieve the best results on all benchmarks, with annualized returns up to 31.2\% (SP500) and Sharpe ratios above 1.5. In contrast, rule-based and ML-based methods show much lower returns (e.g., Buy\&Hold: 9.4\% on DJ30), and DL-based methods generally perform between the two. Notably, SAC consistently delivers higher returns and better risk-adjusted metrics, demonstrating clear advantages for portfolio management.

\begin{table}[htb]
\caption{Comparison of methods for portfolio management.}
\vspace{-0.1cm}
\centering
\footnotesize
\setlength{\tabcolsep}{1pt}
\renewcommand{\arraystretch}{0.85}
\resizebox{0.97\linewidth}{!}{
\begin{tabular}{lcccccccccccc}
\toprule
\multirow{2}{*}{\parbox{1.2cm}{\centering\textbf{Model}}}
    & \multicolumn{6}{c}{DJ30}
    & \multicolumn{6}{c}{SP500} \\

\cmidrule(lr){2-7} \cmidrule(lr){8-13}
& ARR~$\uparrow$ & SR~$\uparrow$ & MDD~$\downarrow$ & CR~$\uparrow$ & SoR~$\uparrow$ & VOL~$\downarrow$
& ARR~$\uparrow$ & SR~$\uparrow$ & MDD~$\downarrow$ & CR~$\uparrow$ & SoR~$\uparrow$ & VOL~$\downarrow$ \\
\midrule

\rowcolor{gray!15}
\multicolumn{13}{c}{\textit{\textbf{Rule-based}}} \\
BUY\&HOLD & 9.4049 & 0.8792 & 9.2585 & 1.0158 & 1.2779 & 0.1089 & 8.9875 & 0.7617 & \underline{13.7434} & 0.6539 & 1.1161 & \underline{0.1228} \\

\rowcolor{gray!15}
\multicolumn{13}{c}{\textit{\textbf{ML-based}}} \\
LightGBM  & 16.3302 & 0.5758 & 30.7444 & 1.1107 & 1.7603 & 0.5162 & 13.4204 & 0.5916 & 30.7037 & 0.7206 & 0.9005 & 0.4149 \\
XGBoost   & 10.6903 & 0.3776 & 30.4737 & 1.6504 & 1.3008 & 0.5167 & 10.2609 & 0.3437 & 30.5994 & 0.8202 & 0.7601 & 0.4135 \\

\rowcolor{gray!15}
\multicolumn{13}{c}{\textit{\textbf{DL-based}}} \\
Autoformer   & -4.1781 & 0.3611 & 34.2251 & -0.0650 & 0.5267 & 0.1123 & 9.7203 & 0.3724 & 33.5112 & 0.4257 & 0.5913 & 0.1563 \\
Crossformer  & 9.6460 & 0.4474 & 33.8498 & 0.1840 & 0.6459 & 0.2135 & 8.2108 & 0.3879 & 29.9936 & 0.5341 & 0.5928 & 0.1289 \\
ETSformer    & 1.5512 & 0.3662 & 31.5739 & 0.0300 & 0.5085 & \underline{0.0918} & 12.0507 & 0.4951 & 30.7685 & 0.7532 & 0.6527 & 0.2342 \\
DLinear      & 17.4357 & 0.7461 & 35.2342 & 0.5593 & 1.0521 & 0.1787 & 17.1605 & 0.7827 & 32.9918 & 1.1329 & 1.1022 & 0.1934 \\
TimesNet     & 16.7235 & 0.8123 & 29.4762 & 0.9821 & 1.0844 & 0.2129 & 12.9701 & 0.6474 & 29.8908 & 0.9891 & 0.9136 & 0.1662 \\
PatchTST     & 13.5049 & 0.7344 & 31.0133 & 0.8961 & 1.0317 & 0.1036 & 9.2306 & 0.4719 & 31.6127 & 0.7654 & 0.8338 & 0.0904 \\
TimeMixer    & 9.1872  & 0.5795 & 30.9910 & 0.6928 & 0.8996 & 0.1861 & 14.4202 & 0.7523 & 30.4321 & 1.0876 & 1.0284 & 0.2177 \\
TimeXer      & 7.4083  & 0.4567 & 32.6578 & 0.5129 & 0.8103 & 0.3184 & 19.8309 & 0.8692 & 28.3254 & 1.2768 & 1.1825 & 0.2841 \\

\rowcolor{gray!15}
\multicolumn{13}{c}{\textit{\textbf{RL-based}}} \\
PPO       & 11.7241 & 1.0339 & 20.1129 & 1.2865 & 1.5246 & 0.2133 & 28.5408 & 1.1136 &  29.5874 & 1.3761 & 1.5589 & 0.3146 \\
SAC       & \underline{19.0272} & \underline{1.5967} & \underline{20.0846} & \underline{2.0945} & \underline{2.4453} & 0.2117 & \underline{31.2104} & \underline{1.5987} & 30.3381 & \underline{2.1362} & \underline{2.4625} & 0.2338 \\

\bottomrule
\toprule

\multirow{2}{*}{\parbox{1.2cm}{\centering\textbf{Model}}}
& \multicolumn{6}{c}{SSE50}
& \multicolumn{6}{c}{HS300} \\

\cmidrule(lr){2-7} \cmidrule(lr){8-13}
& ARR~$\uparrow$ & SR~$\uparrow$ & MDD~$\downarrow$ & CR~$\uparrow$ & SoR~$\uparrow$ & VOL~$\downarrow$
& ARR~$\uparrow$ & SR~$\uparrow$ & MDD~$\downarrow$ & CR~$\uparrow$ & SoR~$\uparrow$ & VOL~$\downarrow$ \\
\midrule

\rowcolor{gray!15}
\multicolumn{13}{c}{\textit{\textbf{Rule-based}}} \\
BUY\&HOLD & 2.8799 & 0.2512 & 0.7742 & 0.1420 & 0.3868 & 0.1706 & 2.7940 & 0.2448 & 19.7412 & 0.1415 & 0.3819 & 0.1738 \\

\rowcolor{gray!15}
\multicolumn{13}{c}{\textit{\textbf{ML-based}}} \\
LightGBM  & 13.4202 & 0.5915 & 0.7032 & 0.5973 & 0.9632 & 0.1472 & 26.5361 & 0.6509 & 0.9407 & 1.2177 & \underline{1.4964} & 0.1693 \\
XGBoost   & 14.3904 & 0.3439 & \underline{0.5991} & 0.5945 & 0.8581 & 0.1358 & 17.2650 & 0.3718 & \underline{0.5823} & 0.5876 & 0.9741 & 0.2116 \\

\rowcolor{gray!15}
\multicolumn{13}{c}{\textit{\textbf{DL-based}}} \\
Autoformer   & 9.4209  & 0.5162 & 0.8093 & 0.8494 & 0.9862 & 0.1586 & 18.7107 & 0.7968 & 1.1079 & 1.2124 & 1.1872 & 0.2193 \\
Crossformer  & 10.5902 & 0.5731 & 0.9107 & 0.9798 & 1.1162 & 0.1741 & 21.8814 & 0.8283 & 1.1523 & \underline{1.2981} & 1.2357 & 0.1642 \\
ETSformer    & 12.1572 & 0.6703 & 0.9755 & 1.0621 & 1.1603 & 0.1814 & 15.2038 & 0.7411 & 1.0635 & 1.0816 & 1.1341 & 0.2328 \\
DLinear      & 17.7206 & 0.8581 & 1.0779 & 1.1941 & 1.2786 & 0.1913 & 29.9904 & 0.8931 & 1.1073 & 1.2425 & 1.2456 & 0.2872 \\
TimesNet     & 14.4103 & 0.8382 & 1.0473 & 1.1495 & 1.2639 & 0.1468 & 24.3309 & 0.9207 & 1.1594 & 1.2930 & 1.2440 & 0.2933 \\
PatchTST     & 15.1308 & 0.9072 & 0.9750 & 1.2366 & 1.2821 & 0.1924 & 22.3607 & 0.9153 & 1.1333 & 1.2645 & 1.1875 & 0.2067 \\
TimeMixer    & 12.8201 & 0.8894 & 0.9456 & 1.2902 & 1.3488 & 0.2031 & 19.1205 & 0.9172 & 1.0581 & 1.1426 & 1.2396 & 0.2486 \\
TimeXer      & 11.4708 & 0.8781 & 0.8942 & 1.3179 & 1.4239 & \underline{0.1172} & 18.1713 & 0.9137 & 1.0503 & 1.1104 & 1.2938 & 0.2173 \\

\rowcolor{gray!15}
\multicolumn{13}{c}{\textit{\textbf{RL-based}}} \\
PPO       & 15.1502 & 1.0013 & 0.6331 & 1.3956 & 1.4481 & 0.1835 & 22.2032 & 1.0578 & 0.8171 & 1.1763 & 1.2541 & \underline{0.1568} \\
SAC       & \underline{20.0670} & \underline{1.0625} & 0.7132 & \underline{1.5320} & \underline{1.5711} & 0.1497 & \underline{39.9005} & \underline{1.1606} & 0.8205 & 1.2032 & 1.3703 & 0.1686 \\
\bottomrule
\end{tabular}
\vspace{-0.3cm}
}
\end{table}

\subsection{LLMs Applications}

\textbf{Dataset Setup}. For comprehensive evaluation, we assess LLMs and LLM Agents on a suite of established financial reasoning benchmarks, including FinQA, FinEval, ConvFinQA, and CFLUE, which cover a broad range of business scenarios. For trading evaluation, we follow the same experimental setup as in algorithmic trading task and use daily-level OHLCV data together with news for the 6 U.S. technology stocks (AAPL, AMZN, GOOGL, META, MSFT, and TSLA) over the period from 2015-05-01 to 2025-05-01.

\noindent\textbf{Metrics}. For financial reasoning task, we evaluate performance using \textbf{Score}. For trading abilities, we use the same six financial metrics as in the above trading task: \textbf{ARR}, \textbf{SR}, \textbf{MDD}, \textbf{CR}, \textbf{SoR}, and \textbf{VOL}.

\noindent\textbf{Methods}. For financial reasoning, we evaluate our proposed models \texttt{FinReasoner} against open-source LLMs on the test dataset, including \textbf{DeepSeek-R1}~\cite{guo2025deepseek}, \textbf{Qwen3-8B}~\cite{yang2025qwen3}, \textbf{Fin-R1-7B}~\cite{guo2025deepseek}, and \textbf{Qwen2.5-7B-Instruct}~\cite{qwen2025qwen25technicalreport}. For trading abilities, we evaluate our own proposed \texttt{FinReasoner} with \textbf{DeepSeek-R1}~\cite{guo2025deepseek}, \textbf{Qwen3-8B}~\cite{yang2025qwen3}, \textbf{Fin-R1-7B}~\cite{guo2025deepseek}, \textbf{Qwen2.5-7B-Instruct}~\cite{qwen2025qwen25technicalreport}, \textbf{GPT-4.1}, \textbf{Claude-4-Sonnet}, all serving as the backbone LLMs within FinAgent framework.

\noindent\textbf{Implementation Details}. LLM training is performed on 16 NVIDIA A100 GPUs, while both LLMs inference and LLMs Agents experiments are conducted on 2 NVIDIA H100 GPUs, with evaluation results reported from a single run.

\begin{figure}
    \centering
    \includegraphics[width=0.9\linewidth]{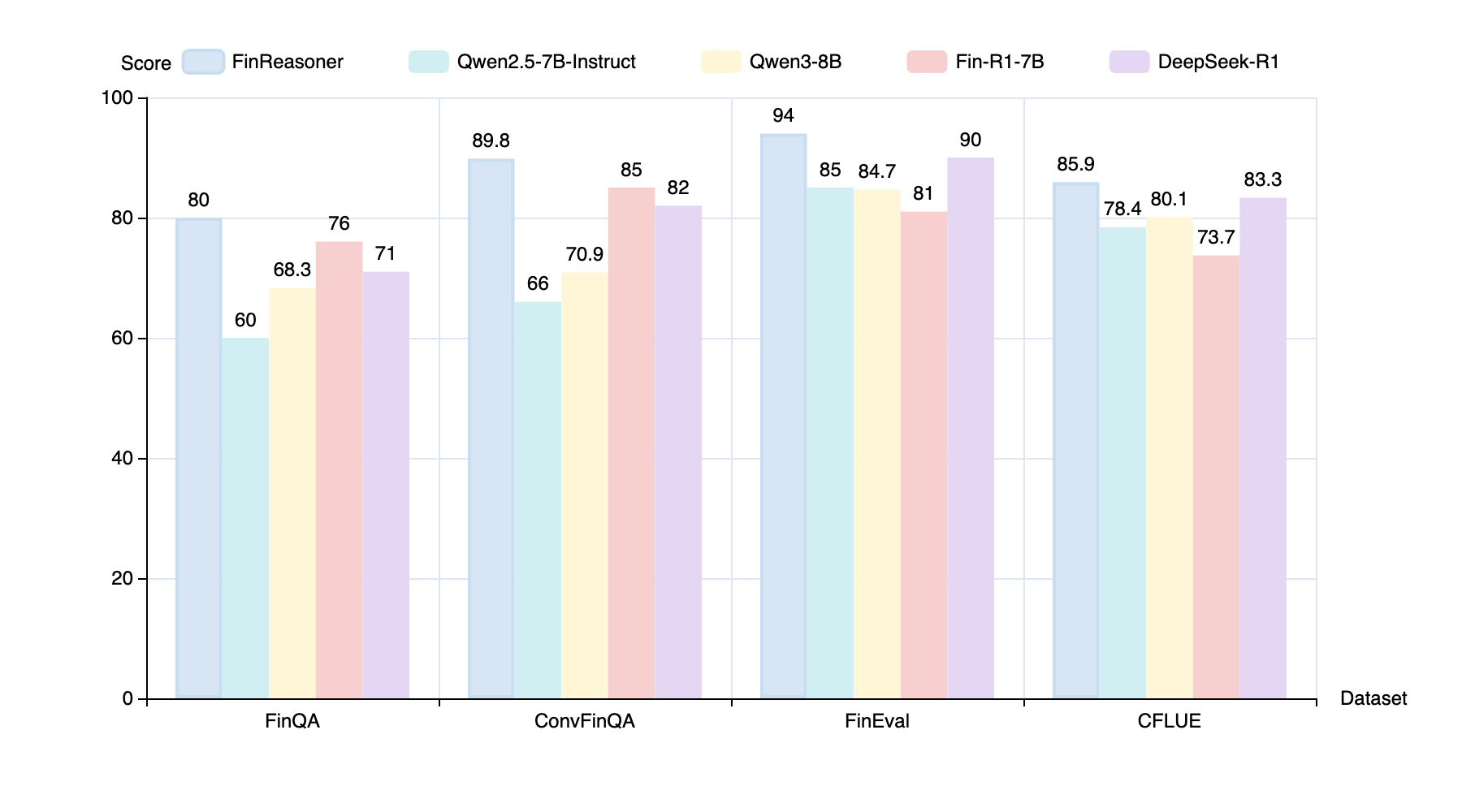}
    \vspace{-1em}
    \caption{Comparison of LLMs on financial reasoning.}
    \label{fig:llm_reasoning}
\end{figure}

\begin{figure}
    \centering
    \vspace{-1em}
    \includegraphics[width=0.8\linewidth]{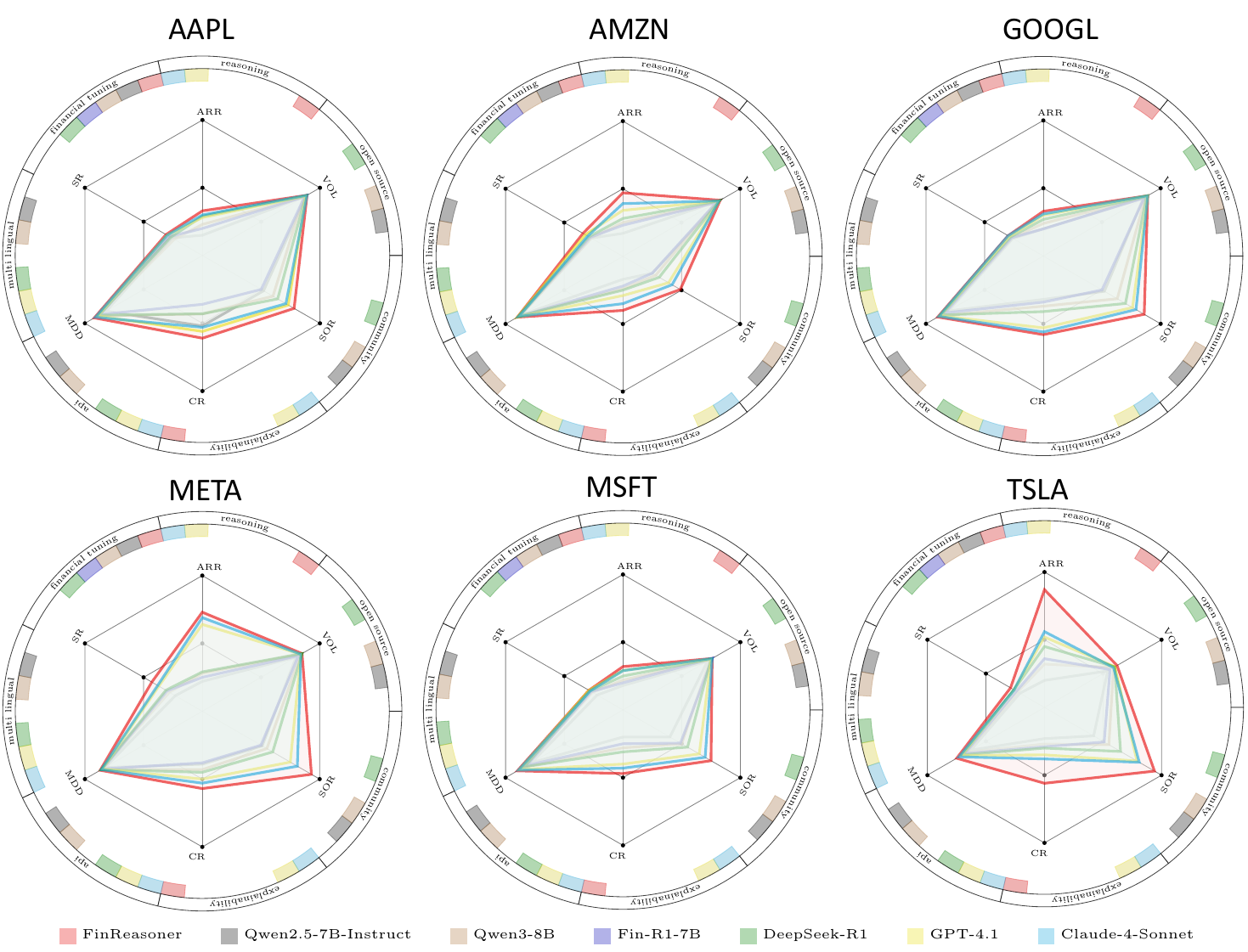}
    \caption{Comparison of LLMs on trading.}
    \label{fig:llm_trading}
    \vspace{-1em}
\end{figure}

\noindent\textbf{Experiment Results} 
As shown in Figure~\ref{fig:llm_reasoning}, \texttt{FinReasoner} leads all four financial-reasoning benchmarks, outperforming every other LLM and underscoring the value of domain-specific reasoning over generic instruction tuning. In Figure~\ref{fig:llm_trading}, we rescale all metrics (with inverted MDD and VOL) to a 0–100 range for area-based comparison. \texttt{FinReasoner} achieves strong performance on all stocks, validating its effectiveness in both reasoning and trading.

\section{Conclusion}
In this paper, we introduce \projectname, a comprehensive framework for financial AI research and development. We provide a detailed description of the framework, including the data, model, training, evaluation, and task layers. We also provide empirical evaluations on four financial AI tasks: time series forecasting, algorithmic trading, portfolio management, and LLMs applications. The results show that \projectname can effectively support financial AI research and development. We will continue to optimize the framework in the future to provide an even better user experience.



\clearpage

\bibliographystyle{ACM-Reference-Format}
\bibliography{main}

@String{Computer = "{IEEE} Computer" }

@String{Chelsea = "Chelsea" }

@article{mmengine2022,
  title   = {{MMEngine: OpenMMLab Foundational Library for Training Deep Learning Models}},
  author  = {MMEngine Contributors},
  howpublished = {\url{https://github.com/open-mmlab/mmengine}},
  year={2022}
}

@misc{fmp2025,
  author       = {Financial Modeling Prep Team},
  title        = {{Financial Modeling Prep}},
  year         = {2025},
  howpublished = {\url{https://site.financialmodelingprep.com/}},
}

@misc{alpaca2025,
  author       = {Alpaca Team},
  title        = {{Alpaca Markets}},
  year         = {2025},
  howpublished = {\url{https://alpaca.markets/}},
}

@article{yang2020qlib,
  title={{Qlib: An AI-oriented Quantitative Investment Platform}},
  author={Yang, Xiao and Liu, Weiqing and Zhou, Dong and Bian, Jiang and Liu, Tie-Yan},
  journal={arXiv preprint arXiv:2009.11189},
  year={2020}
}

@article{wu2022timesnet,
  title={{Timesnet: Temporal 2d-variation Modeling for General Time Series Analysis}},
  author={Wu, Haixu and Hu, Tengge and Liu, Yong and Zhou, Hang and Wang, Jianmin and Long, Mingsheng},
  journal={arXiv preprint arXiv:2210.02186},
  year={2022}
}

@article{dosovitskiy2020image,
  title={{An Image is Worth 16x16 Words: Transformers for Image Recognition at Scale}},
  author={Dosovitskiy, Alexey and Beyer, Lucas and Kolesnikov, Alexander and Weissenborn, Dirk and Zhai, Xiaohua and Unterthiner, Thomas and Dehghani, Mostafa and Minderer, Matthias and Heigold, Georg and Gelly, Sylvain and others},
  journal={arXiv preprint arXiv:2010.11929},
  year={2020}
}

@article{sun2023trademaster,
  title={{TradeMaster: A Holistic Quantitative Trading Platform Empowered by Reinforcement Learning}},
  author={Sun, Shuo and Qin, Molei and Zhang, Wentao and Xia, Haochong and Zong, Chuqiao and Ying, Jie and Xie, Yonggang and Zhao, Lingxuan and Wang, Xinrun and An, Bo},
  journal={Advances in Neural Information Processing Systems},
  volume={36},
  pages={59047--59061},
  year={2023}
}

@article{liu2022finrl,
  title={{FinRL-Meta: Market Environments and Benchmarks for Data-driven Financial Reinforcement Learning}},
  author={Liu, Xiao-Yang and Xia, Ziyi and Rui, Jingyang and Gao, Jiechao and Yang, Hongyang and Zhu, Ming and Wang, Christina and Wang, Zhaoran and Guo, Jian},
  journal={Advances in Neural Information Processing Systems},
  volume={35},
  pages={1835--1849},
  year={2022}
}

@article{yu2025dapo,
  title={{DAPO: An Open-source LLM Reinforcement Learning System at Scale}},
  author={Yu, Qiying and Zhang, Zheng and Zhu, Ruofei and Yuan, Yufeng and Zuo, Xiaochen and Yue, Yu and Dai, Weinan and Fan, Tiantian and Liu, Gaohong and Liu, Lingjun and others},
  journal={arXiv preprint arXiv:2503.14476},
  year={2025}
}

@article{shao2024deepseekmath,
  title={{Deepseekmath: Pushing the Limits of Mathematical Reasoning in Open Language Models}},
  author={Shao, Zhihong and Wang, Peiyi and Zhu, Qihao and Xu, Runxin and Song, Junxiao and Bi, Xiao and Zhang, Haowei and Zhang, Mingchuan and Li, YK and Wu, Y and others},
  journal={arXiv preprint arXiv:2402.03300},
  year={2024}
}

@article{fama2015five,
  title={{A Five-factor Asset Pricing Model}},
  author={Fama, Eugene F and French, Kenneth R},
  journal={Journal of Financial Dconomics},
  volume={116},
  number={1},
  pages={1--22},
  year={2015},
  publisher={Elsevier}
}

@article{ke2017lightgbm,
  title={{Lightgbm: A Highly Efficient Gradient Boosting Decision Tree}},
  author={Ke, Guolin and Meng, Qi and Finley, Thomas and Wang, Taifeng and Chen, Wei and Ma, Weidong and Ye, Qiwei and Liu, Tie-Yan},
  journal={Advances in Neural Information Processing Systems},
  volume={30},
  year={2017}
}

@article{jiang2017deep,
  title={{A Deep Reinforcement Learning Framework for the Financial Portfolio Management Problem}},
  author={Jiang, Zhengyao and Xu, Dixing and Liang, Jinjun},
  journal={arXiv preprint arXiv:1706.10059},
  year={2017}
}

@article{yu2019model,
  title={{Model-based Deep Reinforcement Learning for Dynamic Portfolio Optimization}},
  author={Yu, Pengqian and Lee, Joon Sern and Kulyatin, Ilya and Shi, Zekun and Dasgupta, Sakyasingha},
  journal={arXiv preprint arXiv:1901.08740},
  year={2019}
}

@inproceedings{liu2020adaptive,
  title={{Adaptive Quantitative Trading: An Imitative Deep Reinforcement Learning Approach}},
  author={Liu, Yang and Liu, Qi and Zhao, Hongke and Pan, Zhen and Liu, Chuanren},
  booktitle={Proceedings of the AAAI Conference on Artificial Intelligence},
  pages={2128--2135},
  year={2020}
}

@article{feng2023multi,
  title={{Multi-scale Attention Flow for Probabilistic Time Series Forecasting}},
  author={Feng, Shibo and Miao, Chunyan and Xu, Ke and Wu, Jiaxiang and Wu, Pengcheng and Zhang, Yang and Zhao, Peilin},
  journal={IEEE Transactions on Knowledge and Data Engineering},
  volume={36},
  number={5},
  pages={2056--2068},
  year={2023},
  publisher={IEEE}
}

@inproceedings{ding2020hierarchical,
  title={{Hierarchical Multi-scale Gaussian Transformer for Stock Movement Prediction}},
  author={Ding, Qianggang and Wu, Sifan and Sun, Hao and Guo, Jiadong and Guo, Jian},
  booktitle={IJCAI},
  pages={4640--4646},
  year={2020}
}

@article{liu2020finrl,
  title={{FinRL: A Deep Reinforcement Learning Library for Automated Stock Trading in Quantitative Finance}},
  author={Liu, Xiao-Yang and Yang, Hongyang and Chen, Qian and Zhang, Runjia and Yang, Liuqing and Xiao, Bowen and Wang, Christina Dan},
  journal={arXiv preprint arXiv:2011.09607},
  year={2020}
}

@inproceedings{qin2024earnhft,
  title={{Earnhft: Efficient Hierarchical Reinforcement Learning for High Frequency Trading}},
  author={Qin, Molei and Sun, Shuo and Zhang, Wentao and Xia, Haochong and Wang, Xinrun and An, Bo},
  booktitle={Proceedings of the AAAI Conference on Artificial Intelligence},
  volume={38},
  number={13},
  pages={14669--14676},
  year={2024}
}

@inproceedings{zhang2024reinforcement,
  title={{Reinforcement Learning with Maskable Stock Representation for Portfolio Management in Customizable Stock Pools}},
  author={Zhang, Wentao and Zhao, Yilei and Sun, Shuo and Ying, Jie and Xie, Yonggang and Song, Zitao and Wang, Xinrun and An, Bo},
  booktitle={Proceedings of the ACM Web Conference 2024},
  pages={187--198},
  year={2024}
}

@article{kim2025semi,
  title={{Semi-Decision-Focused Learning with Deep Ensembles: A Practical Framework for Robust Portfolio Optimization}},
  author={Kim, Juhyeong},
  journal={arXiv preprint arXiv:2503.13544},
  year={2025}
}

@article{liu2023fingpt,
  title={{FinGPT: Democratizing Internet-scale Data for Financial Large Language Models}},
  author={Liu, Xiao-Yang and Wang, Guoxuan and Yang, Hongyang and Zha, Daochen},
  journal={arXiv preprint arXiv:2307.10485},
  year={2023}
}

@article{liu2025fin,
  title={{Fin-r1: A Large Language Model for Financial Reasoning Through Reinforcement Learning}},
  author={Liu, Zhaowei and Guo, Xin and Lou, Fangqi and Zeng, Lingfeng and Niu, Jinyi and Wang, Zixuan and Xu, Jiajie and Cai, Weige and Yang, Ziwei and Zhao, Xueqian and others},
  journal={arXiv preprint arXiv:2503.16252},
  year={2025}
}

@article{box1976time,
  title={{Time Series Analysis: Forecasting and Control}},
  author={Box, George EP and Jenkins, Gwilym M and Reinsel, Gregory C and Ljung, Greta M},
  journal={San Francisco: Holden-Day},
  year={1976}
}

@article{engle1982autoregressive,
  title={{Autoregressive Conditional Heteroskedasticity with Estimates of the Variance of United Kingdom Inflation}},
  author={Engle, Robert F},
  journal={Econometrica: Journal of the Econometric Society},
  pages={987--1007},
  year={1982},
  publisher={JSTOR}
}

@article{chen2016xgboost,
  title={{XGBoost: A Scalable Tree Boosting System}},
  author={Chen, Tianqi and Guestrin, Carlos},
  journal={Proceedings of the 22nd ACM SIGKDD International Conference on Knowledge Discovery and Data Mining},
  pages={785--794},
  year={2016}
}

@article{hochreiter1997long,
  title={{Long Short-Term Memory}},
  author={Hochreiter, Sepp and Schmidhuber, J{\"u}rgen},
  journal={Neural Computation},
  volume={9},
  number={8},
  pages={1735--1780},
  year={1997},
  publisher={MIT Press}
}

@article{vaswani2017attention,
  title={{Attention Is All You Need}},
  author={Vaswani, Ashish and Shazeer, Noam and Parmar, Niki and Uszkoreit, Jakob and Jones, Llion and Gomez, Aidan N and Kaiser, {\L}ukasz and Polosukhin, Illia},
  journal={Advances in Neural Information Processing Systems},
  volume={30},
  year={2017}
}

@article{zeng2023are,
  title={{Are Transformers Effective for Time Series Forecasting?}},
  author={Zeng, Ailing and Chen, Muxi and Zhang, Lei and Xu, Qiang},
  journal={Proceedings of the AAAI Conference on Artificial Intelligence},
  volume={37},
  number={9},
  pages={11121--11128},
  year={2023}
}

@article{liu2024timexer,
  title={{TimeXer: Empowering Transformers for Time Series Forecasting with Exogenous Variables}},
  author={Liu, Yong and Wu, Haixu and Wang, Jianmin and Long, Mingsheng},
  journal={arXiv preprint arXiv:2401.13795},
  year={2024}
}

@article{schulman2017proximal,
  title={{Proximal Policy Optimization Algorithms}},
  author={Schulman, John and Wolski, Filip and Dhariwal, Prafulla and Radford, Alec and Klimov, Oleg},
  journal={arXiv preprint arXiv:1707.06347},
  year={2017}
}

@article{mnih2015human,
  title={{Human-level Control Through Deep Reinforcement Learning}},
  author={Mnih, Volodymyr and Kavukcuoglu, Koray and Silver, David and Rusu, Andrei A and Veness, Joel and Bellemare, Marc G and Graves, Alex and Riedmiller, Martin and Fidjeland, Andreas K and Ostrovski, Georg and others},
  journal={nature},
  volume={518},
  number={7540},
  pages={529--533},
  year={2015},
  publisher={Nature Publishing Group UK London}
}

@article{markowitz1952portfolio,
  title={{Portfolio Selection}},
  author={Markowitz, Harry},
  journal={The Journal of Finance},
  volume={7},
  number={1},
  pages={77--91},
  year={1952},
  publisher={Wiley Online Library}
}

@article{heaton2017deep,
  title={{Deep Portfolio Theory}},
  author={Heaton, J and Polson, N and Witte, J},
  journal={Journal of Financial Data Science},
  volume={1},
  number={1},
  pages={1--28},
  year={2017}
}

@article{shah2022flang,
  title={{FLANG: Financial Language Model for Domain-Specific Tasks}},
  author={Shah, Sameena and Chen, Zhiyu and Smiley, Charese and De, Shubhra and Wang, Xiusi and Qin, Ming and Zheng, Yiming and Arora, Pranjal and Liu, Hongxia and others},
  journal={arXiv preprint arXiv:2205.06031},
  year={2022}
}

@article{araci2019finbert,
  title={{FinBERT: Financial Sentiment Analysis with Pre-trained Language Models}},
  author={Araci, Dogu},
  journal={arXiv preprint arXiv:1908.10063},
  year={2019}
}

@article{wu2023bloomberggpt,
  title={{BloombergGPT: A Large Language Model for Finance}},
  author={Wu, Shijie and Irsoy, Ozan and Lu, Steven and Dabravolski, Vadim and Dredze, Mark and Gehrmann, Sebastian and Kambadur, Prabhanjan and Rosenberg, David and Mann, Gideon},
  journal={arXiv preprint arXiv:2303.17564},
  year={2023}
}

@article{qian2025fino1,
  title={{Fino1: On the Transferability of Reasoning-Enhanced LLMs and Reinforcement Learning to Finance}},
  author={Qian, Lingfei and Zhou, Weipeng and Wang, Yan and Peng, Xueqing and Yi, Han and Zhao, Yilun and Huang, Jimin and Xie, Qianqian and Nie, Jian-yun},
  journal={arXiv preprint arXiv:2502.08127},
  year={2025}
}

@article{zhou2024finrobot,
  title={{Finrobot: AI Agent for Equity Research and Valuation with Large Language Models}},
  author={Zhou, Tianyu and Wang, Pinqiao and Wu, Yilin and Yang, Hongyang},
  journal={arXiv preprint arXiv:2411.08804},
  year={2024}
}

@article{yu2024fincon,
  title={{Fincon: A Synthesized LLM Multi-Agent System with Conceptual Verbal Reinforcement for Enhanced Financial Decision Making}},
  author={Yu, Yangyang and Yao, Zhiyuan and Li, Haohang and Deng, Zhiyang and Jiang, Yuechen and Cao, Yupeng and Chen, Zhi and Suchow, Jordan and Cui, Zhenyu and Liu, Rong and others},
  journal={Advances in Neural Information Processing Systems},
  volume={37},
  pages={137010--137045},
  year={2024}
}

@inproceedings{zhang2024multimodal,
  title={{A Multimodal Foundation Agent for Financial Trading: Tool-Augmented, Diversified, and Generalist}},
  author={Zhang, Wentao and Zhao, Lingxuan and Xia, Haochong and Sun, Shuo and Sun, Jiaze and Qin, Molei and Li, Xinyi and Zhao, Yuqing and Zhao, Yilei and Cai, Xinyu and others},
  booktitle={Proceedings of the 30th ACM SIGKDD Conference on Knowledge Discovery and Data Mining},
  pages={4314--4325},
  year={2024}
}

@inproceedings{yu2024finmem,
  title={{Finmem: A Performance-enhanced LLM Trading Agent with Layered Memory and Character Design}},
  author={Yu, Yangyang and Li, Haohang and Chen, Zhi and Jiang, Yuechen and Li, Yang and Zhang, Denghui and Liu, Rong and Suchow, Jordan W and Khashanah, Khaldoun},
  booktitle={Proceedings of the AAAI Symposium Series},
  volume={3},
  number={1},
  pages={595--597},
  year={2024}
}

@article{xiao2024tradingagents,
  title={{TradingAgents: Multi-agents LLM Financial Trading Framework}},
  author={Xiao, Yijia and Sun, Edward and Luo, Di and Wang, Wei},
  journal={arXiv preprint arXiv:2412.20138},
  year={2024}
}

@article{chen2021finqa,
  title={{FinQA: A Dataset of Numerical Reasoning over Financial Data}},
  author={Chen, Zhiyu and Chen, Wenhu and Smiley, Charese and Shah, Sameena and Borova, Iana and Langdon, Dylan and Moussa, Reema and Beane, Matt and Huang, Ting-Hao and others},
  journal={Proceedings of the 2021 Conference on Empirical Methods in Natural Language Processing},
  pages={3697--3711},
  year={2021}
}

@article{feng2025group,
  title={{Group-in-group Policy Optimization for LLM Agent Training}},
  author={Feng, Lang and Xue, Zhenghai and Liu, Tingcong and An, Bo},
  journal={arXiv preprint arXiv:2505.10978},
  year={2025}
}

@misc{agent-r1,
  title        = {{Training Powerful LLM Agents with End-to-End Reinforcement Learning}},
  author       = {Jie Ouyang and Ruiran Yan and Yucong Luo and Mingyue Cheng and Qi Liu and Zirui Liu and Shuo Yu and Daoyu Wang},
  year         = {2025},
  organization = {GitHub},
  url          = {https://github.com/0russwest0/Agent-R1}
}

@article{zhang2023fineval,
  title={{Fineval: A Chinese Financial Domain Knowledge Evaluation Benchmark for Large Language Models}},
  author={Zhang, Liwen and Cai, Weige and Liu, Zhaowei and Yang, Zhi and Dai, Wei and Liao, Yujie and Qin, Qianru and Li, Yifei and Liu, Xingyu and Liu, Zhiqiang and others},
  journal={arXiv preprint arXiv:2308.09975},
  year={2023}
}

@inproceedings{zhu2024cflue,
  title={{Benchmarking Large Language Models on CFLUE - A Chinese Financial Language Understanding Evaluation Dataset}}, 
  author={Zhu, Jie and Li, Junhui and Wen, Yalong and Guo, Lifan},
  booktitle={Proceedings of the 62nd Annual Meeting of the Association for Computational Linguistics (ACL-2024)},
  year={2024}
}

@article{ke2025demystifyingdomainadaptiveposttrainingfinancial,
  title={{Demystifying Domain-adaptive Post-training for Financial LLMs}},
  author={Ke, Zixuan and Ming, Yifei and Nguyen, Xuan-Phi and Xiong, Caiming and Joty, Shafiq},
  journal={arXiv preprint arXiv:2501.04961},
  year={2025}
}

@article{chen2022convfinqa,
  title={{Convfinqa: Exploring the chain of numerical reasoning in conversational finance question answering}},
  author={Chen, Zhiyu and Li, Shiyang and Smiley, Charese and Ma, Zhiqiang and Shah, Sameena and Wang, William Yang},
  journal={arXiv preprint arXiv:2210.03849},
  year={2022}
}

@misc{pyecharts,
  title        = {{Pyecharts: Python Echarts Plotting Library}},
  author       = {Pyecharts Team},
  year         = {2025},
  organization = {GitHub},
  url          = {https://github.com/pyecharts/pyecharts},
  note         = {Version 2.0.8}
}

@article{wu2021autoformer,
  title={{Autoformer: Decomposition Transformers with Auto-correlation for Long-term Series Forecasting}},
  author={Wu, Haixu and Xu, Jiehui and Wang, Jianmin and Long, Mingsheng},
  journal={Advances in Neural Information Processing Systems},
  volume={34},
  pages={22419--22430},
  year={2021}
}

@inproceedings{zhang2023crossformer,
  title={{Crossformer: Transformer Utilizing Cross-dimension Dependency for Multivariate Time Series Forecasting}},
  author={Zhang, Yunhao and Yan, Junchi},
  booktitle={The Eleventh International Conference on Learning Representations},
  year={2023}
}

@article{woo2022etsformer,
  title={{Etsformer: Exponential Smoothing Transformers for Time-series Forecasting}},
  author={Woo, Gerald and Liu, Chenghao and Sahoo, Doyen and Kumar, Akshat and Hoi, Steven},
  journal={arXiv preprint arXiv:2202.01381},
  year={2022}
}

@inproceedings{zeng2023transformers,
  title={{Are Transformers Effective for Time Series Forecasting?}},
  author={Zeng, Ailing and Chen, Muxi and Zhang, Lei and Xu, Qiang},
  booktitle={Proceedings of the AAAI Conference on Artificial Intelligence},
  volume={37},
  number={9},
  pages={11121--11128},
  year={2023}
}

@article{nie2022time,
  title={{A Time Series is Worth 64 Words: Long-term Forecasting with Transformers}},
  author={Nie, Yuqi and Nguyen, Nam H and Sinthong, Phanwadee and Kalagnanam, Jayant},
  journal={arXiv preprint arXiv:2211.14730},
  year={2022}
}

@article{wang2024timemixer,
  title={{Timemixer: Decomposable Multiscale Mixing for Time Series Forecasting}},
  author={Wang, Shiyu and Wu, Haixu and Shi, Xiaoming and Hu, Tengge and Luo, Huakun and Ma, Lintao and Zhang, James Y and Zhou, Jun},
  journal={arXiv preprint arXiv:2405.14616},
  year={2024}
}

@article{wang2024timexer,
  title={{TimeXer: Empowering Transformers for Time Series Forecasting with Exogenous Variables}},
  author={Wang, Yuxuan and Wu, Haixu and Dong, Jiaxiang and Qin, Guo and Zhang, Haoran and Liu, Yong and Qiu, Yunzhong and Wang, Jianmin and Long, Mingsheng},
  journal={Advances in Neural Information Processing Systems},
  volume={37},
  pages={469--498},
  year={2024}
}

@misc{tslib,
  title        = {{Time Series Library}},
  author       = {TSLib Team},
  year         = {2025},
  organization = {GitHub},
  url          = {https://github.com/thuml/Time-Series-Library},
}

@inproceedings{duan2022factorvae,
  title={{FactorVAE: A Probabilistic Dynamic Factor Model Based on Variational Autoencoder for Predicting Cross-sectional Stock Returns}},
  author={Duan, Yitong and Wang, Lei and Zhang, Qizhong and Li, Jian},
  booktitle={Proceedings of the AAAI Conference on Artificial Intelligence},
  volume={36},
  number={4},
  pages={4468--4476},
  year={2022}
}

@article{wei2023hirevae,
  title={{HireVAE: An Online and Adaptive Factor Model Based on Hierarchical and Regime-Switch VAE}},
  author={Wei, Zikai and Rao, Anyi and Dai, Bo and Lin, Dahua},
  journal={arXiv preprint arXiv:2306.02848},
  year={2023}
}

@article{zhao2024storm,
  title={STORM: A Spatio-Temporal Factor Model Based on Dual Vector Quantized Variational Autoencoders for Financial Trading},
  author={Zhao, Yilei and Zhang, Wentao and Yang, Tingran and Jiang, Yong and Huang, Fei and Lim, Wei Yang Bryan},
  journal={arXiv preprint arXiv:2412.09468},
  year={2024}
}

@misc{vae,
  title={{Auto-encoding Variational Bayes}},
  author={Kingma, Diederik P and Welling, Max and others},
  year={2013},
  publisher={Banff, Canada}
}

@article{vqvae,
  title={{Neural Discrete Representation Learning}},
  author={Van Den Oord, Aaron and Vinyals, Oriol and others},
  journal={Advances in Neural Information Processing Systems},
  volume={30},
  year={2017}
}

@inproceedings{mae,
  title={{Masked Autoencoders are Scalable Vision Learners}},
  author={He, Kaiming and Chen, Xinlei and Xie, Saining and Li, Yanghao and Doll{\'a}r, Piotr and Girshick, Ross},
  booktitle={Proceedings of the IEEE/CVF Conference on Computer Vision and Pattern Recognition},
  pages={16000--16009},
  year={2022}
}

@article{haarnoja2018soft,
  title={{Soft Actor-Critic Algorithms and Applications}},
  author={Haarnoja, Tuomas and Zhou, Aurick and Hartikainen, Kristian and Tucker, George and Ha, Sehoon and Tan, Jie and Kumar, Vikash and Zhu, Henry and Gupta, Abhishek and Abbeel, Pieter and others},
  journal={arXiv preprint arXiv:1812.05905},
  year={2018}
}

@article{zhu2025dianjin,
  title={{Dianjin-r1: Evaluating and enhancing financial reasoning in large language models}},
  author={Zhu, Jie and Chen, Qian and Dou, Huaixia and Li, Junhui and Guo, Lifan and Chen, Feng and Zhang, Chi},
  journal={arXiv preprint arXiv:2504.15716},
  year={2025}
}

@article{guo2025deepseek,
  title={{Deepseek-r1: Incentivizing Reasoning Capability in LLMs via Reinforcement Learning}},
  author={Guo, Daya and Yang, Dejian and Zhang, Haowei and Song, Junxiao and Zhang, Ruoyu and Xu, Runxin and Zhu, Qihao and Ma, Shirong and Wang, Peiyi and Bi, Xiao and others},
  journal={arXiv preprint arXiv:2501.12948},
  year={2025}
}

@article{yang2025qwen3,
  title={{Qwen3 Technical Report}},
  author={Yang, An and Li, Anfeng and Yang, Baosong and Zhang, Beichen and Hui, Binyuan and Zheng, Bo and Yu, Bowen and Gao, Chang and Huang, Chengen and Lv, Chenxu and others},
  journal={arXiv preprint arXiv:2505.09388},
  year={2025}
}

@misc{qwen2025qwen25technicalreport,
      title={{Qwen2.5 Technical Report}}, 
      author={Qwen Team},
      year={2025},
      eprint={2412.15115},
      archivePrefix={arXiv},
      primaryClass={cs.CL},
      url={https://arxiv.org/abs/2412.15115}, 
}

@misc{verl,
  title        = {{Volcano Engine Reinforcement Learning for LLMs}},
  author       = {Verl Team},
  year         = {2025},
  organization = {GitHub},
  url          = {https://github.com/volcengine/verl},
}

@misc{ragen,
      title={{RAGEN: Understanding Self-Evolution in LLM Agents via Multi-Turn Reinforcement Learning}}, 
      author={Zihan Wang and Kangrui Wang and Qineng Wang and Pingyue Zhang and Linjie Li and Zhengyuan Yang and Xing Jin and Kefan Yu and Minh Nhat Nguyen and Licheng Liu and Eli Gottlieb and Yiping Lu and Kyunghyun Cho and Jiajun Wu and Li Fei-Fei and Lijuan Wang and Yejin Choi and Manling Li},
      year={2025},
      eprint={2504.20073},
      archivePrefix={arXiv},
      primaryClass={cs.LG},
      url={https://arxiv.org/abs/2504.20073}, 
}

@article{ahn2022can,
  title={{Do as I Can, not as I Say: Grounding Language in Robotic Affordances}},
  author={Ahn, Michael and Brohan, Anthony and Brown, Noah and Chebotar, Yevgen and Cortes, Omar and David, Byron and Finn, Chelsea and Fu, Chuyuan and Gopalakrishnan, Keerthana and Hausman, Karol and others},
  journal={arXiv preprint arXiv:2204.01691},
  year={2022}
}

@inproceedings{yao2023react,
  title={{React: Synergizing Reasoning and Acting in Language Models}},
  author={Yao, Shunyu and Zhao, Jeffrey and Yu, Dian and Du, Nan and Shafran, Izhak and Narasimhan, Karthik and Cao, Yuan},
  booktitle={International Conference on Learning Representations (ICLR)},
  year={2023}
}

@article{wang2023voyager,
  title={{Voyager: An Open-Ended Embodied Agent with Large Language Models}},
  author={Wang, Guanzhi and Xie, Yuqi and Jiang, Yunfan and Mandlekar, Ajay and Xiao, Chaowei and Zhu, Yuke and Fan, Linxi and Anandkumar, Anima},
  journal={arXiv preprint arXiv:2305.16291},
  year={2023}
}

@article{zhang2025agentorchestra,
  title={AgentOrchestra: A Hierarchical Multi-Agent Framework for General-Purpose Task Solving},
  author={Zhang, Wentao and Cui, Ce and Zhao, Yilei and Liu, Yang and An, Bo},
  journal={arXiv preprint arXiv:2506.12508},
  year={2025}
}

\clearpage

\onecolumn

\appendix
\section*{\centering\Huge Appendix}

\section{FinWorld as a Dataset Platform}
\label{appx:dataset}

\begin{figure}[htbp]
  \centering

  \begin{subfigure}[b]{0.45\textwidth}
    \centering
    \includegraphics[width=\textwidth]{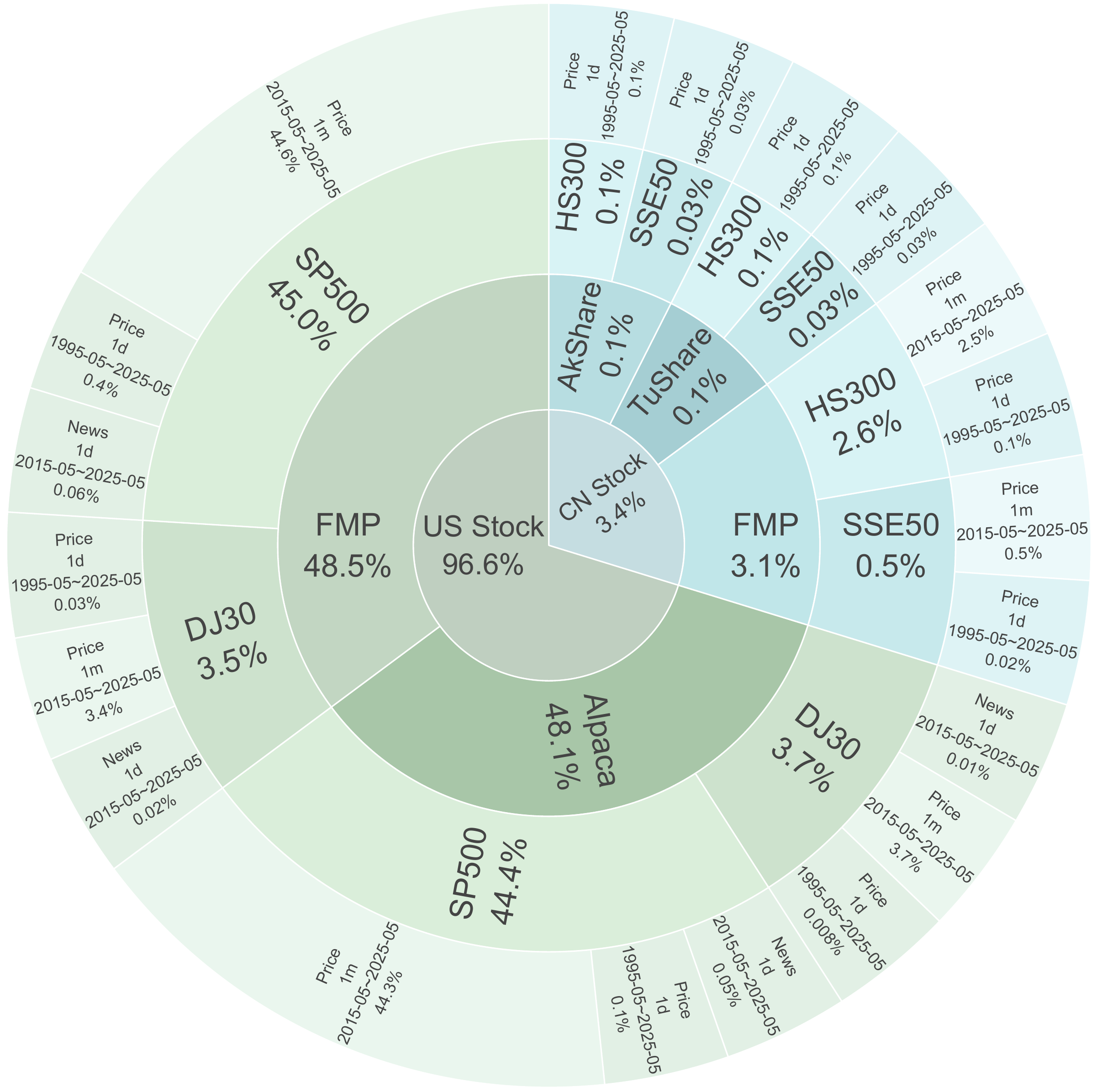}
    \caption{Market datasets.}
    \label{fig:stock_dataset}
  \end{subfigure}%
  \begin{subfigure}[b]{0.45\textwidth}
    \centering
    \includegraphics[width=\textwidth]{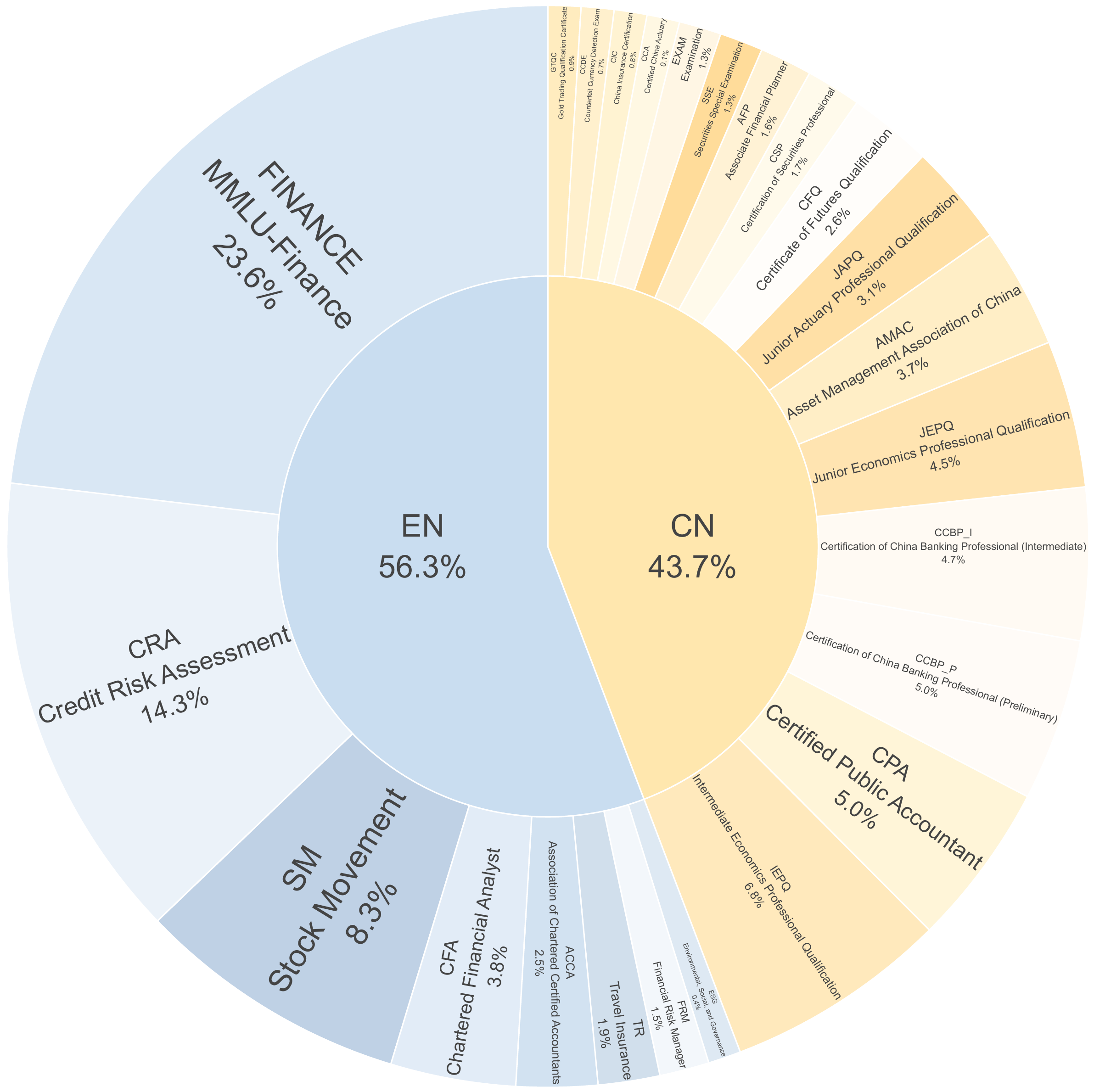}
    \caption{LLM reasoning datasets.}
    \label{fig:llm_dataset}
  \end{subfigure}

  \caption{Overview of dataset.}
  \label{fig:dataset}
\end{figure}

\subsection{Market Dataset}
\label{appx:market_dataset}
We select two representative markets for comprehensive financial AI research: the US market, representing developed markets with high liquidity and extensive data availability, and the Chinese market, representing emerging markets with unique characteristics and regulatory environments. For the US stock pools, we choose the Dow Jones Industrial Average (DJ30) and S\&P 500 (SP500), which are widely recognized benchmarks covering large-cap and blue-chip stocks. For the Chinese stock pools, we select the SSE 50 Index (SSE50) and CSI 300 Index (HS300), representing the most liquid and representative stocks in the Shanghai and Shenzhen markets.

Our dataset integrates multiple data sources to ensure comprehensive coverage and data quality. We utilize four primary data providers: Financial Modeling Prep (FMP), Alpaca Markets, AkShare, and TuShare. These sources provide daily and minute-level price data, along with news data from various financial platforms including Seeking Alpha, Bloomberg, and other major financial news outlets. The dataset spans from 1995 to 2025, offering extensive historical coverage for robust model training and evaluation.

The data processing pipeline includes comprehensive preprocessing steps to ensure data quality and usability. For price data, we implement filtering procedures to remove non-trading days and invalid data points, followed by the calculation of technical indicators using the Alpha158~\cite{yang2020qlib} (Table~\ref{tab:alpha158}) framework, which provides a solid foundation for deep learning modeling. For news text data, we employ a two-stage processing approach: first, we use web scraping scripts to extract raw content from original news links, then we utilize Qwen3-32B to generate concise summaries, enabling efficient text analysis and feature extraction for financial AI applications.

After processing all stock data, we obtain over 800 million valid data points, providing comprehensive support for time series forecasting, algorithmic trading, portfolio management, and LLM applications. Furthermore, we provide unified download and data processing interfaces that theoretically enable unlimited data expansion by simply specifying stock pools and date ranges. Processed data is uniformly uploaded to Hugging Face for storage, and we offer unified dataset encapsulation for both single-stock and multi-stock scenarios, allowing flexible loading of different data types. This constitutes an efficient one-stop framework that integrates data downloading, processing, storage, and retrieval capabilities.


\begin{table}[htb]
    \caption{Alpha158 Technical Indicators Overview}
    \label{tab:alpha158}
    \centering
    \footnotesize
    \setlength{\tabcolsep}{1pt}
    \renewcommand{\arraystretch}{0.9}
    \resizebox{0.98\textwidth}{!}{
    \begin{tabular}{lll}
    \toprule
    \textbf{Indicator} & \textbf{Formula} & \textbf{Description} \\
    \cmidrule(lr){1-1} \cmidrule(lr){2-2} \cmidrule(lr){3-3}
    
    \rowcolor{gray!15}
    \multicolumn{3}{c}{\textbf{Kline-based Indicators}} \\
    kmid & $(close - open) / close$ & K-Mid price ratio \\
    kmid2 & $(close - open) / (high - low)$ & Normalized K-Mid price ratio \\
    klen & $(high - low) / open$ & K-Length price ratio \\
    kup & $(high - max(open, close)) / open$ & K-Up price ratio \\
    kup2 & $(high - max(open, close)) / (high - low)$ & Normalized K-Up price ratio \\
    klow & $(min(open, close) - low) / open$ & K-Low price ratio \\
    klow2 & $(min(open, close) - low) / (high - low)$ & Normalized K-Low price ratio \\
    ksft & $(2 \times close - high - low) / open$ & K-Shift price ratio \\
    ksft2 & $(2 \times close - high - low) / (high - low)$ & Normalized K-Shift price ratio \\
    
    \rowcolor{gray!15}
    \multicolumn{3}{c}{\textbf{Price-based Rolling Indicators (for each window $w$)} } \\
    roc\_$w$ & $close.shift(w) / close$ & Rate of change over $w$ periods \\
    ma\_$w$ & $close.rolling(w).mean() / close$ & Moving average ratio over $w$ periods \\
    std\_$w$ & $close.rolling(w).std() / close$ & Standard deviation ratio over $w$ periods \\
    beta\_$w$ & $(close.shift(w) - close) / (w \times close)$ & Beta coefficient over $w$ periods \\
    max\_$w$ & $close.rolling(w).max() / close$ & Maximum price ratio over $w$ periods \\
    min\_$w$ & $close.rolling(w).min() / close$ & Minimum price ratio over $w$ periods \\
    qtlu\_$w$ & $(close - close.rolling(w).quantile(0.8)) / close$ & Upper quantile ratio over $w$ periods \\
    qtld\_$w$ & $(close - close.rolling(w).quantile(0.2)) / close$ & Lower quantile ratio over $w$ periods \\
    rank\_$w$ & $close.rolling(w).rank(pct=True).iloc[-1] / w$ & Price rank percentile over $w$ periods \\
    
    \rowcolor{gray!15}
    \multicolumn{3}{c}{\textbf{Position-based Indicators (for each window $w$)} } \\
    imax\_$w$ & $high.rolling(w).argmax() / w$ & Position of maximum high over $w$ periods \\
    imin\_$w$ & $low.rolling(w).argmin() / w$ & Position of minimum low over $w$ periods \\
    imxd\_$w$ & $(high.rolling(w).argmax() - low.rolling(w).argmin()) / w$ & Distance between max/min positions \\
    
    \rowcolor{gray!15}
    \multicolumn{3}{c}{\textbf{RSV and Count Indicators (for each window $w$)} } \\
    rsv\_$w$ & $(close - min(low, close.shift(w))) / (max(high, close.shift(w)) - min(low, close.shift(w)))$ & RSV (Raw Stochastic Value) \\
    cntp\_$w$ & $ret1.gt(0).rolling(w).sum() / w$ & Count of positive returns over $w$ periods \\
    cntn\_$w$ & $ret1.lt(0).rolling(w).sum() / w$ & Count of negative returns over $w$ periods \\
    cntd\_$w$ & $cntp\_w - cntn\_w$ & Difference between positive and negative return counts \\
    
    \rowcolor{gray!15}
    \multicolumn{3}{c}{\textbf{Correlation Indicators (for each window $w$)} } \\
    corr\_$w$ & $close.rolling(w).corr(\log(volume + 1).rolling(w))$ & Correlation between close and log volume \\
    cord\_$w$ & $(close / close.shift(1)).rolling(w).corr(\log(volume / volume.shift(1) + 1).rolling(w))$ & Normalized correlation \\
    
    \rowcolor{gray!15}
    \multicolumn{3}{c}{\textbf{Sum-based Indicators (for each window $w$)} } \\
    sump\_$w$ & $pos\_ret1.rolling(w).sum() / abs\_ret1.rolling(w).sum()$ & Sum of positive returns ratio \\
    sumn\_$w$ & $1 - sump\_w$ & Sum of negative returns ratio \\
    sumd\_$w$ & $2 \times sump\_w - 1$ & Difference between positive and negative return sums \\
    
    \rowcolor{gray!15}
    \multicolumn{3}{c}{\textbf{Volume-based Indicators (for each window $w$)} } \\
    vma\_$w$ & $volume.rolling(w).mean() / volume$ & Volume moving average ratio \\
    vstd\_$w$ & $volume.rolling(w).std() / volume$ & Volume standard deviation ratio \\
    wvma\_$w$ & $|close / close.shift(1) - 1|.rolling(w).std() / |close / close.shift(1) - 1|.rolling(w).mean()$ & Weighted volume moving average \\
    vsump\_$w$ & $pos\_vchg1.rolling(w).sum() / abs\_vchg1.rolling(w).sum()$ & Sum of positive volume changes ratio \\
    vsumn\_$w$ & $1 - vsump\_w$ & Sum of negative volume changes ratio \\
    vsumd\_$w$ & $2 \times vsump\_w - 1$ & Difference between positive and negative volume change sums \\
    
    \rowcolor{gray!15}
    \multicolumn{3}{c}{\textbf{Volume Logarithmic Indicator}} \\
    logvol & $\log(volume + 1)$ & Logarithm of volume \\
    
    \bottomrule
    \end{tabular}
    }
    \end{table}

\subsection{LLM Reasoning Dataset}
\label{appx:llm_reasoning_dataset}
Our LLM reasoning dataset collection process is designed to provide comprehensive coverage of financial reasoning capabilities, drawing inspiration from established datasets such as FinQA~\cite{chen2021finqa}, FinEval~\cite{zhang2023fineval}, CFLUE~\cite{zhu2024cflue}, Salesforce FinEval~\cite{ke2025demystifyingdomainadaptiveposttrainingfinancial}, and ConvFinQA~\cite{chen2022convfinqa}. We systematically collect diverse types of financial reasoning data through multiple channels and sources, ensuring broad coverage of financial knowledge domains and reasoning patterns.

The dataset encompasses both Chinese and English content, covering diverse financial scenarios including financial question-answering data, industry-specific financial corpus, professional certification examination materials, business domain knowledge, tabular financial data, market analysis reports, multi-turn conversational interactions, and quantitative investment scenarios. Each scenario is carefully curated to reflect real-world financial applications and reasoning challenges, with particular emphasis on comprehensive coverage of financial industry qualification examination data. 

In total, we have collected over 80k samples including both training and testing data, providing comprehensive support for RL-based LLM reasoning training and evaluation across diverse financial scenarios and professional domains.

\section{FinWorld as a Visualization Toolkit}

Effective visualization is crucial for understanding complex financial AI systems, encompassing datasets, algorithms, models, performance metrics, experimental results and so on. To support comprehensive research and analysis across the entire financial AI workflow, we have developed a comprehensive suite of visualization tools that provide intuitive and interactive representations of our platform's capabilities. These tools enable researchers to quickly grasp the scope and characteristics of our datasets, understand algorithm performance and model behaviors, analyze training dynamics and convergence patterns, compare different approaches across multiple metrics, and visualize complex financial decision-making processes. The visualization framework facilitates informed decision-making in model selection, hyperparameter tuning, performance evaluation, and research design across all four core financial AI tasks. The visualization tools are implemented using Python, LaTex, Pyecharts~\cite{pyecharts}, and other visualization libraries.

\subsection{Dataset Visualization}

We develop a set of visualization tools to help researchers explore and analyze our financial datasets. These tools are designed to provide intuitive and interactive visual representations of dataset composition, distribution, and characteristics, enabling researchers to quickly understand the scope and diversity of available financial data. 

\textbf{Dataset Statistics Tools}. As illustrated in Figure~\ref{fig:dataset}, we employ hierarchical sunburst diagrams to comprehensively visualize the composition, distribution, and characteristics of our financial datasets. These visualizations provide an intuitive representation of data types, proportions, and hierarchical relationships across different market segments and financial instruments. These hierarchical visualizations serve as powerful tools for dataset exploration and analysis, providing researchers with immediate insights into data availability, distribution patterns, and potential applications. The interactive nature of these visualizations allows for dynamic exploration of different data segments, supporting both high-level overview and detailed examination of specific data categories. This comprehensive visualization framework enhances the usability and accessibility of our financial datasets, facilitating efficient research workflow and promoting transparent data understanding across the financial AI community.

\begin{figure}[htbp]
  \centering

  \begin{subfigure}[b]{0.45\textwidth}
    \centering
    \includegraphics[width=\textwidth]{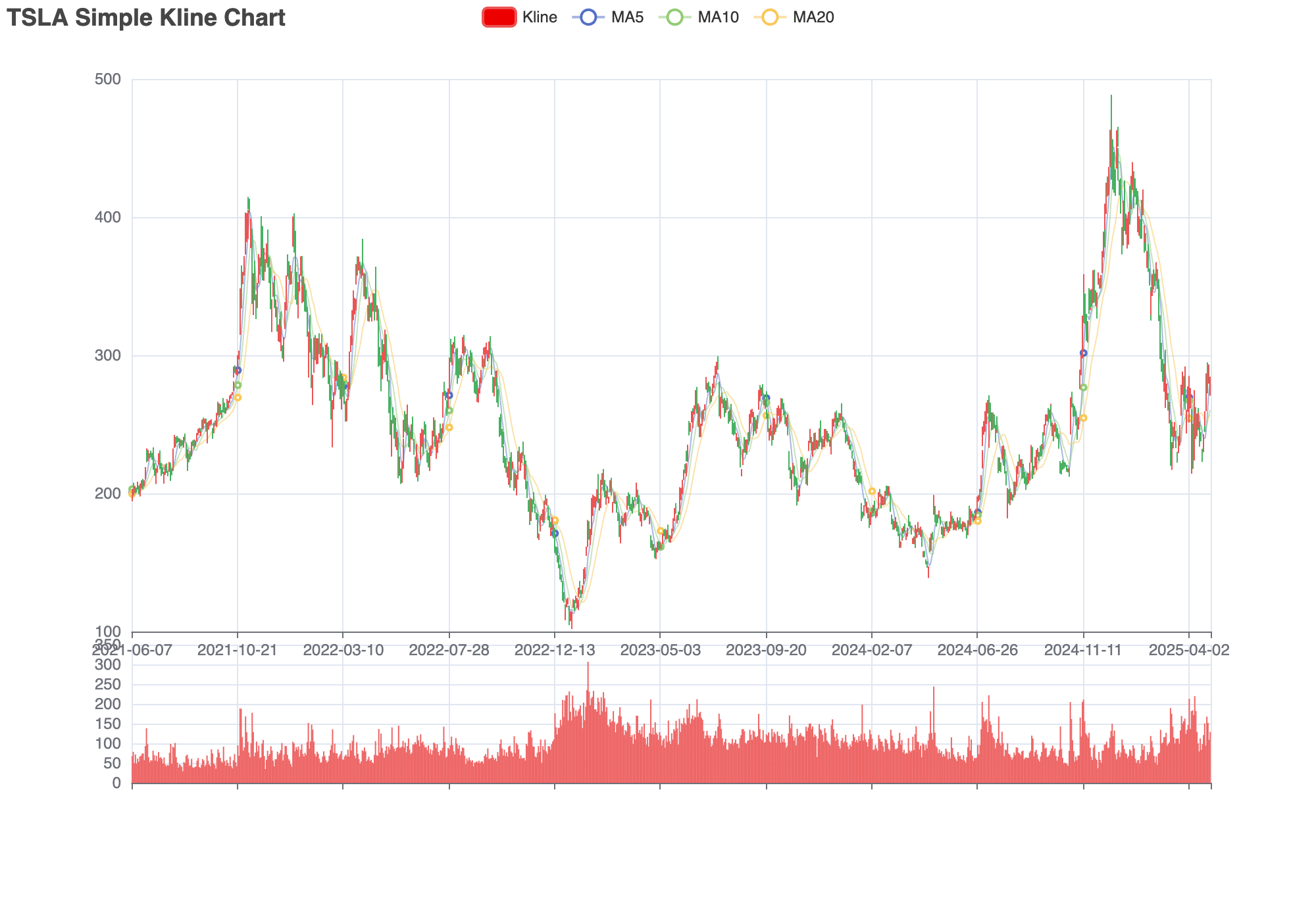}
    \caption{Long-term Kline Chart.}
    \label{fig:long_term_kline}
  \end{subfigure}%
  \begin{subfigure}[b]{0.45\textwidth}
    \centering
    \includegraphics[width=\textwidth]{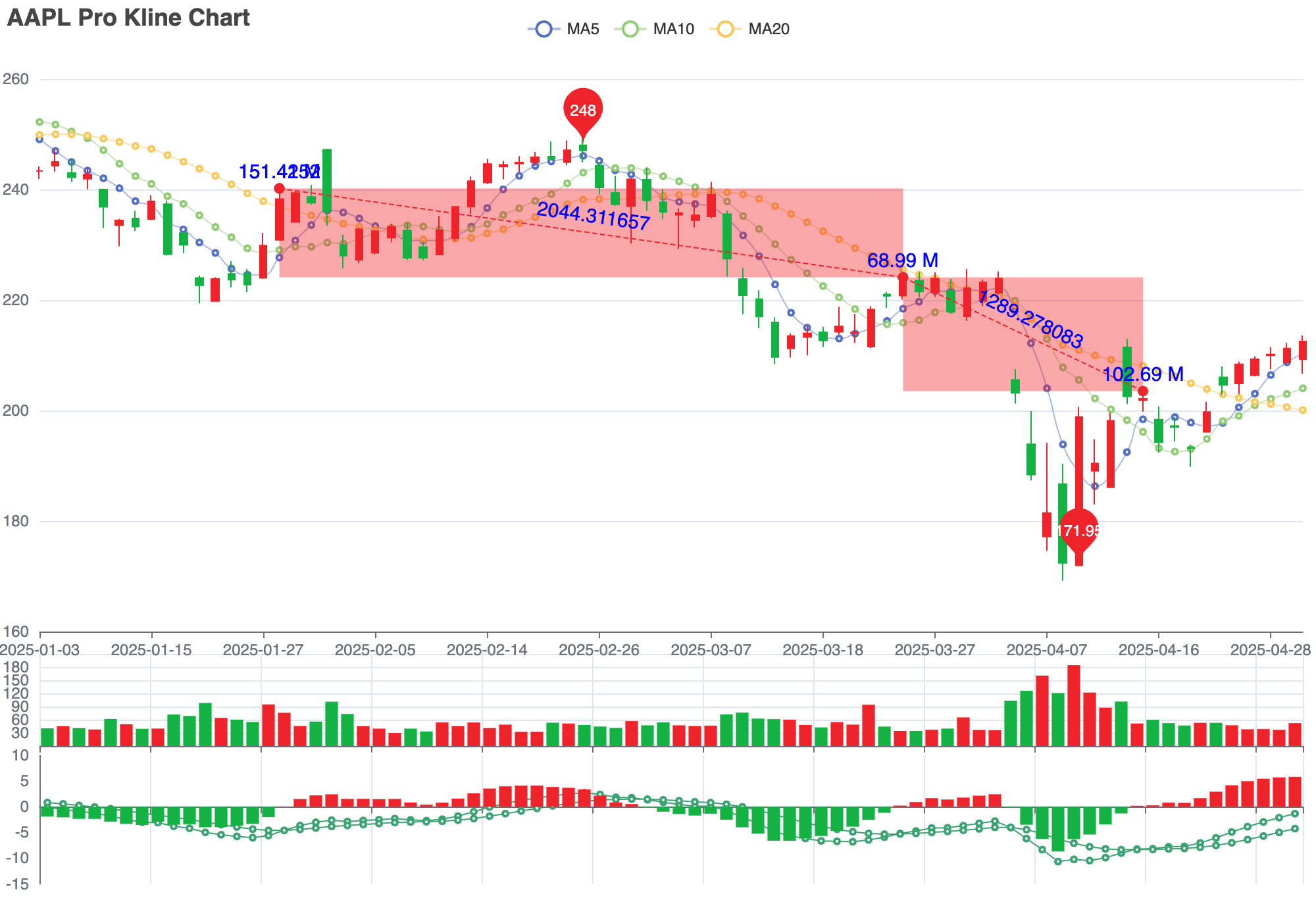}
    \caption{Short-term Kline Chart.}
    \label{fig:short_term_kline}
  \end{subfigure}

  \caption{Overview of double modes of Kline Chart.}
  \label{fig:kline_chart}
\end{figure}

\textbf{Kline Chart Tools}. To effectively visualize financial data, we employ two distinct Kline Chart display modes tailored to different analysis requirements. The first mode is designed for long-term datasets where detailed granularity is not essential, as illustrated in Figure~\ref{fig:long_term_kline}, which displays simplified Kline charts with basic moving averages (MA5, MA10, MA20) and trading volume. This streamlined visualization enables researchers to quickly grasp overall market trends and price movements across extended time periods without being overwhelmed by excessive technical details. Additionally, our Kline Chart visualizations are automatically generated as interactive HTML files, enabling online browsing and interactive exploration of the visualization information through web-based interfaces. 

The second mode is optimized for short-term data analysis requiring detailed technical indicators and trading signals, as shown in Figure~\ref{fig:short_term_kline}. This comprehensive display includes multiple moving averages (MA5, MA10, MA20), MACD (Moving Average Convergence Divergence) with its components (DIF and DEA), and provides real-time technical signals such as golden cross formations, death crosses, and other critical trading indicators. This detailed visualization mode supports precise technical analysis, enabling researchers to identify specific entry and exit points, analyze momentum shifts, and evaluate the effectiveness of various technical trading strategies. The dual-mode approach ensures that researchers can select the appropriate level of detail based on their specific analysis needs, from high-level trend analysis to granular technical signal detection.

\begin{figure}[ht]
  \centering
  \begin{subfigure}[b]{0.40\textwidth}
    \centering
    \includegraphics[width=\textwidth]{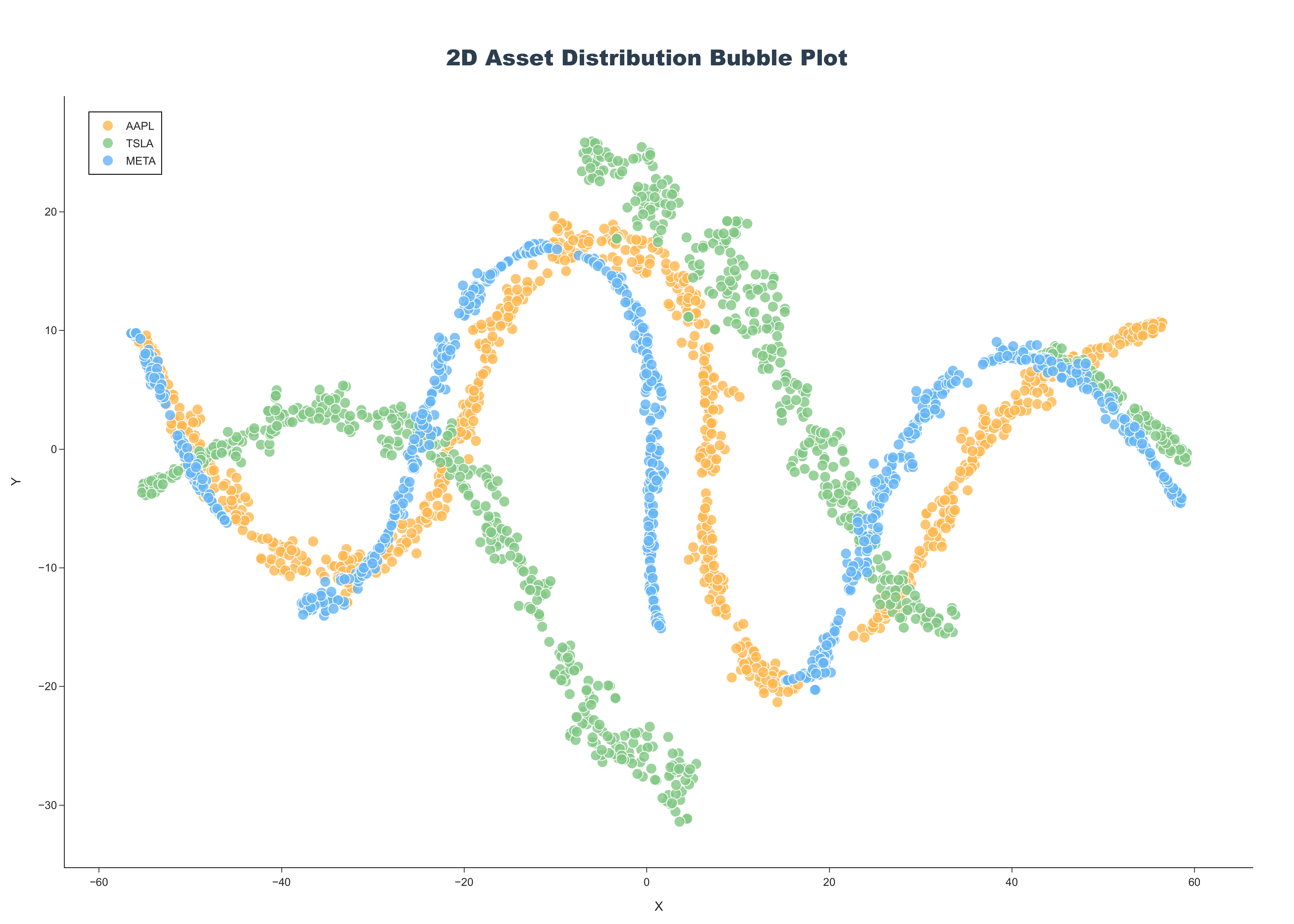}
    \caption{Asset-level t-SNE visualization with 2D bubble.}
    \label{fig:distribution_2d_bubble}
  \end{subfigure}
  \begin{subfigure}[b]{0.40\textwidth}
    \centering
    \includegraphics[width=\textwidth]{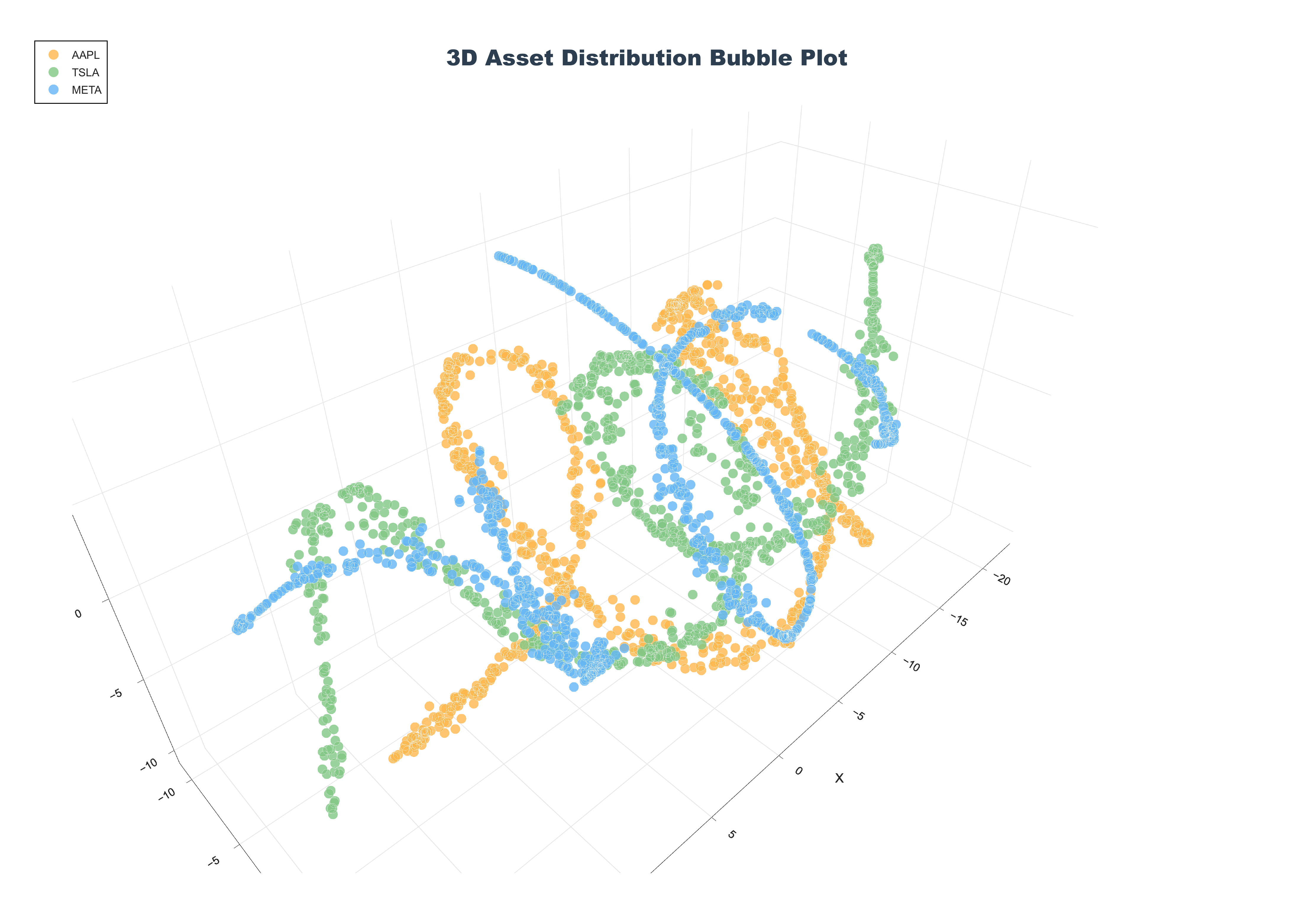}
    \caption{Asset-level t-SNE visualization with 3D bubble.}
    \label{fig:distribution_3d_bubble}
  \end{subfigure}
  \caption{Overview of asset-level t-SNE visualization.}
  \label{fig:asset_level_t_sne_visualization}
\end{figure}

\begin{figure}[ht]
  \centering
  \begin{subfigure}[b]{0.40\textwidth}
    \centering
    \includegraphics[width=\textwidth]{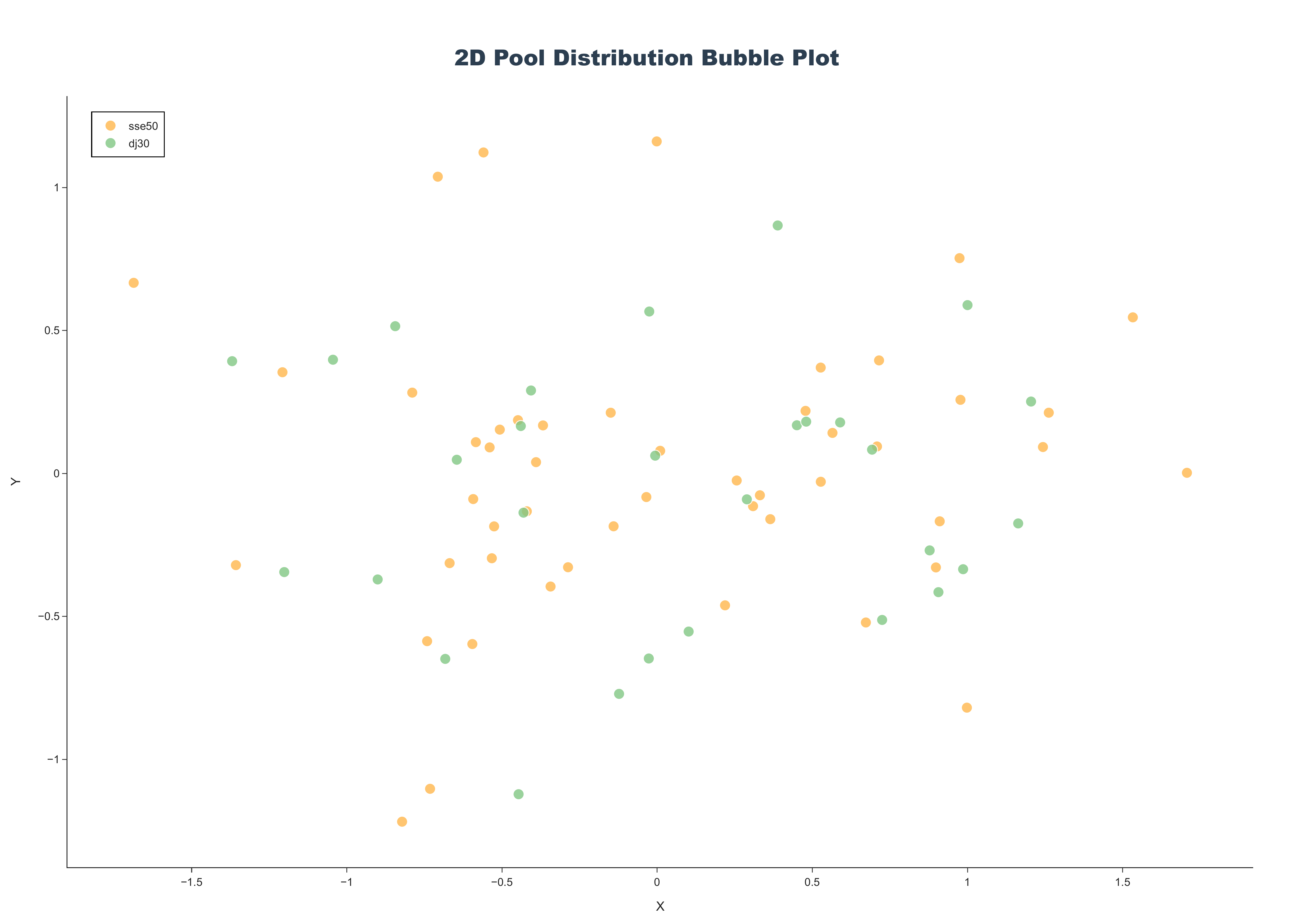}
    \caption{Pool-level t-SNE visualization with 2D bubble.}
    \label{fig:pool_level_t_sne_visualization_2d}
  \end{subfigure}
  \begin{subfigure}[b]{0.40\textwidth}
    \centering
    \includegraphics[width=\textwidth]{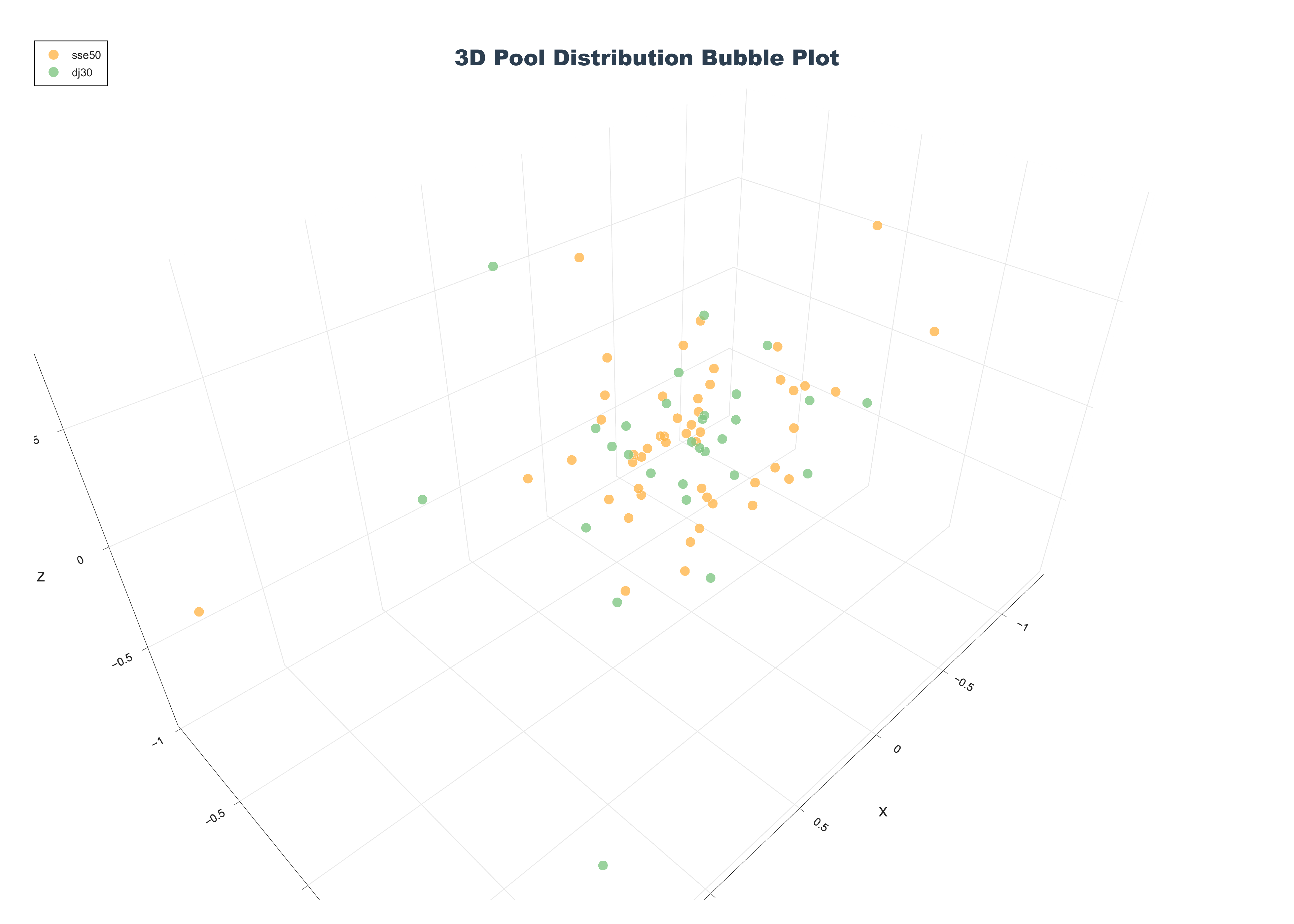}
    \caption{Pool-level t-SNE visualization with 3D bubble.}
    \label{fig:pool_level_t_sne_visualization_3d}
  \end{subfigure}
  \caption{Overview of pool-level t-SNE visualization.}
  \label{fig:pool_level_t_sne_visualization}
\end{figure}

\textbf{Dataset Distribution Tools}. To better understand and analyze the structure of our dataset, we employ a t-SNE-based visualization technique to represent the distribution of price and technical indicators in a low-dimensional space. This approach reveals intrinsic patterns and temporal dynamics embedded in the data. The visualization is conducted in two distinct modes: asset-level and pool-level.

Asset-level visualization focuses on a few selected individual stocks. For each stock, we sample its price at different time points, and then apply t-SNE to project these temporal snapshots into a two-dimensional space. This mode allows us to observe how the price of a single asset evolve over time, and to compare such temporal trajectories across different individual assets.

Pool-level visualization, on the other hand, examines a large number of stocks within different asset pools (e.g., sectors or universes) at the same or selected time points. To construct these visualizations, we compute a comprehensive set of technical indicators using the Alpha158 feature set. The high-dimensional feature representations are then transformed via t-SNE into a more interpretable 2D layout, facilitating insight into the temporal and cross-sectional structure of the data. Here, we sample many stocks from each pool at specific times, and use t-SNE to project their high-dimensional features into two dimensions. This mode helps uncover broader distributional patterns across asset pools—such as clustering behavior, outlier structures, or inter-stock similarities within and across pools.

\subsection{Performance Visualization}

\textbf{Trading Curve Tools}. The proposed module replicates the trading-curve display of FinAgent~\cite{zhang2024multimodal} while focusing on a single, essential feature. It renders an interactive curve that tracks cumulative return and overlays each model decision with a clearly labelled BUY or SELL marker. An accompanying block of interpretable text records the reasoning behind every trade, summarising the key signals and risk constraints that guided the action. This streamlined presentation allows researchers to relate model logic directly to observed performance, offering a transparent basis for auditing algorithmic behaviour and evaluating how trading decisions drive the evolution of returns over time.

\begin{figure}[ht]
  \centering
  \includegraphics[width=0.9\textwidth]{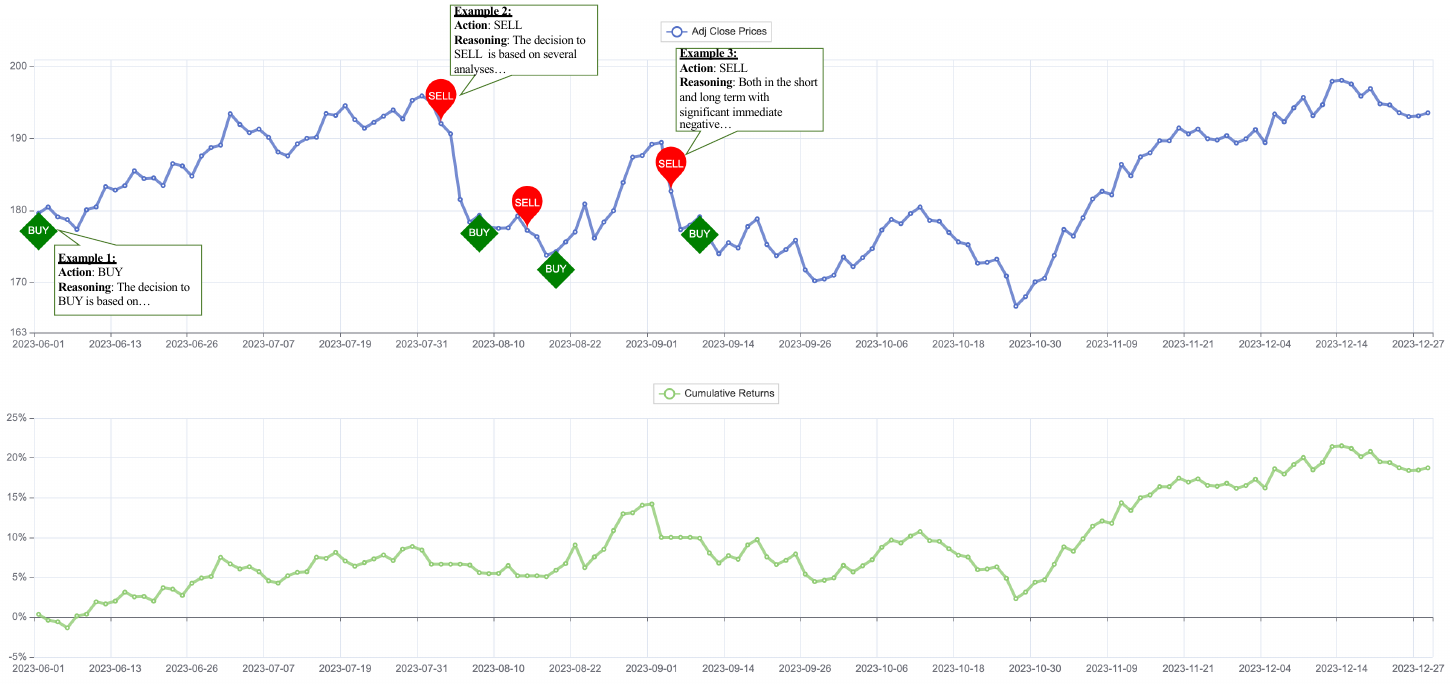}
  \caption{Overview of trading curve visualization.}
  \label{fig:trading_curve}
\end{figure}

\textbf{Score Visualization Tools}. As illustrated in Figure~\ref{fig:llm_reasoning}, the module renders an interactive bar chart that aggregates the performance of multiple large language models (LLMs) across a suite of reasoning benchmarks. Each bar encodes the absolute score of an individual model on a given task, while colour grouping facilitates quick visual association between benchmarks. By presenting all results in a single, normalised view, Score Visualization Tools makes it straightforward to identify relative strengths and weaknesses, rank models at a glance, and track progress as new architectures or training regimes are evaluated.

\textbf{Trading Performance Visualization Tools}. As illustrated in Figure \ref{fig:llm_trading}, the module presents an interactive radar chart that overlays multi metrics (e.g., ARR, SR, MDD, CR, SoR, VOL) for every evaluated model. Each polygon traces the scores of one model, and the enclosed area provides an intuitive proxy for holistic performance: the larger the area, the better the overall trade-off between return and risk. The radial format allows direct, dimension-by-dimension comparison, making it straightforward to pinpoint where a model excels or lags relative to its peers. Although the example focuses on trading-specific indicators, the framework is extensible; additional axes can represent intrinsic LLMs attributes such as reasoning abilities or computational efficiency, thereby enabling a unified visual assessment of both financial and cognitive capabilities.

\section{FinWorld as a Development Framework}

Beyond the four core financial AI tasks (time series forecasting, algorithmic trading, portfolio management, and LLM applications) discussed in the main text, \projectname is designed as a comprehensive and extensible development framework that supports unlimited expansion for diverse financial applications. The platform's modular architecture and flexible design principles enable researchers and practitioners to easily implement and experiment with various financial AI methodologies beyond the standard tasks.

The framework supports additional financial AI tasks such as factor model, financial market representation learning, and other domain-specific applications. For factor model, \projectname provides built-in support for traditional statistical factors, and deep learning factor extraction methods. The platform's representation learning capabilities enable the development of sophisticated market embeddings that capture complex financial relationships and temporal dynamics.

The extensible nature of \projectname is achieved through its layered architecture, where each component can be independently modified or extended without affecting other parts of the system. This design allows for seamless integration of new algorithms, models, and tasks while maintaining consistency with the existing framework. Researchers can leverage the platform's unified data interfaces, standardized evaluation metrics, and comprehensive visualization tools to develop and validate novel financial AI approaches.

\subsection{Factor Model}

\projectname provides comprehensive support for factor model, encompassing both traditional statistical approaches and advanced deep learning methodologies. For traditional statistical factors, we primarily utilize the Alpha158. For deep learning dynamic factor models, we implement state-of-the-art approaches including FactorVAE~\cite{duan2022factorvae} and HireVAE~\cite{wei2023hirevae}. FactorVAE leverages variational autoencoders to learn latent factor representations that capture complex market dynamics and temporal dependencies. HireVAE extends this framework with hierarchical structures to model multi-scale temporal patterns in financial data. Notably, the factor model STORM~\cite{zhao2024storm} is developed based on this framework, demonstrating the platform's capability to support cutting-edge research and development in financial AI.

The factor model framework in \projectname is designed with modularity and extensibility in mind, allowing researchers to easily implement and experiment with new factor extraction methods. The platform provides standardized interfaces for factor computation, evaluation, and backtesting, ensuring consistent and reproducible results across different factor models and datasets.

\subsection{Financial Market Representation Learning}

\projectname implements comprehensive financial market representation learning capabilities, focusing on stock pool-based representation learning approaches. The platform supports various state-of-the-art representation learning methods, including VAE (Variational Autoencoder)~\cite{vae}, MAE (Masked Autoencoder)~\cite{mae}, and VQVAE (Vector Quantized Variational Autoencoder)~\cite{vqvae} for capturing complex market dynamics and relationships. These representation learning methods enable the development of sophisticated market embeddings that capture both local and global patterns in financial data: VAE captures smooth state transitions via continuous latents, MAE learns robust features through masked reconstruction, and VQVAE models discrete regimes with categorical representations.

The representation learning framework in \projectname is designed to work seamlessly with different stock pools and market segments, allowing researchers to develop market-specific embeddings that capture the unique characteristics of different financial instruments and market conditions. This capability is particularly valuable for understanding cross-asset relationships, market regime detection, and developing more sophisticated financial AI models.

\section{Implementation Details}
\label{appx:implementation}

\subsection{Time Series Forecasting}
\label{appx:forecasting}

\textbf{Dataset Setup}. We focus on daily-level financial data for time series forecasting. We select DJ30 and SP500 from the U.S. market and SSE50 and HS300 from the Chinese market, covering both daily OHLCV data with Alpha158 technical indicators for the period 2015-05-01 to 2025-05-01, sourced from FMP. We employ a StandardScaler for feature normalization and utilize both dense features (145 dimensions from Alpha158 indicators) and sparse temporal features (4 dimensions: days, months, weekdays, years). The data is split at 2023-05-01 for training and validation separation. All features are standardized for each stock individually to ensure comparability across assets and time periods.

\textbf{Metrics}. We evaluate forecasting performance using four metrics: 
i) \textbf{Mean Absolute Error (MAE)}, which measures point prediction accuracy by averaging the absolute differences between predicted and true values; 
ii) \textbf{Mean Squared Error (MSE)}, which computes the average of the squared differences between predictions and actual outcomes; 
iii) \textbf{Rank Correlation Coefficient (RankIC)}, which assesses the correlation between the predicted and actual rankings using Spearman's rank correlation; and 
iv) \textbf{RankIC Information Ratio (RankICIR)}, which evaluates the stability of ranking performance over time by dividing the average RankIC by its standard deviation. 
MAE and MSE focus on absolute prediction accuracy, while RankIC and RankICIR are particularly relevant for evaluating the model’s effectiveness in capturing return-based or ranking-based financial relationships. Detailed information can be found in Table~\ref{appx:forecasting_metrics}.

\begin{table}[htb]
  \caption{Forecasting Metrics Overview}
  \label{appx:forecasting_metrics}
  \centering
  \footnotesize
  \setlength{\tabcolsep}{4pt}
  \renewcommand{\arraystretch}{1.0}
  \resizebox{0.98\textwidth}{!}{
  \begin{tabular}{lclp{0.45\textwidth}}
  \toprule
  \textbf{Name} & \textbf{Direction} & \textbf{Formula} & \textbf{Description} \\
  \midrule

  MAE & $\downarrow$ & $\mathrm{MAE} = \frac{1}{TN} \sum_{t=1}^{T} \sum_{i=1}^{N} \left| \hat{y}_{i,t} - y_{i,t} \right|$ 
      & Mean Absolute Error; average absolute deviation over all assets and time steps. \\

  MSE & $\downarrow$ & $\mathrm{MSE} = \frac{1}{TN} \sum_{t=1}^{T} \sum_{i=1}^{N} \left( \hat{y}_{i,t} - y_{i,t} \right)^2$
      & Mean Squared Error; penalizes large prediction errors more heavily. \\

  RankIC$_t$ & $\uparrow$ &
  $\displaystyle \mathrm{RankIC}_t = \frac{ \sum_{i=1}^{N} \left( R(\hat{y}_{i,t}) - \bar{R}_{\hat{y},t} \right)\left( R(y_{i,t}) - \bar{R}_{y,t} \right) }{ \sqrt{ \sum_{i=1}^{N} \left( R(\hat{y}_{i,t}) - \bar{R}_{\hat{y},t} \right)^2 } \sqrt{ \sum_{i=1}^{N} \left( R(y_{i,t}) - \bar{R}_{y,t} \right)^2 } }$ 
      & Spearman rank correlation computed across assets at each time step $t$. \\

  RankIC & $\uparrow$ & $\displaystyle \mathrm{RankIC} = \frac{1}{T} \sum_{t=1}^{T} \mathrm{RankIC}_t$ 
      & Average cross-sectional Spearman correlation over all time steps. \\

  RankICIR & $\uparrow$ &
  $\displaystyle \mathrm{RankICIR} = \frac{ \frac{1}{T} \sum_{t=1}^{T} \mathrm{RankIC}_t }{ \sqrt{ \frac{1}{T - 1} \sum_{t=1}^{T} \left( \mathrm{RankIC}_t - \bar{\rho} \right)^2 } }$
      & Rank Information Coefficient Information Ratio; mean RankIC divided by its standard deviation over time, with $\bar{\rho}$ being the mean RankIC. \\

  \bottomrule
  \end{tabular}
  }
\end{table}

\textbf{Methods}. We evaluate several state-of-the-art ML-based and DL-based time series forecasting models, including: i) ML-based methods: \textbf{LightGBM}~\cite{yang2020qlib} and \textbf{XGBoost}~\cite{yang2020qlib}; ii) DL-based methods: \textbf{Autoformer}~\cite{wu2021autoformer}, \textbf{Crossformer}~\cite{zhang2023crossformer}, \textbf{ETSformer}~\cite{woo2022etsformer}, \textbf{DLinear}~\cite{zeng2023transformers}, \textbf{TimesNet}~\cite{wu2022timesnet}, \textbf{PatchTST}~\cite{nie2022time}, \textbf{TimeMixer}~\cite{wang2024timemixer}, and \textbf{TimeXer}~\cite{wang2024timexer}. These models are selected based on their effectiveness in capturing complex financial time series patterns and their ability to handle high-dimensional data.

\textbf{Implementation Details}. All experiments are conducted on 2 NVIDIA H100 GPUs. Following the implementation practices of Qlib~\cite{yang2020qlib} and TSLib~\cite{tslib}, we adopt consistent architectural configurations across different models while preserving their unique characteristics.
For dataset, we utilize a multi-GPU parallel dataloader with a batch size of 64. Each batch contains sequences of 64 historical days and 32 future days, meaning the model observes the past 64 days and predicts the returns over the next 32 days. For network architecture, we use a standardized setup with encoder depth of 4 and decoder depth of 4, where the decoder typically consists of a single MLP layer for output mapping unless the model has specific decoder requirements. The embedding dimension is set to 128 across all models. For optimization, we employ AdamW optimizer with a learning rate of 1e-5 and implement a warmup scheduler for learning rate adjustment, with maximum training epochs of 2,000 and warmup epochs of 200. The loss function is Mean Squared Error (MSE). Additional model-specific configurations are applied based on each model's architectural requirements and design principles. The reported results are averaged over three runs with different random seeds. 

\subsection{Algorithmic Trading}
\label{appx:trading}

\textbf{Dataset Setup}. We focus on daily-level financial data for trading. We select AAPL, AMZN, GOOGL, META, MSFT, and TSLA from the U.S. market, covering both daily OHLCV data with Alpha158 technical indicators for the period from 1995-05-01 to 2025-05-01, sourced from FMP. We employ a StandardScaler for feature normalization and utilize both dense features (145 dimensions from Alpha158 indicators) and sparse temporal features (4 dimensions: days, months, weekdays, years). The data is split at 2023-05-01 for training and validation separation. All features are standardized for each stock individually to ensure comparability across assets and time periods.

\textbf{Metrics}. We evaluate algorithmic trading performance using six financial metrics:  
i) \textbf{Annual Rate of Return (ARR)}, which measures the annualized profitability of a strategy based on the change in portfolio value over time, adjusted by an annualization factor (e.g., 252 for daily trading);  
ii) \textbf{Sharpe Ratio (SR)}, which quantifies risk-adjusted return by comparing the average return to its standard deviation;  
iii) \textbf{Maximum Drawdown (MDD)}, which measures the largest peak-to-trough decline in the cumulative return, indicating the worst observed loss during a period;  
iv) \textbf{Calmar Ratio (CR)}, which evaluates the return-to-risk tradeoff by comparing average return to maximum drawdown;  
v) \textbf{Sortino Ratio (SoR)}, which is similar to the Sharpe Ratio but considers only downside volatility, thus emphasizing negative return risks.  
vi) \textbf{Volatility (VOL)}, which captures the standard deviation of returns and reflects the level of return fluctuation over time.

ARR reflects pure profitability, SR, CR, and SoR assess performance adjusted for different aspects of risk, and MDD and VOL evaluate risk exposure. Together, these metrics offer a comprehensive assessment of trading strategy effectiveness from both return and risk perspectives.

\begin{table}[htb]
  \caption{Trading Metrics Overview}
  \label{appx:trading_metrics}
  \centering
  \footnotesize
  \setlength{\tabcolsep}{4pt}
  \renewcommand{\arraystretch}{1.0}
  \resizebox{0.98\textwidth}{!}{
  \begin{tabular}{lclp{0.45\textwidth}}
  \toprule
  \textbf{Name} & \textbf{Direction} & \textbf{Formula} & \textbf{Description} \\
  \midrule

  ARR & $\uparrow$ & $\mathrm{ARR} = \left( \prod_{t=1}^{T} (1 + \mathrm{r}_t) \right)^{\frac{N}{T}} - 1$
      & Annualized Rate of Return; compounded growth rate over time, using $N$ periods per year and return series $\mathrm{rets}$. \\

  SR & $\uparrow$ & $\mathrm{SR} = \frac{\mathbb{E}[\mathrm{rets}] - r_f}{\mathrm{Std}(\mathrm{rets})} \times \sqrt{N}$
      & Sharpe Ratio; annualized excess return per unit of total volatility, computed from return series $\mathrm{rets}$. \\

  MDD & $\downarrow$ & $\mathrm{MDD} = \max_{t \in [1,T]} \left( \frac{\max_{s \in [1,t]} V_s - V_t}{\max_{s \in [1,t]} V_s} \right)$
  & Maximum Drawdown; largest observed decline from peak to trough in cumulative portfolio value derived from $\mathrm{rets}$. \\

  CR & $\uparrow$ & $\mathrm{CR} = \frac{\mathrm{ARR}(\mathrm{rets})}{|\mathrm{MDD}(\mathrm{rets})|}$
      & Calmar Ratio; annualized return divided by absolute maximum drawdown, both computed from $\mathrm{rets}$. \\    

  SoR & $\uparrow$ & $\mathrm{SoR} = \frac{\mathbb{E}[\mathrm{rets}] - r_f}{\mathrm{DD}(\mathrm{rets})} \times \sqrt{N}$
      & Sortino Ratio; risk-adjusted return using downside deviation, penalizing only negative returns in $\mathrm{rets}$. \\

  VOL & $\downarrow$ & $\mathrm{VOL} = \mathrm{Std}(\mathrm{rets}) \times \sqrt{N}$
      & Volatility; annualized standard deviation of return series $\mathrm{rets}$, capturing total return variability. \\

  DD & $\downarrow$ & $\mathrm{DD} = \sqrt{ \frac{1}{T} \sum_{t=1}^{T} \min(\mathrm{r}_t - r_f, 0)^2 } \times \sqrt{N}$
      & Downside Deviation; annualized standard deviation of returns below the risk-free rate, focusing on downside risk. \\
  
  \midrule
  \multicolumn{4}{l}{\textit{Note:} $\mathrm{rets} = [r_1, r_2, \ldots, r_T]$ is the return series, where $r_t$ is the return on day $t$; $N$ is the number of periods per year; $r_f$ is the risk-free rate.} \\
  \bottomrule
  \end{tabular}
  }
\end{table}

\textbf{Methods}. We systematically evaluate several Rule-based, ML-based, DL-based, and RL-based methods, including i) Rule-based methods: \textbf{BUY\&HOLD}, \textbf{MACD}; ii) ML-based Methods: \textbf{LightGBM}, \textbf{XGBoost}; iii) DL-based Methods: \textbf{Transformer}~\cite{vaswani2017attention}, \textbf{LSTM}, \textbf{DLinear}~\cite{zeng2023transformers};  iv) RL-based Methods: \textbf{PPO}~\cite{schulman2017proximal}, \textbf{SAC}~\cite{haarnoja2018soft}. These models represent a diverse set of approaches commonly used in quantitative finance research and practice.

\textbf{Implementation Details}. All experiments are conducted on 2 NVIDIA H100 GPUs. For all DL-based methods, we use similar dataset splits, network architectures, and optimizer configurations as in the forecasting task. For RL-based methods, the initial cash in the environment is set to $1 \times 10^5$, and the transaction fee rate is $1 \times 10^{-4}$. We use 4 parallel environments for data sampling, with a sampling step size of 512. The policy mini-batch size is set to 128 and the value mini-batch size to 16. The optimizer used is AdamW, with the policy network learning rate set to $1 \times 10^{-5}$ and the value network learning rate set to $1 \times 10^{-6}$. The maximum number of training steps is $1 \times 10^8$, and the learning rate warmup is applied starting from step 0. Additional model-specific configurations are applied based on each model's architectural requirements and design principles. The reported results are averaged over three runs with different random seeds. 

\subsection{Portfolio Management}
\label{appx:portfolio}

\textbf{Dataset Setup}. The dataset setup is the same as in the time series forecasting task.

\textbf{Metrics}. The evaluation metrics for portfolio management are consistent with those used in the algorithmic trading task. Specifically, we report \textbf{ARR}, \textbf{SR}, \textbf{MDD}, \textbf{CR}, \textbf{SoR}, and \textbf{VOL}, following the definitions and formulas described in the previous section.

\textbf{Methods}. We systematically evaluate several ML-based, DL-based, and RL-based methods. Since we can apply the top-$k$ dropout strategy ~\cite{yang2020qlib} after any ML\&DL-based forecasting model to construct a portfolio, our ML\&DL-based methods include all ML\&DL-based forecasting models. Specifically, we consider: i) Rule-based methods: \textbf{BUY\&HOLD}; ii) ML-based Methods: \textbf{LightGBM}, \textbf{XGBoost}; iii) DL-based Methods: \textbf{Autoformer}~\cite{wu2021autoformer}, \textbf{Crossformer}~\cite{zhang2023crossformer}, \textbf{ETSformer}~\cite{woo2022etsformer}, \textbf{DLinear}~\cite{zeng2023transformers}, \textbf{TimesNet}~\cite{wu2022timesnet}, \textbf{PatchTST}~\cite{nie2022time}, \textbf{TimeMixer}~\cite{wang2024timemixer}, and \textbf{TimeXer}~\cite{wang2024timexer}; iv) RL-based Methods: \textbf{PPO}~\cite{schulman2017proximal}, \textbf{SAC}~\cite{haarnoja2018soft}. These models represent a diverse set of approaches commonly used in quantitative finance research and practice.

\textbf{Implementation Details}. All experiments are conducted on 2 NVIDIA H100 GPUs. For all DL-based methods, we use similar dataset splits, network architectures, and optimizer configurations as in the forecasting task. In addition, all ML\&DL-based methods use the Top-$k$ Dropout strategy~\cite{yang2020qlib} for portfolio construction. For RL-based methods, the initial cash in the environment is set to $1 \times 10^5$, and the transaction fee rate is $1 \times 10^{-4}$. We use 4 parallel environments for data sampling, with a sampling step size of 512. The policy mini-batch size is set to 128 and the value mini-batch size to 16. The optimizer used is AdamW, with the policy network learning rate set to $1 \times 10^{-5}$ and the value network learning rate set to $1 \times 10^{-6}$. The maximum number of training steps is $1 \times 10^8$, and the learning rate warmup is applied starting from step 0. Additional model-specific configurations are applied based on each model's architectural requirements and design principles.The reported results are averaged over three runs with different random seeds. 

\subsection{LLM Applications} 

Given that LLM Applications encompass both the training and deployment of LLMs as well as LLM Agents, we have allocated a dedicated chapter to this subject. Further details can be found in Appendix \ref{appx:llm_applications}.

\section{LLM Applications}
\label{appx:llm_applications}

In this section, we provide a comprehensive overview of the development background of both LLMs and LLM agents in financial AI. We discuss the historical context, the key advancements that have shaped their evolution, and the main challenges encountered along the way. Furthermore, we elaborate on several specific implementation details within our own work, illustrating how these developments have informed and influenced our approach. This in-depth analysis aims to offer readers a deeper understanding of the foundations and practical considerations involved in working with LLMs and LLM agents.

\subsection{Background}

\textbf{LLMs for Financial AI.}
The integration of LLMs into financial decision-making processes has fundamentally transformed the landscape of financial AI, giving rise to two predominant developmental trajectories. The first, traditional paradigm leverages large-scale pre-training on generic corpora, followed by supervised fine-tuning (SFT) on domain-specific datasets, a process that has enabled the creation of financial-domain LLMs such as FinBERT~\cite{araci2019finbert}, FLANG~\cite{shah2022flang}, and FinGPT~\cite{liu2023fingpt}. These models excel at classic financial NLP tasks including sentiment analysis, news classification, and financial document understanding, benefiting from the inclusion of specialized financial terminology and context during pre-training. For example, BloombergGPT~\cite{wu2023bloomberggpt} introduced domain-adapted tokenization and large-scale financial corpus curation, enabling robust out-of-domain generalization while maintaining strong in-domain accuracy on a wide range of financial tasks. Similarly, FinQA~\cite{chen2021finqa} targets numerical reasoning within financial texts, highlighting the importance of reasoning abilities in addition to conventional text understanding.

The second, more recent development path centers on enhancing the reasoning capabilities of LLMs using RL, inspired by breakthroughs such as DeepSeek-R1~\cite{guo2025deepseek}, which demonstrated substantial improvements in complex reasoning and decision-making tasks. In the context of financial AI, RL-based approaches have facilitated the creation of models capable of more sophisticated financial reasoning, including complex question answering (QA) and multi-step analysis over noisy and unstructured market data. State-of-the-art systems such as Fin-R1~\cite{liu2025fin}, Fin-o1~\cite{qian2025fino1}, and Dianjin-R1~\cite{zhu2025dianjin} typically adopt a two-stage pipeline: they first apply SFT on curated financial QA datasets to capture domain knowledge, then employ RL to further optimize the models’ reasoning and generalization abilities with task-specific reward functions on challenging financial benchmarks. Notably, this line of work has shown that RL fine-tuning can help mitigate some limitations of SFT-only models, such as shallow pattern matching or overfitting to specific data distributions. However, it is also evident that LLM-based methods remain challenged by issues such as the absence of explicit sequential decision-making mechanisms, high computational overhead (particularly with RL-based training), and an inability to robustly handle the non-stationarity and regime shifts that typify real-world financial markets. Moreover, most current benchmarks still emphasize static QA or single-step predictions, rather than end-to-end financial decision-making in dynamic environments.

\textbf{LLM Agents for Financial AI.}
To overcome the intrinsic limitations of static LLM-based models in handling sequential and interactive decision-making tasks, recent research has shifted towards constructing LLM-powered financial agents equipped with advanced agentic mechanisms. These mechanisms, such as external memory, dynamic tool use, self-reflection, and collaborative reasoning, enable LLMs to act not just as passive information processors, but as proactive, context-aware agents capable of adaptive and continual learning. Early efforts, including FinMem~\cite{yu2024finmem} and FinRobot~\cite{zhou2024finrobot}, demonstrated that the addition of structured memory modules and chain-of-thought reasoning could substantially boost single-asset trading performance and stability, particularly in volatile or data-scarce settings. These systems typically maintain layered or episodic memories that allow agents to recall past market events, dynamically update strategies, and profile assets or markets over time.

Building on these foundations, more recent agentic systems such as FinAgent~\cite{zhang2024multimodal} have introduced multimodal reasoning capabilities, allowing LLMs to jointly process textual news, structured numerical data, and even K-line (candlestick) chart images via tool-augmented and dual-reflection architectures. This multimodal approach is shown to yield substantial profit gains and improved generalization across diverse datasets. Meanwhile, there has been a pronounced trend toward collaborative and hierarchical multi-agent systems, inspired by real-world financial organizations. For instance, FinCon~\cite{yu2024fincon} adopts a manager–analyst architecture, combining conceptual verbal reinforcement and episodic self-critique to facilitate robust sequential decision-making in both single-stock and multi-asset (portfolio) trading. Similarly, TradingAgents~\cite{xiao2024tradingagents} simulate an entire trading firm with specialized agents (e.g., fundamental, technical, sentiment, and risk analysts), whose structured debate and coordination have been shown to improve both returns and risk control.

Recent advances have further explored the combination of RL with agentic LLMs to enable dynamic exploration and continual adaptation in non-stationary environments~\cite{feng2025group,agent-r1}. For example, group-based RL finetuning enables multiple LLM agents to interact, learn, and adapt strategies in simulated market environments, fostering resilience against sudden market changes. Beyond performance, there is an increasing focus on interpretability and safety in financial LLM agents, with research exploring techniques such as role assignment, self-consistency checking, and real-time human-in-the-loop supervision. Despite these advances, key open challenges remain: efficiently scaling agentic LLMs, mitigating hallucinations and overfitting, improving sample efficiency in RL settings, and reliably deploying such systems in the face of adversarial and ever-evolving market conditions.

In summary, based on the development trajectory of LLMs and LLM agents described above, we can categorize the most popular LLM applications in the financial sector into four main types of tasks. Three of them require training, including \textbf{SFT for LLMs}, \textbf{RL for LLMs}, and \textbf{RL for LLM Agents}. The fourth type does not require training and mainly relies on prompt engineering for \textbf{LLM Agents} to perform downstream tasks. In \projectname, the primary focus is on two types of tasks: \textbf{RL for LLMs} and \textbf{LLMs Agents}.

\subsection{Reinforcement Learning for LLMs}
\label{appx:llm_reasoning}

The reinforcement learning implementation for LLMs in \projectname follows a two-stage training paradigm designed to progressively develop both reasoning capabilities and decision-making skills in financial contexts.

\textbf{Stage I: Financial Reasoning Fine-tuning.} In the first stage, LLMs undergo reinforcement learning fine-tuning on conventional financial LLM reasoning datasets (refer to the LLM reasoning dataset section~\ref{appx:llm_reasoning_dataset}). This stage focuses on developing fundamental financial reasoning abilities, including understanding financial concepts, analyzing market data, and performing quantitative calculations. The model learns to generate coherent and accurate responses to financial questions, interpret complex financial scenarios, and apply domain-specific knowledge. However, this stage alone is insufficient for real-world financial decision-making applications, as it lacks exposure to the dynamic and uncertain nature of actual market environments.

\textbf{Stage II: Market Environment Learning.} The second stage addresses this limitation by immersing the LLM in real market environments (refer to the market dataset section~\ref{appx:market_dataset}) for exploration, trial-and-error learning, and decision-making skill development. In this stage, the model learns to make sequential decisions under uncertainty, adapt to changing market conditions, and optimize for long-term financial outcomes. The model interacts with simulated or real market data, receiving feedback on its decisions through reward signals that reflect financial performance metrics such as returns, risk-adjusted performance, and transaction costs. This stage enables the LLM to develop practical decision-making capabilities that go beyond theoretical reasoning, preparing it for deployment in actual financial applications. We primarily focus on the single asset trading task in this stage.

The two-stage approach ensures that LLMs first establish a solid foundation in financial reasoning before developing the practical skills necessary for effective decision-making in dynamic market environments. This progressive learning strategy enhances both the theoretical understanding and practical applicability of LLM-based financial AI systems.

\noindent\textbf{Stage I: Financial Reasoning Fine-tuning}

\subsubsection{Training Dataset.} The training dataset for Stage I is used to train the LLM to reason about financial questions. We use the financial reasoning dataset provided by the LLM reasoning dataset section~\ref{appx:llm_reasoning_dataset}. We collected over 80k training samples to ensure comprehensive coverage of financial reasoning patterns and scenarios. We evaluate our model on test datasets from FinQA, FinEval, ConvFinQA, and CFLUE to assess its reasoning capabilities across diverse financial domains. Since our dataset contains both Chinese and English content, we follow conventional practices by using Chinese prompts for Chinese data and English prompts for English data templates.

The templates of the prompts for English and Chinese are as follows:

\begin{adjustwidth}{1em}{0pt}
  \label{appx:stage_1_prompt_template}
  \begin{tcolorbox}[
      colback=myblue!80!white, 
      colframe=blue!50!black, 
      arc=4mm, 
      boxrule=0.8pt, 
      left=1mm, right=1mm, top=1mm, bottom=1mm,
      width=\textwidth
      ]
      
      \centering
      {\bfseries English Prompt Template}

      {You FIRST think about the reasoning process as an internal monologue and then provide the final answer. The reasoning process MUST BE enclosed within <think> </think> tags. The final answer MUST BE put in \textbackslash boxed\{\}.
      }

      \centering
      {\bfseries Chinese Prompt Template}

      {
        \begin{CJK}{UTF8}{gbsn} 
        你首先将思考过程作为内部独白，然后给出最终答案。推理过程必须用<think> </think>标签括起来。最终答案必须放在\textbackslash boxed\{\}中。
        \end{CJK}
      }

  \end{tcolorbox}
\end{adjustwidth}

\subsubsection{Training Procedure.} For our financial reasoning tasks, we employ Group Relative Policy Optimization (GRPO)~\cite{shao2024deepseekmath} as the primary training methodology. GRPO is specifically designed for training large language models in reasoning-intensive scenarios, making it particularly well-suited for financial question answering and document analysis tasks. The algorithm enhances the model's capability to generate accurate and well-reasoned financial responses through group-based reward normalization and policy optimization.

In our implementation, GRPO samples groups of candidate outputs for each financial task instance and computes group-normalized advantages,
{
\begin{equation}
    \hat{A}_{i, t} = \frac{r_i - \mathrm{mean}(\{R_j\}_{j=1}^G)}{\mathrm{std}(\{R_j\}_{j=1}^G)},
\end{equation}
}
and optimizes a clipped objective:
{
\begin{equation}
\begin{split}
\mathcal{J}_{\mathrm{GRPO}}(\theta) =\; & \mathbb{E}_{(q,a)\sim\mathcal{D},\, \{o_i\}_{i=1}^G \sim \pi_{\theta_{\mathrm{old}}}} \Bigg[
    \frac{1}{G} \sum_{i=1}^G \frac{1}{|o_i|} \sum_{t=1}^{|o_i|} \Big(
        \min \Big( r_{i,t}(\theta) \hat{A}_{i,t},\, \\ 
        & \mathrm{clip}\left(r_{i,t}(\theta), 1-\epsilon, 1+\epsilon\right) \hat{A}_{i,t} \Big) \qquad - \beta D_{\mathrm{KL}}(\pi_\theta \| \pi_{\mathrm{ref}})
    \Big) 
\Bigg],
\end{split}
\end{equation}
} where $r_{i,t}(\theta) = \frac{\pi_\theta(o_{i,t} \mid q, o_{i, < t})}{\pi_{\theta_{\mathrm{old}}}(o_{i,t} \mid q, o_{i, < t})}$ is the importance sampling ratio, and $\epsilon$ is the clipping parameter. Following the approach in DAPO~\cite{yu2025dapo}, we omit the KL divergence regularization term as it has been shown to be unnecessary for chain-of-thought reasoning scenarios.

Following the design of DeepSeek-R1~\cite{guo2025deepseek}, our reward function consists of two main components: format reward and accuracy reward, designed to encourage both proper reasoning structure and correct answers.

\textbf{Format Reward.} The format reward encourages the model to follow a specific output structure with reasoning steps enclosed in \texttt{<think>...</think>} tags and the final answer in \texttt{\textbackslash boxed\{\}} tag. The format reward is defined as:

\begin{equation}
R_{\text{f}}(y) = \begin{cases}
1, & \text{if format matches} \\
0, & \text{otherwise}
\end{cases}
\end{equation}

where $y$ denotes the model's output. Format matching requires the output to contain exactly one pair of \texttt{<think>} tags and one \texttt{\textbackslash boxed\{\}} tag, with no additional content outside these tags.

\textbf{Accuracy Reward.} Given the complexity of financial scenarios where exhaustive enumeration of answer patterns through rule-based methods is challenging, we adopt Qwen3~\cite{yang2025qwen3} as the judge for answer evaluation. The content within the \texttt{\textbackslash boxed\{\}} tag is extracted using regular expressions and serves as the solution. The accuracy reward is defined as:

\begin{equation}
R_{\text{a}}(y, y^*) = \begin{cases}
1, & \text{if } y = y^* \\
0, & \text{otherwise}
\end{cases}
\end{equation}

where $y$ is the model's output (extracted from \texttt{\textbackslash boxed\{\}} tag) and $y^*$ is the standard answer. In our specific implementation, since financial problems are mostly multiple-choice questions, multi-select questions, financial calculation problems, and QA questions, we have implemented a set of regular expression matching rules to determine answer correctness based on the answer type. 

The final composite reward is computed as:

\begin{equation}
R = \alpha \cdot R_{\text{f}}(y) + \beta \cdot R_{\text{a}}(y, y^*)
\end{equation}

where $\alpha$ and $\beta$ are weighting parameters (typically $\alpha = 0.1$, $\beta = 0.9$) to balance format compliance and answer correctness.

\textbf{Metrics.} For evaluating the performance of our financial reasoning models, we employ \textbf{Score} (accuracy score) as the primary metric. The score measures the proportion of correctly answered questions in the test dataset and is defined as:

\begin{equation}
\text{Score} = \frac{N_{\text{correct}}}{N_{\text{total}}} \times 100\%
\end{equation}

where $N_{\text{correct}}$ represents the number of questions answered correctly and $N_{\text{total}}$ denotes the total number of questions in the evaluation dataset. A question is considered correctly answered if the model's output within the \texttt{\textbackslash boxed\{\}} tags matches the standard answer according to our regular expression matching rules.

\textbf{Methods}. We evaluate our model \textbf{FinReasoner} against open-source LLMs on the test dataset. We systematically evaluate several open-source LLM methods, including \textbf{DeepSeek-R1}~\cite{guo2025deepseek}, \textbf{Qwen3-8B}~\cite{yang2025qwen3}, \textbf{Fin-R1-7B}~\cite{guo2025deepseek}, and \textbf{Qwen2.5-7B-Instruct}~\cite{qwen2025qwen25technicalreport}. We use the same evaluation protocol as the original paper to ensure fair comparison.

\textbf{Implementation Details.} Our model, \textbf{FinReasoner}, is built on the \textbf{Qwen-8B} language model as the base LLM. In our implementation, we integrate Verl 0.5.0~\cite{verl} as an external dependency library within the \texttt{libs} directory. We customize both the trainer and the reward computation as plugins, which are injected into Verl’s trainer (details can be found in the \texttt{mverl} module of our codebase). This implementation approach offers significant advantages: since Verl is frequently updated, the plugin injection method provides greater flexibility for maintaining synchronization with the official Verl library.

All experiments are conducted on 16 NVIDIA A100 GPUs distributed across 2 nodes with 8 GPUs per node. We employ tensor model parallelism with a parallel size of 4 to efficiently handle large model training. The training configuration includes a batch size of 64, with micro batch sizes of 16 per GPU for both PPO training and log probability computation. We set the maximum prompt length and response length to 4096 tokens each, with left truncation for input sequences. For GRPO training, we use a rollout size of 8 and train for 50 epochs. The learning rate is set to 1e-6 for stable convergence during the reinforcement learning phase.

\noindent\textbf{Stage II: Market Environment Learning}

\subsubsection{Training Dataset.} 

\begin{figure}[ht]
  \centering

  \begin{subfigure}[b]{0.45\textwidth}
    \centering
    \includegraphics[width=\textwidth]{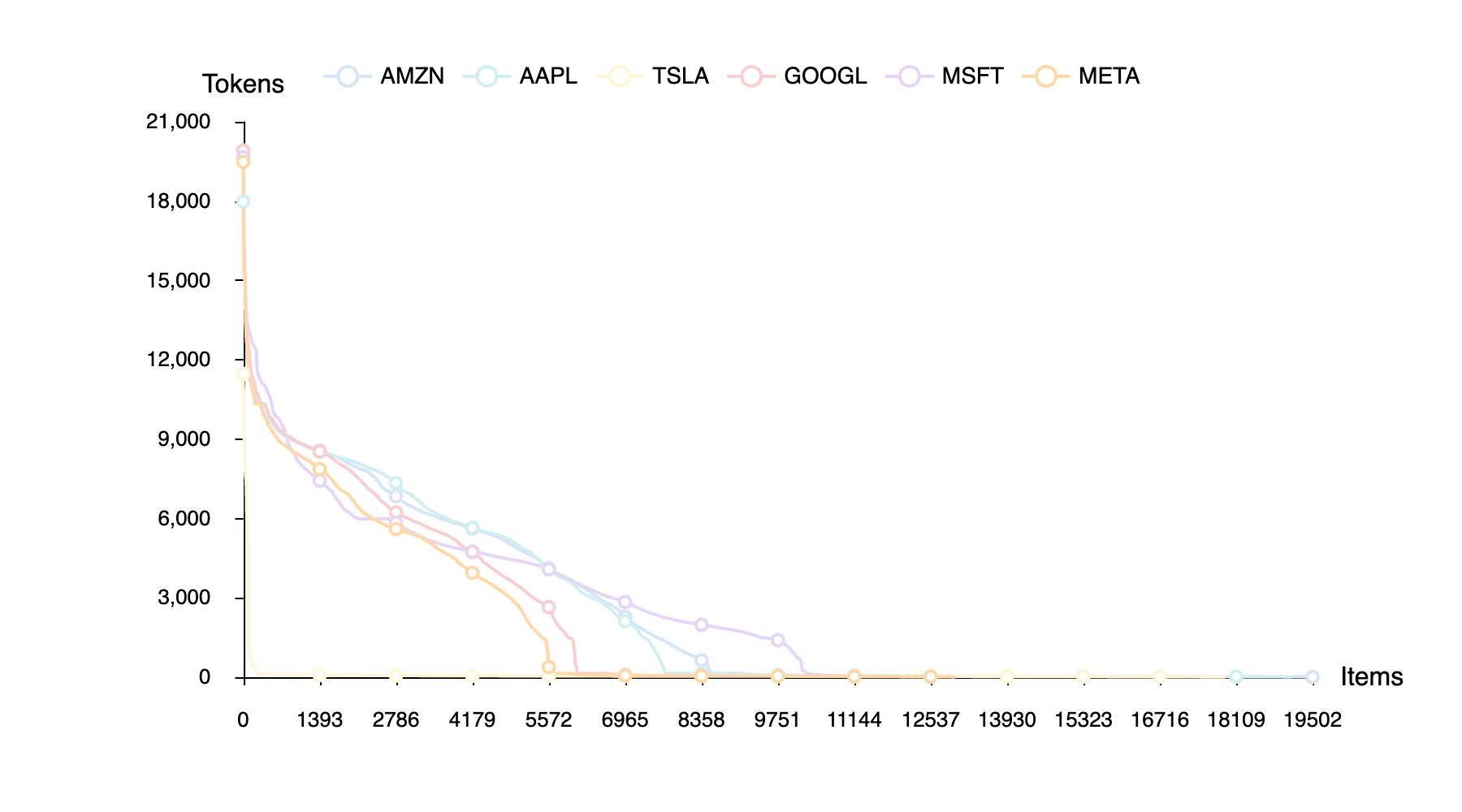}
    \caption{Original news tokens distribution.}
    \label{fig:original_tokens_distribution}
  \end{subfigure}%
  \begin{subfigure}[b]{0.45\textwidth}
    \centering
    \includegraphics[width=\textwidth]{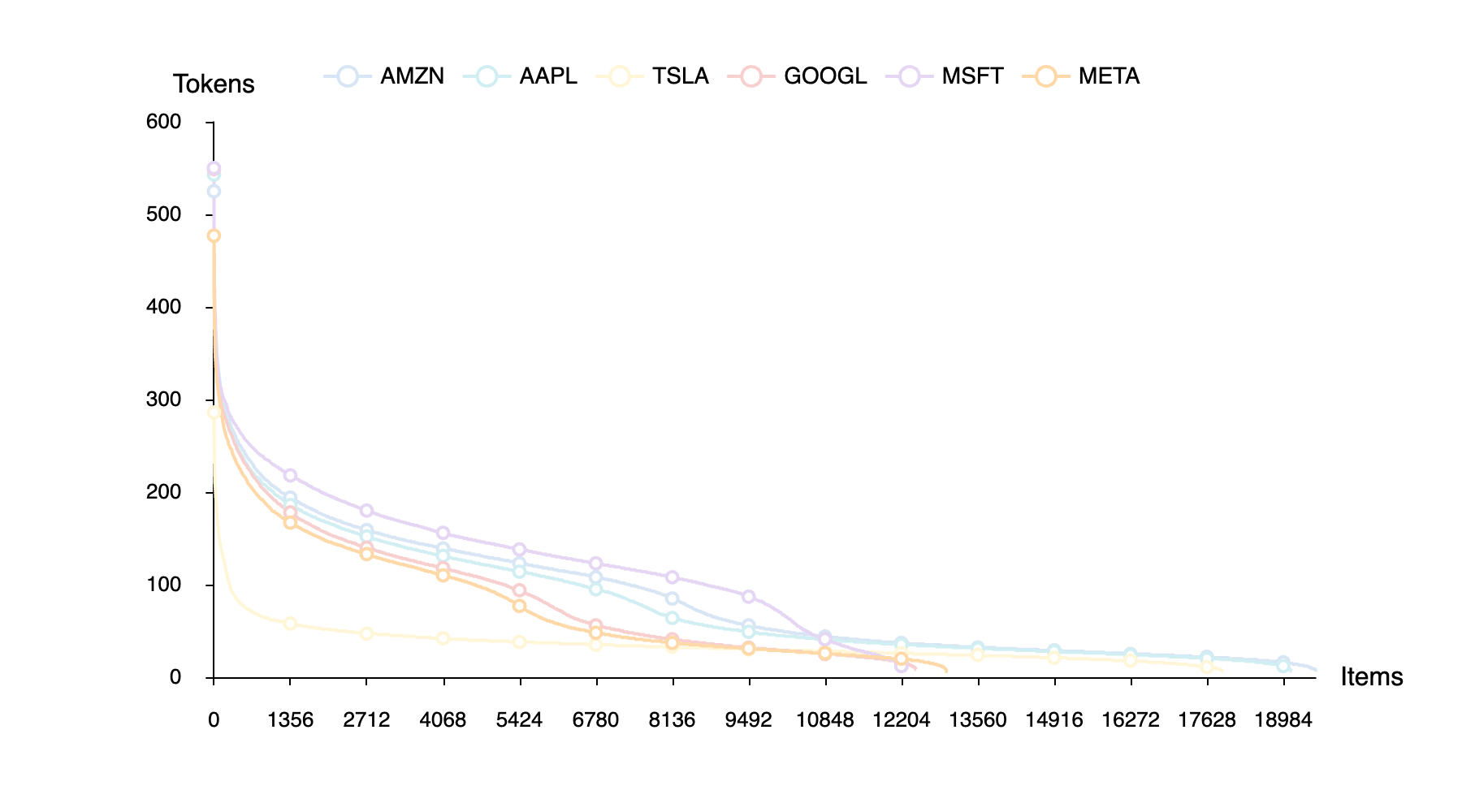}
    \caption{Summarized news tokens distribution.}
    \label{fig:summarized_tokens_distribution}
  \end{subfigure}

  \caption{News tokens distribution before and after summarization.}
  \label{fig:tokens_distribution}
\end{figure}

For Stage II training, we employ the market dataset, where we use prompt templates to provide historical OHLCV data and news text for LLM analysis and reasoning. We select 6 US stocks (AAPL, AMZN, GOOGL, META, MSFT, and TSLA) with nearly 10 years of data from 2015-05-01 to 2025-05-01. In the data processing pipeline, we first download OHLCV and news data from FMP. However, since the news text provided by FMP contains some advertisements and irrelevant content, we employ web crawling tools to scrape original news webpages, then filter out extremely long and short texts. Subsequently, we use Qwen3-32B to summarize and extract effective textual information from each news webpage, and finally use the summary as part of the news content in our prompts. The distribution of news tokens before and after summarization is shown in Figure~\ref{fig:tokens_distribution}.

In our specific implementation, we use historical 7-day OHLCV data and news as input. Since there are many news texts per day, we sample 3-5 news articles per day and organize historical prices, news, and historical effective decisions (BUY, SELL) records into prompts as LLM input. The prompt structure is as follows:

\begin{adjustwidth}{1em}{0pt}
  \label{appx:stage_2_prompt_template}
  \begin{tcolorbox}[
      colback=myblue!80!white, 
      colframe=blue!50!black, 
      arc=4mm, 
      boxrule=0.8pt, 
      left=1mm, right=1mm, top=1mm, bottom=1mm,
      width=\textwidth
      ]
      
      \centering
      {\bfseries Trading Prompt Template}
      
      \begin{adjustwidth}{0pt}{0pt}
      \begin{lstlisting}[
        basicstyle=\footnotesize,
        breaklines=true,
        breakatwhitespace=true,
        columns=flexible,
        frame=none,
        backgroundcolor=\color{myblue!80!white},
        lineskip=-2pt
      ]
You are a highly experienced trader. Your task is to carefully analyze the provided financial news, current price, and historical trading records. Based on your professional judgment, provide the single best trading decision for the current situation.

Here is the task:
# Name: Apple Inc., Symbol: (AAPL)

## Price (7 days OHLCV data)
|   close |   high |    low |   open |      volume |
|--------:|-------:|-------:|-------:|------------:|
|  173.56 | 174.03 | 171.9  | 173.02 | 5.37245e+07 | ...

## News (3-5 news articles)
**Timestamp | Title | Content**
2023-05-10 15:39:19 | Unleash Your iPhone's Inner Energizer Bunny: Battery Life Tips For The Tech-Savvy | One of the most common concerns among **Apple Inc** ...

## Historical Trading Records (7 days historical trading records)
|   Timestamp |   Decision |   Price |   Volume |
|------------:|-----------:|--------:|---------:|
| 2023-05-01 09:30:00 | BUY | 173.56 | 5.37245e+07 | ...

## Record
| timestamp   |   open |   high |    low |   close |      volume |   price |        cash |   position |   pre_value | action   |   post_value |           ret |
|:------------|-------:|-------:|-------:|--------:|------------:|--------:|------------...
| 2023-04-25  | 165.19 | 166.31 | 163.73 |  163.77 | 4.87141e+07 |  169.59 | 100000...

## History Valid Action
| timestamp   |   open |   high |    low |   close |      volume |   price |        cash |   position |   pre_value | action   |   post_value |           ret |
|:------------|-------:|-------:|-------:|--------:|------------:|--------:|------------...
| 2023-05-05  | 170.98 | 174.3  | 170.76 |  173.57 | 1.13453e+08 |  173.57 | 100000...

## Note
1. `timestamp`: the timestamp of the record
2. `open`: Open price
3. `high`: High price
4. `low`: Low price
5. `close`: Close price
6. `volume`: Volume of the asset traded
7. `price`: Current price (adj_close price)
8. `cash`: Current cash
9. `position`: Current position
10. `pre_value`: Previous total value, `value = cash + position * price`
11. `action`: Action taken, `BUY`, `SELL`, or `HOLD`
12. `post_value`: Current total value
13. `ret`: Return, `ret = (post_value - pre_value) / pre_value`

Today is 2023-05-16 23:59:59, and the current price, cash, and position are 172.07, 98359.28, and 0000.

You should follow the instructions below in detail when making your decision.
  
1. Your full reasoning process must be enclosed within <think></think> tags.
2. Your final answer must be one of the following three options: BUY, HOLD, or SELL, and be presented only inside a single \boxed{}.
3. Do not output anything else except the reasoning in <think>...</think> and the final answer in \boxed{}.

Example output:
<think>Based on the news indicating strong earnings, a rising price trend, and positive historical returns, a BUY decision is justified.</think>
\boxed{BUY}

      \end{lstlisting}
      \end{adjustwidth}

  \end{tcolorbox}
\end{adjustwidth}

\subsubsection{Training Procedure.} 

Following the design principles of RAGEN~\cite{ragen}, our reward function in the single-stock trading environment consists of two main components: format reward and trading reward, which are designed to ensure standardized action outputs and effective trading decisions.

\textbf{Format Reward.}
The format reward encourages the model to adhere to a strict action output format. For each trading action $a$, the format reward is defined as:
\begin{equation}
R_{\mathrm{f}}(a) = 
\begin{cases}
1, & \text{if the action format is correct} \\
0, & \text{otherwise}
\end{cases}
\end{equation}
where the action output must specify a valid trading operation (e.g., buy, sell, or hold) and the corresponding quantity, without any extraneous or malformed content.

\textbf{Trading Reward.}
The trading reward evaluates the profit and loss generated by the model's trading decisions over a trajectory of $T$ trading days. It is computed as the relative change in the agent’s portfolio value, taking transaction costs into account:
\begin{equation}
R_{\mathrm{t}}(\tau) = \frac{V_T - V_0}{V_0}
\end{equation}
where $V_t$ is the portfolio value at time $t$:
\begin{equation}
V_t = \text{Cash}_t + \text{Position}_t \times \text{Price}_t
\end{equation}
Here, $\text{Position}_t$ is the number of shares held at time $t$, and $\text{Price}_t$ is the stock price at time $t$. At each trading step, a transaction cost proportional to the traded volume is deducted:
\begin{equation}
C_t = \lambda |\Delta \text{Position}_t| \times \text{Price}_t
\end{equation}
\begin{equation}
V_t \leftarrow V_t - C_t
\end{equation}
where $\lambda$ is the transaction fee rate (e.g., $1 \times 10^{-4}$).

The final reward for each trajectory is a weighted combination of the format and trading rewards:
\begin{equation}
R = \gamma R_{\mathrm{f}}(a) + (1-\gamma) R_{\mathrm{t}}(\tau)
\end{equation}
where $\gamma$ is a hyperparameter (e.g., $\gamma=0.1$) balancing the importance of action format and trading performance.

\textbf{Metrics.} We evaluate the performance of the trained model using the following trading metrics: \textbf{ARR}, \textbf{SR}, \textbf{MDD}, \textbf{CR}, \textbf{SoR}, and \textbf{VOL}.

\textbf{Methods}. We evaluate our trading models using a simplified FinAgent-based AI Agent framework developed in this work. This framework supports both commercial models (e.g., GPT-4.1, Claude-4-Sonnet) and local models (e.g., Qwen), allowing the underlying LLM to be flexibly replaced. Therefore, our comparison includes \textbf{DeepSeek-R1}~\cite{guo2025deepseek}, \textbf{Qwen3-8B}~\cite{yang2025qwen3}, \textbf{Fin-R1-7B}~\cite{guo2025deepseek}, \textbf{Qwen2.5-7B-Instruct}~\cite{qwen2025qwen25technicalreport}, \textbf{GPT-4.1}, \textbf{Claude-4-Sonnet}, as well as our own \textbf{FinReasoner}, all serving as the backbone LLMs for the FinAgent. It is worth noting that a detailed description of our FinAgent (LLMs Agents) implementation will be provided in the next section; please refer to Appendix~\ref{appx:llms_agents} for details.

\textbf{Implementation Details}. Similar to Stage I, we implement this using Verl and external plugins to inject additional trainers and evaluation functions. Notably, we also need to inject an environment implementation to enable dynamic interaction between the LLMs and the environment.

All experiments are conducted on 16 NVIDIA A100 GPUs distributed across 2 nodes with 8 GPUs per node. We employ tensor model parallelism with a parallel size of 4 to efficiently handle large model training. The training configuration includes a batch size of 64, with micro batch sizes of 16 per GPU for both PPO training and log probability computation. We set the maximum prompt length and response length to 8192 tokens each, with left truncation for input sequences. For GRPO training, we use a rollout size of 8 and train for 50 epochs. The learning rate is set to 1e-6 for stable convergence during the reinforcement learning phase. 

\subsection{LLMs Agents for Financial AI}
\label{appx:llms_agents}

Before introducing financial AI agents, it is important to first accept the concept of a Tool-Calling Agent. In the early stages of agent development, action-based agents played a central role. Notable examples include SayCan~\cite{ahn2022can} and ReAct~\cite{yao2023react}, which are built around action selection: they interpret high-level instructions via a language model and choose among a set of predefined skills or actions based on feasibility and expected utility. In contrast, agents such as Voyager~\cite{wang2023voyager} represent the action generation paradigm, where executable behaviors are dynamically synthesized, often in open-ended environments, allowing the agent to build a library of composable skills without being constrained by a fixed action space.

With the emergence of function calling, the rise of tools became evident. Most contemporary agents began to define functions as tools through standardized interfaces, with large language models learning to decide when and how to invoke these tools. It is important to distinguish between the concepts of ``action'' and ``tool'': a tool can serve as a unified interface encapsulating multiple actions, and in some cases, a tool may itself be implemented as an agent.

A further shift occurred with the advent of Anthropic's Model Context Protocol (MCP), which introduced a standardized method for tool invocation via a JSON-RPC protocol. The appearance of MCP led to the development of a new generation of tool-calling agents that are constructed on this protocol, enabling agents to reliably discover and utilize external tools in a more dynamic and interoperable manner.

These advancements culminate in the design of our general-purpose task agent AgentOrchestra~\cite{zhang2025agentorchestra}, which is a hierarchical multi-agent framework designed to systematically address the key challenges of generalization, multimodal reasoning, scalability, and collaboration in complex task-solving environments. The framework adopts a two-tier architecture. At the top level, a planning agent is responsible for understanding user tasks, decomposing them into manageable sub-tasks, and orchestrating the workflow by dynamically assigning these sub-tasks to specialized lower-level agents. The lower level consists of several specialized agents: a Deep Analyzer for in-depth analysis of input data and extraction of key insights, a Deep Researcher that conducts comprehensive research and automatically generates research reports or knowledge summaries, and a Browser Use agent that automates browser-based web searches and information extraction to assist the research process. Additionally, a General Tool Calling Agent provides a unified interface for invoking various external tools and APIs through function calling, enabling efficient execution of specific tasks and integration with diverse external services. Through this layered design, the system achieves flexible coordination, efficient information processing, and robust tool integration.

\begin{figure}[htbp]
  \centering
  \includegraphics[width=\textwidth]{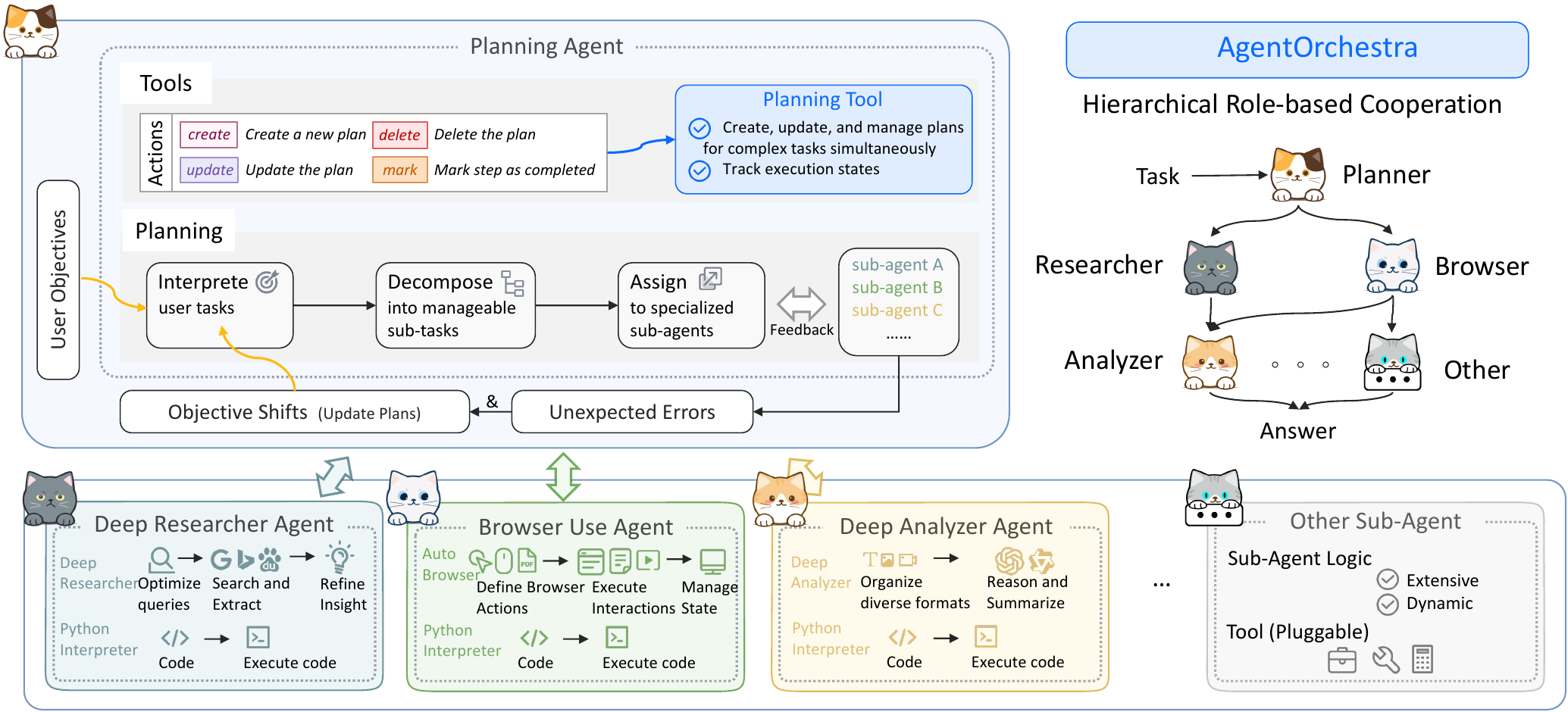}
  \caption{AgentOrchestra Architecture.}
  \label{fig:agentorchestra}
\end{figure}

AgentOrchestra incorporates several innovative technical features to ensure flexible and efficient handling of complex tasks. It leverages asynchronous coroutine scheduling for high-concurrency collaboration among agents, greatly improving system responsiveness and throughput. The framework natively supports seamless switching between leading commercial large models and open-source local models, balancing capability, privacy, and cost. Comprehensive support for both local and remote MCP, enables secure and unified integration of agents and tools across on-premises and cloud environments. Additionally, AgentOrchestra is fully compatible with OpenAI Function Calling and standard JSON invocation, allowing structured, automated cooperation between tools and sub-agents, and significantly enhancing interoperability and automation. 

We have now integrated AgentOrchestra into the \projectname framework. The core modules consist of three parts: agents, models, and tools. The agent module defines the basic async coroutine concurrent agents, as well as the execution logic and prompt templates for deep analyzers, browser-use agents, general agents and so on. As described in Appendix~\ref{appx:model_layer}, we encapsulate the unified interface for commercial LLMs within the LLM API module. The tools module provides several basic tools, such as web crawlers, python code interpreters, arbitrary format file readers, and asynchronous tools like the MCP tools. Furthermore, after integrating the AgentOrchestra modules into \projectname, it can be used to develop other agents. For example, it can be used to build agents that can retrieve the latest news from the internet in real time and make decisions based on the news, or to develop multi-agent systems capable of quantitative company analysis. This framework provides a fundamental infrastructure for developing financial agents and offers the flexibility to create customized financial agents based on individual ideas and creativity.

Our FinAgent is developed based on the general agent, and its prompt template, as described in Trading Prompt Template~\ref{appx:stage_2_prompt_template}, enables the agent to focus on financial decision-making tasks. The main workflow of FinAgent is to retrieve the last two rounds of dialogue history from memory, embed this historical information into the current prompt template, and then use function calling to let the LLM decide whether the action is BUY, HOLD, or SELL. This process is repeated for each day's decision until the task is completed. To fairly compare the open-source FinR1 and our Finreasoner models, we only need to load the corresponding model from the models module.

\section{Architecture of FinWorld}
\label{appx:architecture}

\projectname adopts a layered architecture design where each layer has clear responsibilities and functional boundaries. This layered design not only improves system modularity and maintainability but also provides flexible usage patterns for users at different levels. This section provides a detailed introduction to each layer of \projectname, including the configuration layer, data layer, model layer, task layer, and interface layer.

\subsection{Configuration Layer}

The configuration layer of \projectname is built on top of \texttt{mmengine}~\cite{mmengine2022}, providing a unified and extensible configuration system based on Python dictionaries. This design allows all experimental parameters such as dataset sources, model architectures, training procedures, and evaluation settings to be defined in a centralized, human-readable, and programmatically modifiable format. By leveraging mmengine’s support for hierarchical and modular configuration, users can construct complex experimental setups through inheritance, parameter overriding, and dynamic imports. This approach enhances reproducibility, transparency, and collaboration while separating configuration logic from implementation details.

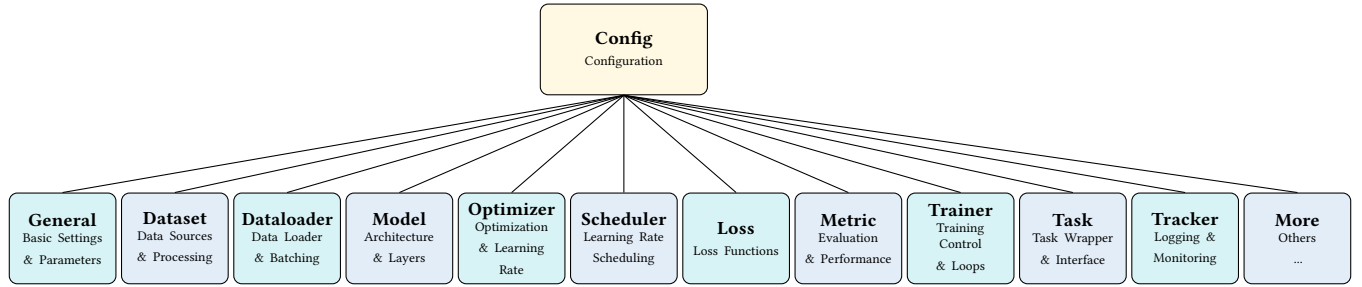
\begin{figure}[htbp]
  \centering
  \begin{tikzpicture}[
    every node/.style={rectangle, rounded corners=3pt, draw, text width=1.2cm, text centered, minimum height=1.2cm, font=\footnotesize}
  ]
    \def\spacing{1.35cm} 
    \def\levelspacing{2.5cm} 
    
    \node[fill=yellow2, text width=2cm] (root) at (0,0) {\textbf{Config} \\ \tiny Configuration};
    
    \node[fill=green2] (general) at ({-5.5*\spacing},-\levelspacing) {\textbf{General} \\ \tiny Basic Settings \\ \tiny \& Parameters};
    \node[fill=blue2] (dataset) at ({-4.4*\spacing},-\levelspacing) {\textbf{Dataset} \\ \tiny Data Sources \\ \tiny \& Processing};
    \node[fill=green2] (dataloader) at ({-3.3*\spacing},-\levelspacing) {\textbf{Dataloader} \\ \tiny Data Loader \\ \tiny \& Batching};
    \node[fill=blue2] (model) at ({-2.2*\spacing},-\levelspacing) {\textbf{Model} \\ \tiny Architecture \\ \tiny \& Layers};
    \node[fill=green2] (optimizer) at ({-1.1*\spacing},-\levelspacing) {\textbf{Optimizer} \\ \tiny Optimization \\ \tiny \& Learning Rate};
    \node[fill=blue2] (scheduler) at (0,-\levelspacing) {\textbf{Scheduler} \\ \tiny Learning Rate \\ \tiny Scheduling};
    \node[fill=green2] (loss) at ({1.1*\spacing},-\levelspacing) {\textbf{Loss} \\ \tiny Loss Functions};
    \node[fill=blue2] (metric) at ({2.2*\spacing},-\levelspacing) {\textbf{Metric} \\ \tiny Evaluation \\ \tiny \& Performance};
    \node[fill=green2] (trainer) at ({3.3*\spacing},-\levelspacing) {\textbf{Trainer} \\ \tiny Training Control \\ \tiny \& Loops};
    \node[fill=blue2] (task) at ({4.4*\spacing},-\levelspacing) {\textbf{Task} \\ \tiny Task Wrapper \\ \tiny \& Interface};
    \node[fill=green2] (tracker) at ({5.5*\spacing},-\levelspacing) {\textbf{Tracker} \\ \tiny Logging \& \\ \tiny Monitoring};
    \node[fill=blue2] (more) at ({6.6*\spacing},-\levelspacing) {\textbf{More} \\ \tiny Others \\ \tiny ...};
    
    \draw (root.south) -- (general.north);
    \draw (root.south) -- (dataset.north);
    \draw (root.south) -- (dataloader.north);
    \draw (root.south) -- (model.north);
    \draw (root.south) -- (optimizer.north);
    \draw (root.south) -- (scheduler.north);
    \draw (root.south) -- (loss.north);
    \draw (root.south) -- (metric.north);
    \draw (root.south) -- (trainer.north);
    \draw (root.south) -- (task.north);
    \draw (root.south) -- (tracker.north);
    \draw (root.south) -- (more.north);
  \end{tikzpicture}
  \caption{Configuration Layer Architecture of FinWorld}
  \label{fig:config_layer}
\end{figure}

To enable flexible and decoupled component management, \projectname adopts a registry mechanism for managing all class declarations. Key components such as \texttt{Dataset} and \texttt{Environment} are registered under a global registry and referenced through the \texttt{type} field in the configuration. When loading configurations, the system retrieves the corresponding class from the registry and instantiates it using the provided arguments. This object-oriented design supports extensibility and consistent runtime behavior across the entire system.

\subsection{Dataset Layer}

The dataset layer of \projectname comprises multiple functional modules designed to enable standardized, extensible, and task-oriented data management for financial AI research. This layer abstracts the complexities of diverse data sources and modalities, offering a unified interface for data acquisition, preprocessing, and task-specific organization. Specifically, it consists of five main modules: data downloader module, data processor module, dataset module, dataloader module, and environment module, each supporting flexible customization and extensibility.

\begin{figure}[htbp]
  \centering
  \begin{tikzpicture}[
    every node/.style={rectangle, rounded corners=3pt, draw, text width=1.2cm, text centered, minimum height=1.2cm, font=\footnotesize}
  ]

    \def\spacing{1.35cm} 
    \def\levelspacing{2.5cm} 
    
    \node[fill=yellow2, text width=2cm] (root) at (0,0) {\textbf{Dataset} \\ \tiny Dataset Sources};
    
    \node[fill=green2] (fmp) at (-6,-\levelspacing) {\textbf{FMP} \\ \tiny Financial \\ \tiny Modeling Prep};
    \node[fill=blue2] (alpaca) at (-4.5,-\levelspacing) {\textbf{Alpaca} \\ \tiny Alpaca \\ \tiny API};
    \node[fill=green2] (akshare) at (-3,-\levelspacing) {\textbf{AkShare} \\ \tiny Chinese \\ \tiny Market Data};
    \node[fill=blue2] (tushare) at (-1.5,-\levelspacing) {\textbf{TuShare} \\ \tiny Chinese \\ \tiny Market Data};
    \node[fill=green2] (more1) at (0,-\levelspacing) {\textbf{More} \\ \tiny Others \\ \tiny ...};
    
    \node[fill=blue2] (finqa) at (1.5,-\levelspacing) {\textbf{FinQA} \\ \tiny Financial \\ \tiny Q\&A Dataset};
    \node[fill=green2] (convfinqa) at (3,-\levelspacing) {\textbf{ConvFinQA} \\ \tiny Conversational \\ \tiny Financial Q\&A};
    \node[fill=blue2] (finance) at (4.5,-\levelspacing) {\textbf{Finance} \\ \tiny Financial \\ \tiny Knowledge};
    \node[fill=green2] (acca) at (6,-\levelspacing) {\textbf{ACCA/CFA} \\ \tiny Professional \\ \tiny Certifications};
    \node[fill=blue2] (more2) at (7.5,-\levelspacing) {\textbf{More} \\ \tiny Others \\ \tiny ...};
    
    \node[fill=yellow2, text width=2.3cm] (aggproc) at (-3,-2 * \levelspacing) {\textbf{AggProcessor} \\ \tiny Market Data \\ \tiny Processing};
    \node[fill=yellow2, text width=2.3cm] (llmproc) at (4.5,-2 * \levelspacing) {\textbf{LLMProcessor} \\ \tiny LLM Data \\ \tiny Processing};
    
    \node[fill=green2] (single) at (-2, -3 * \levelspacing) {\textbf{SingleAsset} \\ \tiny Dataset};
    \node[fill=blue2] (multi) at (2, -3 * \levelspacing) {\textbf{MultiAsset} \\ \tiny Dataset};
    
    \node[fill=yellow2] (dataloader) at (-3,-4 * \levelspacing) {\textbf{DataLoader} \\ \tiny Batch Loading \\ \tiny \& Iteration};
    \node[fill=green2] (trading) at (0,-4 * \levelspacing) {\textbf{Trading} \\ \tiny Trading Environment};
    \node[fill=blue2] (portfolio) at (2,-4 * \levelspacing) {\textbf{Portfolio} \\ \tiny Portfolio Management};
    \node[fill=yellow2] (llmenv) at (4,-4 * \levelspacing) {\textbf{LLM} \\ \tiny LLM Environment};
    
    \draw (root.south) -- (fmp.north);
    \draw (root.south) -- (alpaca.north);
    \draw (root.south) -- (akshare.north);
    \draw (root.south) -- (tushare.north);
    \draw (root.south) -- (more1.north);
    \draw (root.south) -- (finqa.north);
    \draw (root.south) -- (convfinqa.north);
    \draw (root.south) -- (finance.north);
    \draw (root.south) -- (acca.north);
    \draw (root.south) -- (more2.north);
    
    \draw (fmp.south) -- (aggproc.north);
    \draw (alpaca.south) -- (aggproc.north);
    \draw (akshare.south) -- (aggproc.north);
    \draw (tushare.south) -- (aggproc.north);
    \draw (more1.south) -- (aggproc.north);
    
    \draw (finqa.south) -- (llmproc.north);
    \draw (convfinqa.south) -- (llmproc.north);
    \draw (finance.south) -- (llmproc.north);
    \draw (acca.south) -- (llmproc.north);
    \draw (more2.south) -- (llmproc.north);
    
    \draw (aggproc.south) -- (single.north);
    \draw (aggproc.south) -- (multi.north);
    \draw (llmproc.south) -- (single.north);
    \draw (llmproc.south) -- (multi.north);
    
    \draw (single.south) -- (dataloader.north);
    \draw (single.south) -- (trading.north);
    \draw (single.south) -- (portfolio.north);
    \draw (single.south) -- (llmenv.north);
    \draw (multi.south) -- (dataloader.north);
    \draw (multi.south) -- (trading.north);
    \draw (multi.south) -- (portfolio.north);
    \draw (multi.south) -- (llmenv.north);
    
  \end{tikzpicture}
  \caption{Dataset Layer Architecture of FinWorld}
  \label{fig:dataset_layer}
\end{figure}
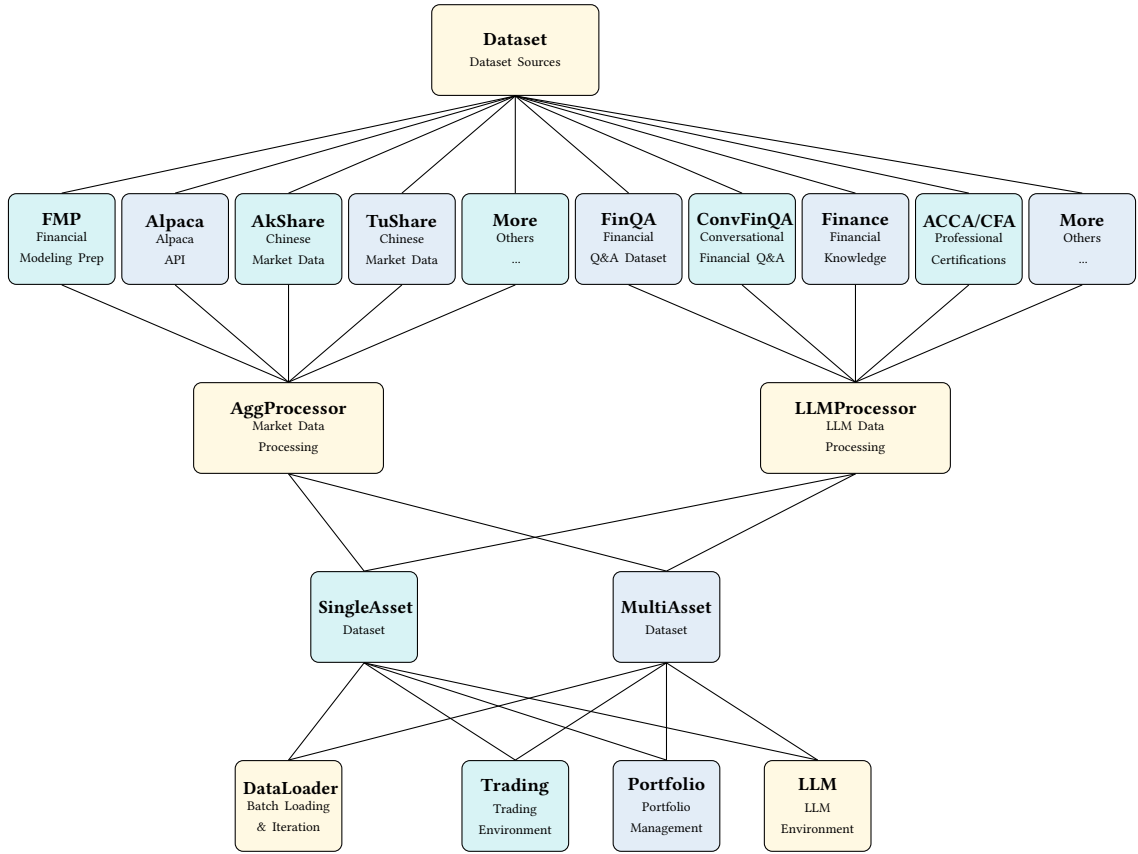

\textbf{Downloader Module.} This module is composed of two tightly integrated components. The first is the Market Data Downloader, which provides a unified interface for heterogeneous financial data providers such as FMP~\cite{fmp2025} and Alpaca~\cite{alpaca2025}. It supports multiple temporal resolutions including daily and minute level data, and accommodates diverse data modalities such as OHLCV time series, fundamentals, news, and alternative signals. The component performs schema harmonization, symbol normalization, and calendar alignment across exchanges. It manages authentication, pagination, and provider specific rate limits, and includes robust retry logic and fault tolerance. All ingested records are validated against a canonical schema and enriched with provenance metadata, including provider, version, retrieval timestamp, and transformation history, in order to ensure complete reproducibility. The second component is the LLM Reasoning Dataset Downloader, which offers programmatic access to financial reasoning corpora such as FinQA, professional certification materials including ACCA and CFA, and curated financial knowledge bases. It enforces licensing checks, version pinning, and deterministic train, validation, and test splits. For textual resources, it provides optional document cleaning, de duplication, and citation preservation. Together, these components implement a single abstraction for acquisition from application programming interfaces and databases, with caching and incremental updates that support scalable and repeatable data collection for research.

\textbf{Processor Module.} Building on standardized access, this module implements configurable preprocessing pipelines for both numerical and textual data. For market data, it supports factor construction such as alpha158~\cite{yang2020qlib}, rolling window feature engineering, calendar and timezone alignment, corporate action adjustment, missing value imputation, outlier handling, and normalization at the asset level or the cross sectional level. Leakage prevention is enforced through proper temporal ordering, expanding windows, and walk forward splits. For reasoning datasets, the module performs tokenization, span alignment, answer normalization, and the construction of chain of thought or rationales when available. It also supports weak supervision signals, label smoothing, and schema mapping across heterogeneous datasets. All pipelines are declared through a concise configuration that is version controlled and executable, enabling exact reproduction of preprocessing steps across experiments. Comprehensive logging records parameters, random seeds, and checksums for every intermediate artifact.

\textbf{Dataset Module.} This module materializes processed data into task specific datasets that are directly consumable by downstream models. For single asset or algorithmic trading tasks, it packages aligned windows of features and targets at the individual asset level with optional masks for market holidays and missing intervals. For portfolio management and multi asset settings, it builds unified panels that synchronize symbols onto a common trading calendar and provide consistent feature spaces, along with instrument level metadata such as sector, exchange, and liquidity attributes. The module natively supports multiple modalities, including numerical time series, structured tabular fundamentals, and unstructured text such as news, filings, and analyst commentary. It can produce formats suitable for supervised learning, sequence to sequence modeling, retrieval augmented generation, and question answering over financial documents. Dataset objects include explicit train, validation, and test partitions with clear time boundaries, as well as standardized evaluation splits for rigorous comparison.

\textbf{Dataloader Module.} This module exposes performant dataloaders for machine learning, deep learning, and applications that rely on large language models. It provides streaming and memory mapped reading for large time series, efficient mini batch construction with sliding or expanding windows, negative and positive sampling strategies, class balancing, and sequence padding with attention masks. The interface integrates naturally with common training loops, supports mixed precision, and enables deterministic shuffling that respects temporal constraints. Collation utilities handle variable length sequences, multi asset batching, and multimodal inputs that combine numerical features with tokenized text. These capabilities reduce input output overhead and simplify integration with external frameworks.

\textbf{Environment Module.} This module encapsulates data within interactive environments designed for reinforcement learning. It provides standardized definitions of episodes, actions, and rewards for tasks such as single asset trading, multi asset allocation, and execution with partial fills. The environment simulates realistic market frictions, including transaction costs, slippage, liquidity constraints, position limits, and optional market impact models. Evaluation protocols include walk forward backtesting, out of sample testing on rolling windows, and live paper trading interfaces for prospective validation. In addition to conventional agents, the environment supports agents driven by large language models through tool interfaces for observation summarization, query of external knowledge, and action justification. All environment configurations are specified declaratively and are accompanied by seeded randomness and event logs, which promotes rigorous ablation studies and reproducibility across a broad range of agent based learning tasks.

\subsection{Model Layer}
\label{appx:model_layer}

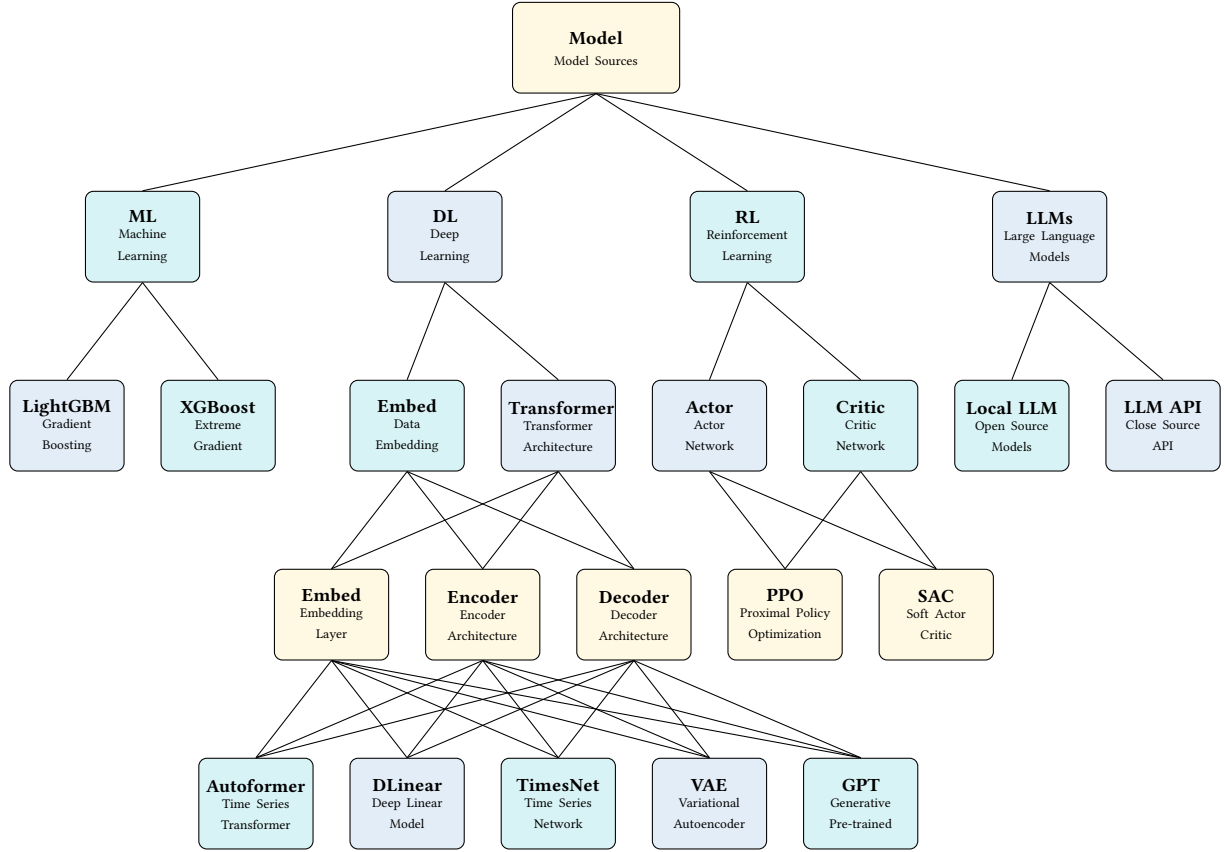
\begin{figure}[htbp]
  \centering
  \begin{tikzpicture}[
    every node/.style={rectangle, rounded corners=3pt, draw, text width=1.3cm, text centered, minimum height=1.2cm, font=\footnotesize}
  ]
    \def\spacing{1.35cm} 
    \def\levelspacing{2.5cm} 
    
    \node[fill=yellow2, text width=2cm] (root) at (0,0) {\textbf{Model} \\ \tiny Model Sources};
    
    \node[fill=green2] (ml) at (-6,- \levelspacing) {\textbf{ML} \\ \tiny Machine \\ \tiny Learning};
    \node[fill=blue2] (dl) at (-2,- \levelspacing) {\textbf{DL} \\ \tiny Deep \\ \tiny Learning};
    \node[fill=green2] (rl) at (2,- \levelspacing) {\textbf{RL} \\ \tiny Reinforcement \\ \tiny Learning};
    \node[fill=blue2] (llms) at (6,- \levelspacing) {\textbf{LLMs} \\ \tiny Large Language \\ \tiny Models};
    
    \node[fill=blue2] (lightgbm) at (-7,-2 * \levelspacing) {\textbf{LightGBM} \\ \tiny Gradient \\ \tiny Boosting};
    \node[fill=green2] (xgboost) at (-5,-2 * \levelspacing) {\textbf{XGBoost} \\ \tiny Extreme \\ \tiny Gradient};
    
    \node[fill=green2] (embed2) at (-2.5,-2 * \levelspacing) {\textbf{Embed} \\ \tiny Data \\ \tiny Embedding};
    \node[fill=blue2] (transformer) at (-0.5,-2 * \levelspacing) {\textbf{Transformer} \\ \tiny Transformer \\ \tiny Architecture};
    
    \node[fill=blue2] (actor) at (1.5,-2 * \levelspacing) {\textbf{Actor} \\ \tiny Actor \\ \tiny Network};
    \node[fill=green2] (critic) at (3.5,-2 * \levelspacing) {\textbf{Critic} \\ \tiny Critic \\ \tiny Network};
    
    \node[fill=green2] (localllm) at (5.5,-2 * \levelspacing) {\textbf{Local LLM} \\ \tiny Open Source \\ \tiny Models};
    \node[fill=blue2] (remotellm) at (7.5,-2 * \levelspacing) {\textbf{LLM API} \\ \tiny Close Source \\ \tiny API};
    
    \node[fill=yellow2] (embed) at (-3.5,-3 * \levelspacing) {\textbf{Embed} \\ \tiny Embedding \\ \tiny Layer};
    \node[fill=yellow2] (encoder) at (-1.5,-3 * \levelspacing) {\textbf{Encoder} \\ \tiny Encoder \\ \tiny Architecture};
    \node[fill=yellow2] (decoder) at (0.5,-3 * \levelspacing) {\textbf{Decoder} \\ \tiny Decoder \\ \tiny Architecture};
    
    \node[fill=yellow2] (ppo) at (2.5,-3 * \levelspacing) {\textbf{PPO} \\ \tiny Proximal Policy \\ \tiny Optimization};
    \node[fill=yellow2] (sac) at (4.5,-3 * \levelspacing) {\textbf{SAC} \\ \tiny Soft Actor \\ \tiny Critic};
    
    \node[fill=green2] (autoformer) at (-4.5,-4 * \levelspacing) {\textbf{Autoformer} \\ \tiny Time Series \\ \tiny Transformer};
    \node[fill=blue2] (dlinear) at (-2.5,-4 * \levelspacing) {\textbf{DLinear} \\ \tiny Deep Linear \\ \tiny Model};
    \node[fill=green2] (timesnet) at (-0.5,-4 * \levelspacing) {\textbf{TimesNet} \\ \tiny Time Series \\ \tiny Network};
    \node[fill=blue2] (vae) at (1.5,-4 * \levelspacing) {\textbf{VAE} \\ \tiny Variational \\ \tiny Autoencoder};
    \node[fill=green2] (gpt) at (3.5,-4 * \levelspacing) {\textbf{GPT} \\ \tiny Generative \\ \tiny Pre-trained};

    \draw (root.south) -- (ml.north);
    \draw (root.south) -- (dl.north);
    \draw (root.south) -- (rl.north);
    \draw (root.south) -- (llms.north);
    
    \draw (ml.south) -- (lightgbm.north);
    \draw (ml.south) -- (xgboost.north);
    
    \draw (dl.south) -- (embed2.north);
    \draw (dl.south) -- (transformer.north);
    
    \draw (rl.south) -- (actor.north);
    \draw (rl.south) -- (critic.north);
    
    \draw (llms.south) -- (localllm.north);
    \draw (llms.south) -- (remotellm.north);
    
    \draw (embed2.south) -- (embed.north);
    \draw (embed2.south) -- (encoder.north);
    \draw (embed2.south) -- (decoder.north);
    
    \draw (transformer.south) -- (embed.north);
    \draw (transformer.south) -- (encoder.north);
    \draw (transformer.south) -- (decoder.north);
    
    \draw (actor.south) -- (ppo.north);
    \draw (actor.south) -- (sac.north);
    \draw (critic.south) -- (ppo.north);
    \draw (critic.south) -- (sac.north);
    
    \draw (embed.south) -- (autoformer.north);
    \draw (embed.south) -- (dlinear.north);
    \draw (embed.south) -- (timesnet.north);
    \draw (embed.south) -- (vae.north);
    \draw (embed.south) -- (gpt.north);
    
    \draw (encoder.south) -- (autoformer.north);
    \draw (encoder.south) -- (dlinear.north);
    \draw (encoder.south) -- (timesnet.north);
    \draw (encoder.south) -- (vae.north);
    \draw (encoder.south) -- (gpt.north);
    
    \draw (decoder.south) -- (autoformer.north);
    \draw (decoder.south) -- (dlinear.north);
    \draw (decoder.south) -- (timesnet.north);
    \draw (decoder.south) -- (vae.north);
    \draw (decoder.south) -- (gpt.north);
    
  \end{tikzpicture}
  \caption{Model Layer Architecture of FinWorld}
  \label{fig:model_layer}
\end{figure}

The model layer of \projectname comprises a set of specialized yet interoperable modules that support a unified definition, management, and invocation of heterogeneous modeling paradigms across the platform. This layer abstracts both construction and orchestration for traditional machine learning, deep neural architectures, and LLMs, exposing standardized interfaces for training, inference, and downstream integration in financial AI workflows. By decoupling model specification from execution and deployment, it enables consistent behavior across tasks, reproducible configurations, and smooth interchangeability of components.

\textbf{Machine Learning Models.}
This module provides a declarative specification for classical models under a uniform schema. Supported classes include linear and logistic regression, decision trees, random forests, and gradient‑boosting ensembles (e.g., XGBoost and LightGBM). Each model is described by a consistent input/output contract and an explicit task head (regression, binary classification, or multiclass classification). Structural attributes, such as regularization choices, tree growth policy, and ensemble topology, are expressed as first‑class parameters, enabling transparent comparison across alternatives and systematic hyperparameter control. The uniform abstraction ensures that training and scoring pipelines, metrics, and model persistence follow the same conventions regardless of the underlying estimator.

\textbf{Deep Learning Models.}
Neural architectures are organized into composable \emph{components}, \emph{layers}, and \emph{models} with stable interfaces. Components include data embedding for financial time series~\cite{wu2022timesnet}, patch embedding~\cite{dosovitskiy2020image}, positional encodings, attention mechanisms, transformer blocks, and common activation functions. Layers assemble these components into an embedding layer, an encoder, and a decoder that can be mixed and matched without changing external contracts. Networks such as Autoformer, VAE, and GPT‑style decoders are realized by assembling the layers under one specification that remains independent of training and inference. This separation supports rapid iteration on architecture variants while preserving consistent preprocessing, batching, and evaluation semantics.

\textbf{Reinforcement Learning Models.}
This module defines RL network structures using an actor–critic abstraction. Policy and value networks are composed from the Deep Learning components, with backbones chosen from MLP, LSTM/GRU, or Transformer families. Specifications cover discrete or continuous action heads and value heads, allow shared or separate encoders, and optionally incorporate memory. Financial constraints are captured as structural hooks (transaction costs, slippage and market impact, risk limits, and trading calendars) so that environment dynamics and trading/portfolio objectives can be expressed alongside the network definition. The result is a consistent interface for simulation and live inference that aligns with the rest of the platform’s training and serving patterns.

\textbf{LLMs Models.}
This module centralizes access to proprietary and open‑source LLMs behind a unified interface. It supports seamless switching among commercial providers (e.g., GPT‑4.1, Claude‑4‑Sonnet, Gemini‑2.5‑Pro) and efficient local models (e.g., Qwen2.5, Qwen3), with standardized controls for context length, sampling behavior, and cost/latency tracking. Built‑in capabilities include function calling, tool use, and retrieval‑augmented generation tailored to financial documents. The module integrates naturally with agents and downstream models, enabling document understanding, information extraction, and instruction following under the same invocation patterns as other \projectname models. This symmetry reduces integration overhead and promotes consistent monitoring and evaluation across modalities.

\subsection{Training Layer}

\begin{figure}[htbp]
  \centering
  \begin{tikzpicture}[
    every node/.style={rectangle, rounded corners=3pt, draw, text width=1.2cm, text centered, minimum height=1.2cm, font=\footnotesize}
  ]
    \def\spacing{1.5cm} 
    \def\levelspacing{2.5cm} 

    \node[fill=yellow2, text width=2cm] (root) at (0,0) {\textbf{Training} \\ \tiny Training Layer};
    
    \node[fill=green2] (optimizer) at (-6,-\levelspacing) {\textbf{Optimizer} \\ \tiny Optimization \\ \tiny Algorithms};
    \node[fill=blue2] (scheduler) at (-3,-\levelspacing) {\textbf{Scheduler} \\ \tiny Learning Rate \\ \tiny Scheduling};
    \node[fill=green2] (loss) at (0,-\levelspacing) {\textbf{Loss} \\ \tiny Loss Functions \\ \tiny \& Objectives};
    \node[fill=blue2] (metrics) at (3,-\levelspacing) {\textbf{Metrics} \\ \tiny Evaluation \\ \tiny \& Monitoring};
    \node[fill=green2] (trainer) at (6,-\levelspacing) {\textbf{Trainer} \\ \tiny Training \\ \tiny Pipeline};
    
    \node[fill=green2] (adamw) at (-5.5*\spacing,-2 * \levelspacing) {\textbf{AdamW} \\ \tiny Adaptive \\ \tiny Optimizer};
    \node[fill=blue2] (adam) at (-4.5*\spacing,-2 * \levelspacing) {\textbf{Adam} \\ \tiny Adaptive \\ \tiny Optimizer};
    \node[fill=green2] (cosine) at (-3.5*\spacing,-2 * \levelspacing) {\textbf{Cosine} \\ \tiny Cosine \\ \tiny Scheduler};
    \node[fill=blue2] (linear) at (-2.5*\spacing,-2 * \levelspacing) {\textbf{Linear} \\ \tiny Linear \\ \tiny Scheduler};
    \node[fill=green2] (mae) at (-1.5*\spacing,-2 * \levelspacing) {\textbf{MAE Loss} \\ \tiny Mean Abs \\ \tiny Error};
    \node[fill=blue2] (mse) at (-0.5*\spacing,-2 * \levelspacing) {\textbf{MSE Loss} \\ \tiny Mean Squared \\ \tiny Error};
    \node[fill=green2] (acc) at (0.5*\spacing,-2 * \levelspacing) {\textbf{ACC} \\ \tiny Accuracy \\ \tiny Metric};
    \node[fill=blue2] (arr) at (1.5*\spacing,-2 * \levelspacing) {\textbf{ARR} \\ \tiny Annual Return \\ \tiny Rate};
    \node[fill=green2] (msemetric) at (2.5*\spacing,-2 * \levelspacing) {\textbf{MSE} \\ \tiny Mean Squared \\ \tiny Error};
    \node[fill=blue2] (train) at (3.5*\spacing,-2 * \levelspacing) {\textbf{Train} \\ \tiny Training \\ \tiny Process};
    \node[fill=green2] (valid) at (4.5*\spacing,-2 * \levelspacing) {\textbf{Valid} \\ \tiny Validation \\ \tiny Process};
    \node[fill=blue2] (test) at (5.5*\spacing,-2 * \levelspacing) {\textbf{Test} \\ \tiny Testing \\ \tiny Process};
    
    \draw (root.south) -- (optimizer.north);
    \draw (root.south) -- (scheduler.north);
    \draw (root.south) -- (loss.north);
    \draw (root.south) -- (metrics.north);
    \draw (root.south) -- (trainer.north);
    
    \draw (optimizer.south) -- (adamw.north);
    \draw (optimizer.south) -- (adam.north);
    
    \draw (scheduler.south) -- (cosine.north);
    \draw (scheduler.south) -- (linear.north);
    
    \draw (loss.south) -- (mae.north);
    \draw (loss.south) -- (mse.north);
    
    \draw (metrics.south) -- (acc.north);
    \draw (metrics.south) -- (arr.north);
    \draw (metrics.south) -- (msemetric.north);
    
    \draw (trainer.south) -- (train.north);
    \draw (trainer.south) -- (valid.north);
    \draw (trainer.south) -- (test.north);
    
  \end{tikzpicture}
  \caption{Training Layer Architecture of FinWorld}
  \label{fig:training_layer}
\end{figure}
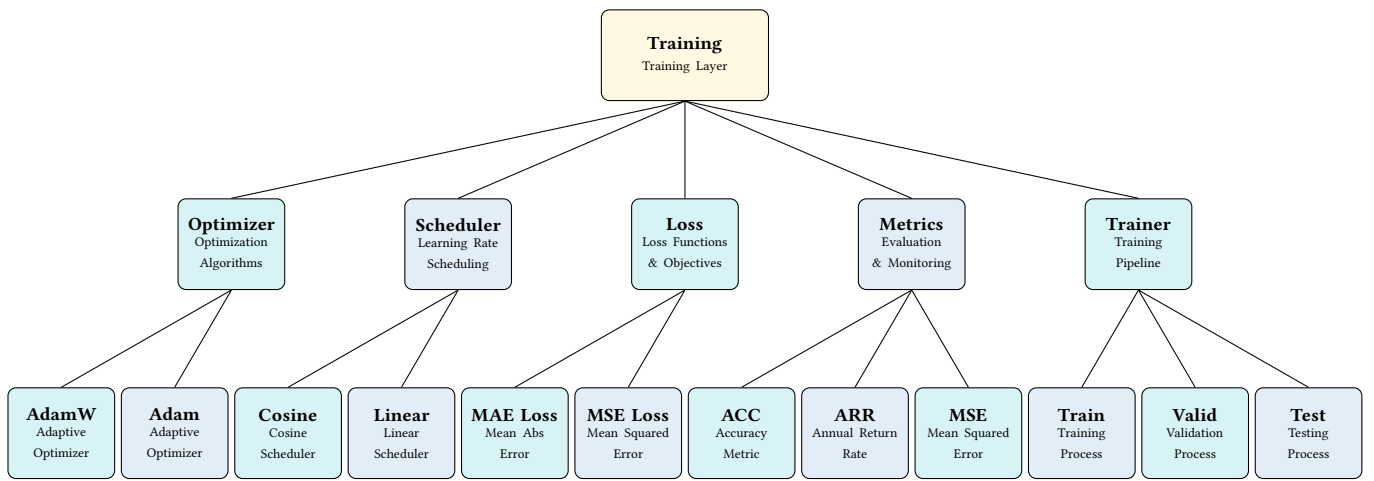

The training layer in \projectname offers a modular scaffold that abstracts every element required to optimise all method pipelines for financial applications. A uniform interface guarantees experiment reproducibility, smooth scaling from single GPU notebooks to distributed clusters, and quick transfer of best practices across tasks such as time series forecasting, trading, portfolio management, and LLM applications.

\textbf{Optimizer}. A rich catalogue of first order methods, ranging from classic stochastic gradient descent (SGD) to adaptive variants such as Adam and AdamW, lets practitioners match optimisation dynamics to task characteristics. A common wrapper normalises hyperparameter signatures and supports gradient centralisation, decoupled weight decay, and mixed precision updates, ensuring robust convergence on noisy and non stationary market data.

\textbf{Loss}. A flexible factory produces objective functions that cover regression losses (mean squared error and mean absolute error), classification losses for event prediction, and reinforcement learning surrogates for policy and value objectives. Composite losses can be declared with a single line, enabling multi task learning or the joint optimisation of risk adjusted return and prediction accuracy.

\textbf{Scheduler}. Static strategies (step, cosine, linear) and adaptive schemes (warm up then decay, reduce on plateau) are available for both learning rates and regularisation coefficients. Schedulers are time aware, resuming the exact trajectory when checkpoints are reloaded.

\textbf{Metrics}. In addition to generic accuracy and error scores, the library ships with finance specific diagnostics such as annual return, Sharpe ratio, maximum drawdown, and tail risk measures. Each metric is logged at configurable intervals and can trigger early stopping or hyper parameter sweeps, enabling data driven model selection.

\textbf{Trainer}. Acting as the orchestrator, the trainer pipes data loading, forward and backward passes, gradient clipping, metric evaluation, checkpointing, and experiment logging (TensorBoard or WandB). Specialised variants provide task specific logic, for example the forecasting trainer, trading trainer, portfolio trainer, and large language model trainer. Clear callback hooks let researchers inject custom steps, such as on the fly data augmentation or bespoke risk constraints, without changing the core loop.

Together, these components form a coherent architecture that accelerates experimentation, strengthens reproducibility, and lowers the barrier to launching state of the art AI solutions in the demanding environment of financial markets.

\subsection{Evaluation Layer}

The evaluation layer of \projectname\ is designed to provide a unified, extensible, and highly modular evaluation system for financial AI tasks. At its core, the evaluation layer organizes all assessment methods into two main branches: \texttt{metrics} and \texttt{plot}, each responsible for different but complementary aspects of model evaluation. 

\textbf{Metrics}. The metrics defined in the training layer are reused in the evaluation implementation.

\textbf{Plot}. The plot module provides advanced visualization tools that enable intuitive analysis and diagnosis of model performance. Supported plot types include K-line (candlestick) charts, cumulative return curves, compass plots, star plots, and sunburst diagrams. This modular design ensures that both quantitative and qualitative evaluation needs are met in a systematic and reproducible manner.

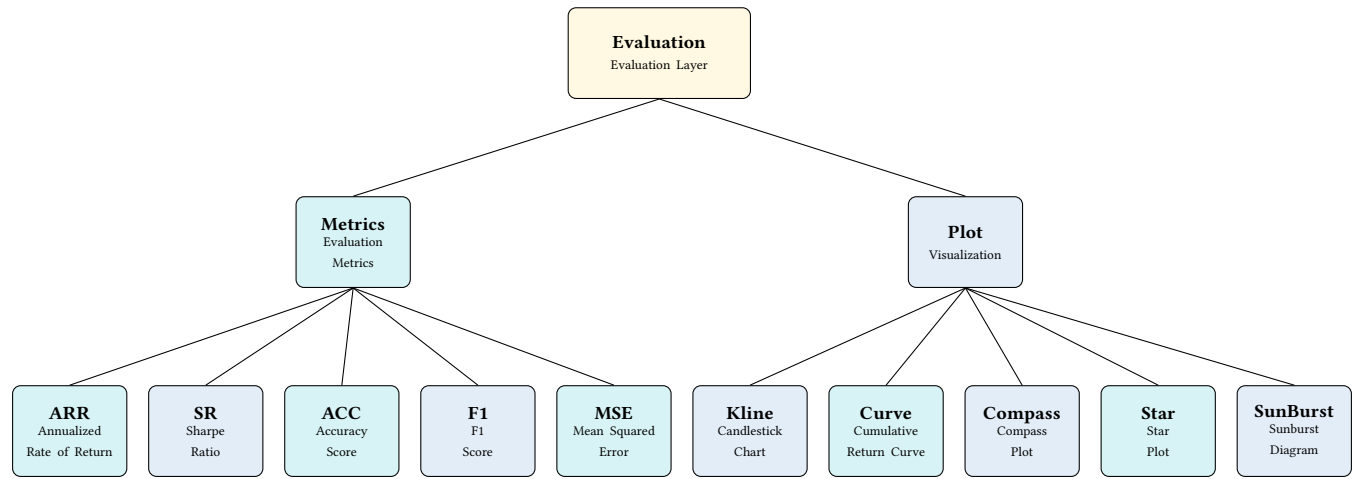
\begin{figure}[htbp]
  \centering
  \begin{tikzpicture}[
    every node/.style={rectangle, rounded corners=3pt, draw, text width=1.3cm, text centered, minimum height=1.2cm, font=\footnotesize}
  ]
    \def\spacing{1.5cm} 
    \def\levelspacing{2.5cm} 

    \node[fill=yellow2, text width=2.2cm] (root) at (0,0) {\textbf{Evaluation} \\ \tiny Evaluation Layer};

    \node[fill=green2] (metrics) at (-2.7*\spacing,-\levelspacing) {\textbf{Metrics} \\ \tiny Evaluation \\ \tiny Metrics};
    \node[fill=blue2]  (plot)    at ( 2.7*\spacing,-\levelspacing) {\textbf{Plot} \\ \tiny Visualization};

    \node[fill=green2] (arr) at (-5.2*\spacing,-2* \levelspacing) {\textbf{ARR} \\ \tiny Annualized \\ \tiny Rate of Return};
    \node[fill=blue2]  (sr)  at (-4.0*\spacing,-2* \levelspacing) {\textbf{SR} \\ \tiny Sharpe \\ \tiny Ratio};
    \node[fill=green2] (acc) at (-2.8*\spacing,-2* \levelspacing) {\textbf{ACC} \\ \tiny Accuracy \\ \tiny Score};
    \node[fill=blue2]  (f1)  at (-1.6*\spacing,-2* \levelspacing) {\textbf{F1} \\ \tiny F1 \\ \tiny Score};
    \node[fill=green2] (mse) at (-0.4*\spacing,-2* \levelspacing) {\textbf{MSE} \\ \tiny Mean Squared \\ \tiny Error};

    \node[fill=blue2]  (kline)    at (0.8*\spacing,-2* \levelspacing) {\textbf{Kline} \\ \tiny Candlestick \\ \tiny Chart};
    \node[fill=green2] (curve)    at (2.0*\spacing,-2* \levelspacing) {\textbf{Curve} \\ \tiny Cumulative \\ \tiny Return Curve};
    \node[fill=blue2]  (compass)  at (3.2*\spacing,-2* \levelspacing) {\textbf{Compass} \\ \tiny Compass \\ \tiny Plot};
    \node[fill=green2] (star)     at (4.4*\spacing,-2* \levelspacing) {\textbf{Star} \\ \tiny Star \\ \tiny Plot};
    \node[fill=blue2]  (sunburst) at (5.6*\spacing,-2* \levelspacing) {\textbf{SunBurst} \\ \tiny Sunburst \\ \tiny Diagram};

    \draw (root.south) -- (metrics.north);
    \draw (root.south) -- (plot.north);

    \draw (metrics.south) -- (arr.north);
    \draw (metrics.south) -- (sr.north);
    \draw (metrics.south) -- (acc.north);
    \draw (metrics.south) -- (f1.north);
    \draw (metrics.south) -- (mse.north);

    \draw (plot.south) -- (kline.north);
    \draw (plot.south) -- (curve.north);
    \draw (plot.south) -- (compass.north);
    \draw (plot.south) -- (star.north);
    \draw (plot.south) -- (sunburst.north);

  \end{tikzpicture}
  \caption{Evaluation Layer Architecture of FinWorld}
  \label{fig:evaluation_layer}
\end{figure}

This adaptive evaluation design simplifies the extension and customization of assessment routines, while ensuring consistent and standardized evaluation across the platform. By streamlining model assessment for diverse financial tasks, it enables systematic comparison, rapid diagnosis of strengths and weaknesses, and iterative improvement of financial AI methods. In our implementation, the evaluation layer is integrated into the trainer’s validation and test stages, allowing metric and plotting methods to be shared with the training phase.

\subsection{Task Layer}

\begin{figure}[htbp]
  \centering
  \begin{tikzpicture}[
    every node/.style={
      rectangle, rounded corners=3pt, draw, 
      text width=1.2cm, text centered, 
      minimum height=1.2cm, font=\footnotesize
    }
  ]
    \def\spacing{1.35cm}      
    \def\levelspacing{2.5cm}  

    \node[fill=yellow2, text width=2cm] (root) at (0,0) {\textbf{Task} \\ \tiny Task Layer};

    \node[fill=green2] (train) at ({-1*\spacing}, -\levelspacing) {\textbf{Train} \\ \tiny Training};
    \node[fill=blue2]  (test)  at ({1*\spacing}, -\levelspacing)  {\textbf{Test} \\ \tiny Testing};

    \node[fill=green2] (forecast)  at ({-2.7*\spacing}, -2*\levelspacing) {\textbf{Forecasting} \\ \tiny Time Series};
    \node[fill=blue2]  (trading)   at ({-1.35*\spacing}, -2*\levelspacing) {\textbf{Trading} \\ \tiny Algo Trading};
    \node[fill=green2] (portfolio) at (0, -2*\levelspacing) {\textbf{Portfolio} \\ \tiny Management};
    \node[fill=blue2]  (llms)      at ({1.35*\spacing}, -2*\levelspacing) {\textbf{LLMs} \\ \tiny FM Tasks};
    \node[fill=green2] (agent)     at ({2.7*\spacing}, -2*\levelspacing) {\textbf{Agents} \\ \tiny Agent};

    \draw (root.south) -- (train.north);
    \draw (root.south) -- (test.north);

    \draw (train.south) -- (forecast.north);
    \draw (train.south) -- (trading.north);
    \draw (train.south) -- (portfolio.north);
    \draw (train.south) -- (llms.north);
    \draw (train.south) -- (agent.north);

    \draw (test.south) -- (forecast.north);
    \draw (test.south) -- (trading.north);
    \draw (test.south) -- (portfolio.north);
    \draw (test.south) -- (llms.north);
    \draw (test.south) -- (agent.north);

  \end{tikzpicture}
  \caption{Task Layer Architecture of FinWorld}
  \label{fig:task_layer}
\end{figure}
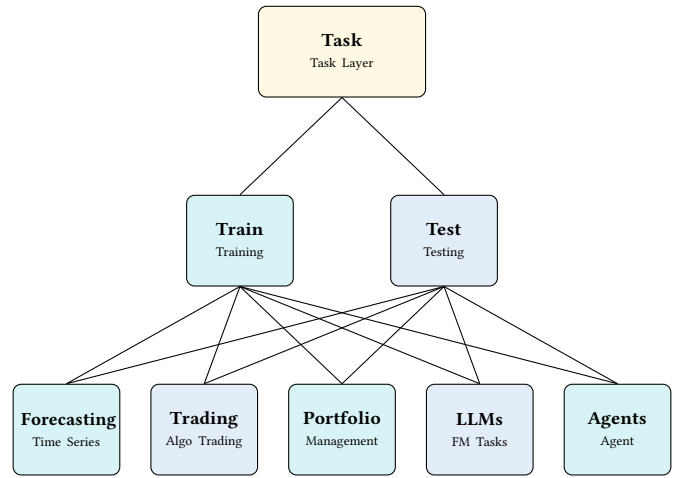

At its core, the task layer is responsible for the systematic definition, abstraction, and encapsulation of financial AI task types. It systematically supports a wide spectrum of financial AI tasks, providing unified abstractions and modular interfaces that facilitate integration with upstream data modules and downstream modeling components. This layer centers on several core financial tasks, including time series forecasting, algorithmic trading, portfolio management, and LLMs applications, which are formally defined in Section~\ref{sec:preliminaries}. 

Each task is specified by configurable input/output schemas and standardized evaluation protocols, supporting both established benchmarks and user-customized scenarios. The unified task architecture enables rapid prototyping, cross-task generalization, and reproducible research across financial AI applications. Additionally, the layer provides a flexible framework for deploying LLM-based agents in asynchronous, single-agent, or multi-agent configurations, allowing users to compose and customize financial agents for diverse research and application needs.

\subsection{Presentation Layer}

\begin{figure}[htbp]
  \centering
  \begin{tikzpicture}[
    every node/.style={
      rectangle, rounded corners=3pt, draw, 
      text width=1.2cm, text centered, 
      minimum height=1.2cm, font=\footnotesize
    }
  ]
    \def\spacing{1.5cm}      
    \def\levelspacing{2.5cm}  

    \node[fill=yellow2, text width=2cm] (root) at (0,0) {\textbf{Presentation} \\ \tiny Presentation Layer};

    \node[fill=green2] (github) at ({-3*\spacing}, -\levelspacing) {\textbf{Github} \\ \tiny Repo};
    \node[fill=blue2]  (html)   at ({-2*\spacing}, -\levelspacing) {\textbf{HTML} \\ \tiny Web Page};
    \node[fill=green2] (latex)  at ({-1*\spacing}, -\levelspacing) {\textbf{LaTeX} \\ \tiny Tech Report};
    \node[fill=blue2]  (pdf)    at (0, -\levelspacing)             {\textbf{PDF} \\ \tiny Report};
    \node[fill=green2] (png)    at ({1*\spacing}, -\levelspacing)  {\textbf{PNG} \\ \tiny Figures};
    \node[fill=blue2]  (csv)    at ({2*\spacing}, -\levelspacing)  {\textbf{CSV} \\ \tiny Tables};
    \node[fill=green2] (json)   at ({3*\spacing}, -\levelspacing)  {\textbf{JSON} \\ \tiny Artifacts};

    \draw (root.south) -- (github.north);
    \draw (root.south) -- (html.north);
    \draw (root.south) -- (latex.north);
    \draw (root.south) -- (pdf.north);
    \draw (root.south) -- (png.north);
    \draw (root.south) -- (csv.north);
    \draw (root.south) -- (json.north);

  \end{tikzpicture}
  \caption{Presentation Layer Architecture of FinWorld}
  \label{fig:presentation_layer}
\end{figure}
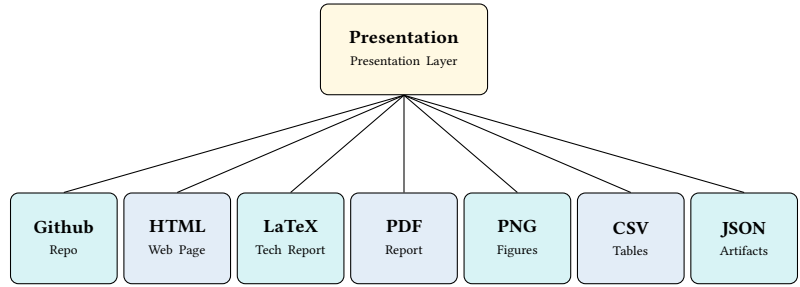

The presentation layer in \projectname is designed for automated and multi-modal dissemination and documentation of experimental results. At the center of this layer is a dedicated presentation agent, which coordinates the aggregation of evaluation outputs, the automatic generation of technical reports, and the creation of interactive web pages for result interpretation and sharing. Experimental findings, along with key visualizations and benchmark summaries, are systematically compiled into structured documents and published to collaborative platforms such as GitHub and GitHub Pages (github.io), ensuring transparent sharing and long-term accessibility. Seamless integration with experiment tracking tools like Wandb also enables real-time visualization and comparison of metrics throughout the research lifecycle.

This automated presentation pipeline is highly configurable. Users can customize report templates, select specific evaluation results to highlight, and choose the output format that best fits their audience, including detailed PDF or LaTeX technical reports, interactive dashboards, and web-based summaries. The layer supports multi-channel publishing, allowing results to be distributed simultaneously across research repositories, organizational knowledge bases, and public web platforms. This significantly expands the reach and impact of the research. By centralizing the documentation and presentation process, the layer ensures that all experimental records, plots, and analyses are consistently organized and easily retrievable, supporting systematic archiving and knowledge transfer.

The presentation layer also emphasizes both human and machine interpretability. Generated reports include not only visual and tabular summaries, but also machine-readable artifacts such as JSON or CSV exports, which can be used for downstream analysis or automated benchmarking. Support for version control and changelog generation makes it straightforward to track the evolution of results over time, aiding reproducibility and collaborative development. By combining automation, configurability, and multi-modal output, the presentation layer streamlines communication, supports peer review and compliance, and enhances the visibility, transparency, and long-term value of financial AI research within the platform.

\end{document}